\PassOptionsToPackage{dvipsnames}{xcolor}
\documentclass[singlecolumn]{NTUAI4XPaperEN}

\usepackage{mathtools}
\usepackage{placeins}
\usepackage[english]{babel}
\usepackage{xspace}
\setthemecolor{19365b}

\usepackage{booktabs}     
\usepackage{algorithm}
\usepackage{algpseudocode}
\usepackage{adjustbox}
\usepackage{wrapfig}

\usepackage{microtype}      

\usepackage{graphicx}
\usepackage{subcaption} 

\usepackage{amsthm}

\usepackage{multirow}
\usepackage{makecell}
\usepackage{titletoc}

\crefname{section}{\S}{\S}
\crefname{table}{Tab.}{Tabs.}
\crefname{figure}{Fig.}{Figs.}

\Crefname{table}{Table}{Tables}
\crefname{equation}{Eq.}{Eqs.}
\Crefname{equation}{Eq.}{Eqs.}
\crefname{algorithm}{Alg.}{Algs.}
\Crefname{algorithm}{Alg.}{Algs.}

\definecolor{oursblue}{RGB}{235,243,251}
\definecolor{oursgreen}{RGB}{235,247,237}
\definecolor{oursgreen1}{RGB}{34,85,60}
\definecolor{blockgray}{RGB}{244,245,246}

\usepackage{tcolorbox}
\tcbuselibrary{breakable}
\newcommand{\pvar}[1]{%
	\textcolor{oursgreen1}{\texttt{<#1>}}%
}
\newtcolorbox{promptbox}[1]{
	breakable,
	colback=blockgray,
	colframe=black!18,
	boxrule=0.45pt,
	arc=1.5pt,
	left=5pt,
	right=5pt,
	top=4pt,
	bottom=4pt,
	before skip=4pt,
	after skip=5pt,
	title={#1},
	fonttitle=\small\bfseries,
	coltitle=black
}

\titlecontents{section}
[2em]
{\bfseries}
{\contentslabel{1.8em}}
{}
{\titlerule*[0.8pc]{.}\contentspage}

\titlecontents{subsection}
[4.5em]
{}
{\contentslabel{3em}}
{}
{\titlerule*[0.8pc]{.}\contentspage}

\definecolor{Light}{rgb}{0.92, 0.99, 0.95}
\definecolor{mygray}{HTML}{E6E6FA}

\definecolor{gvnavy}{RGB}{1,33,105}
\definecolor{gvamber}{RGB}{232,163,61}
\definecolor{gvamberd}{RGB}{199,127,26}
\definecolor{gvice}{RGB}{242,245,250}
\definecolor{gvgreen}{RGB}{63,128,63}
\definecolor{gvred}{RGB}{178,34,34}
\definecolor{gvgray}{RGB}{107,118,137}
\definecolor{gvblue}{RGB}{53,108,170}
\definecolor{rowhl}{RGB}{251,241,221}
\definecolor{rowref}{RGB}{238,238,238}

\setthemecolor{1a2a6c}

\titlelogocontent{%
	\begin{minipage}{\textwidth}
		\begin{minipage}[c]{0.35\textwidth}
			\raggedright
			\includegraphics[height=1cm]{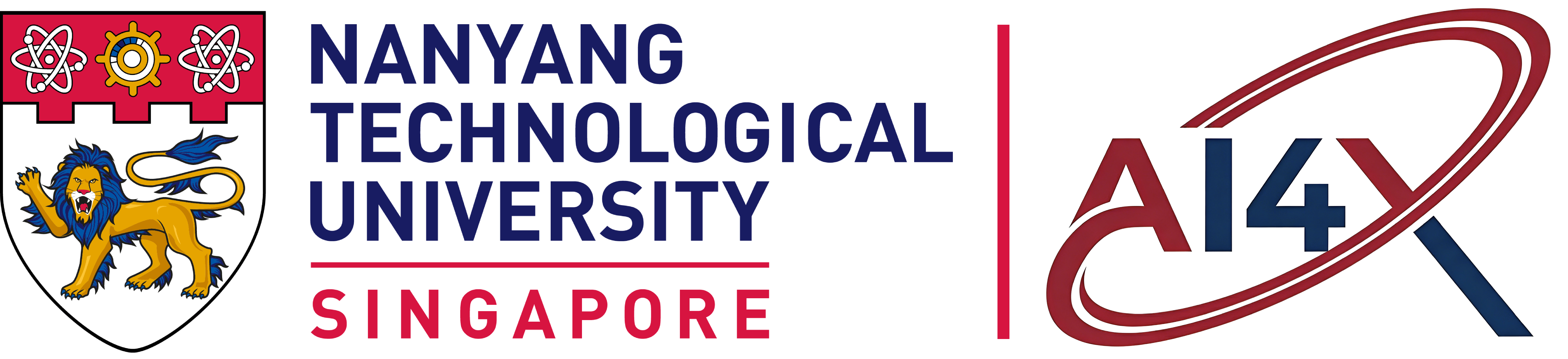}
		\end{minipage}%
		\hfill
		\begin{minipage}[c]{0.60\textwidth}
			\raggedleft
			\includegraphics[height=1cm]{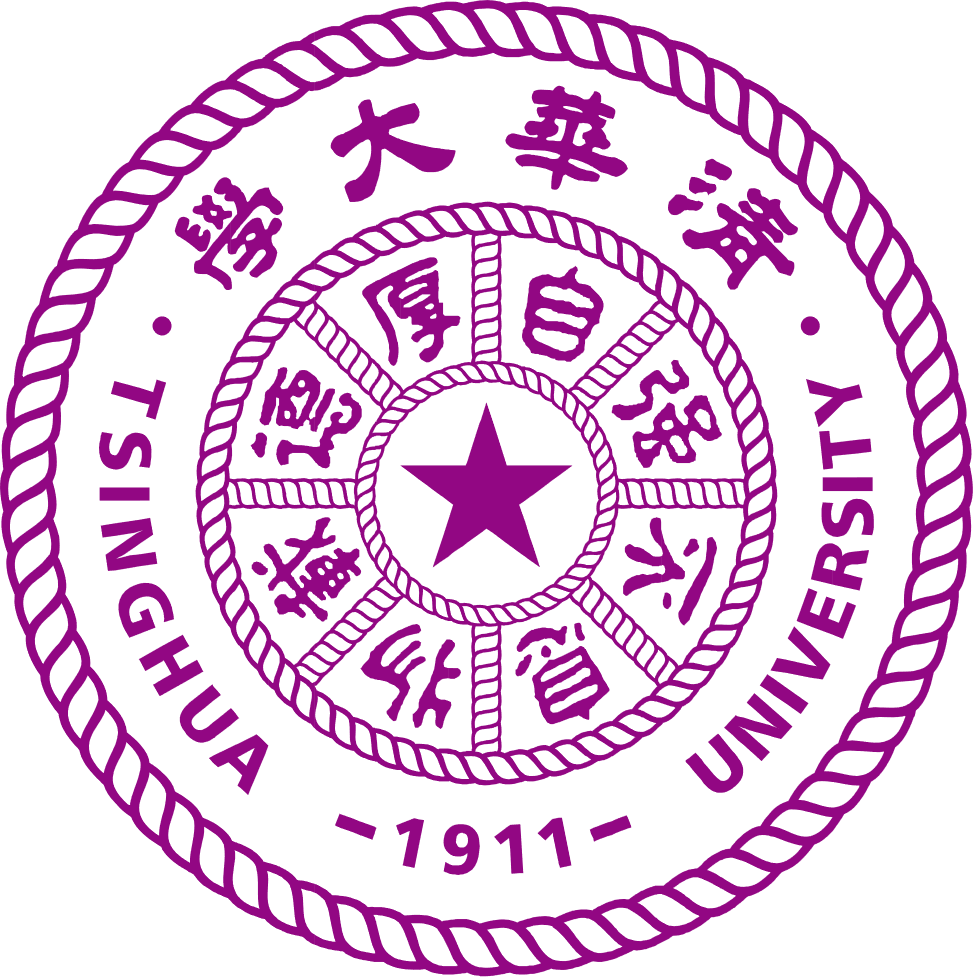}\hspace{0.2em}
			\includegraphics[height=1cm]{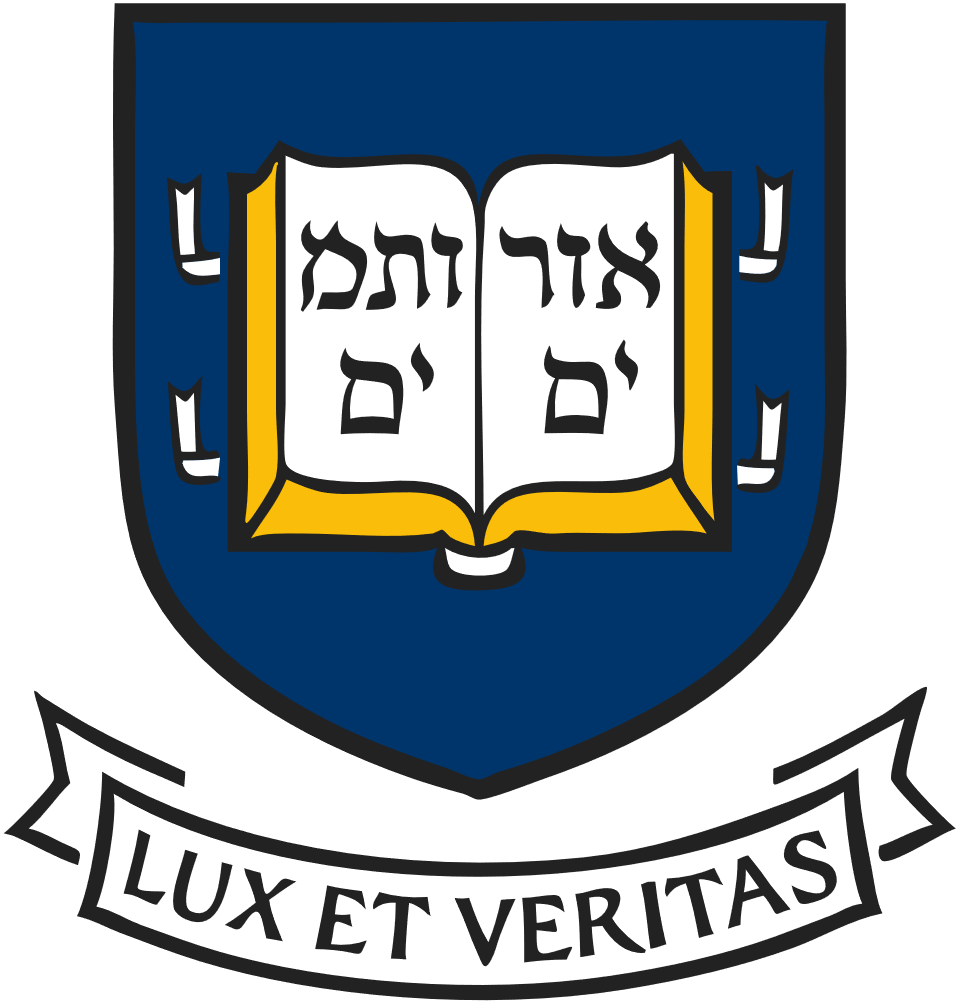}\hspace{0.2em}
	\raisebox{0.3cm}{\includegraphics[height=0.5cm]{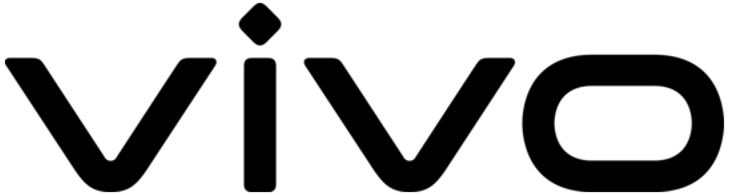}}\hspace{0.2em}
			\includegraphics[height=1.0cm]{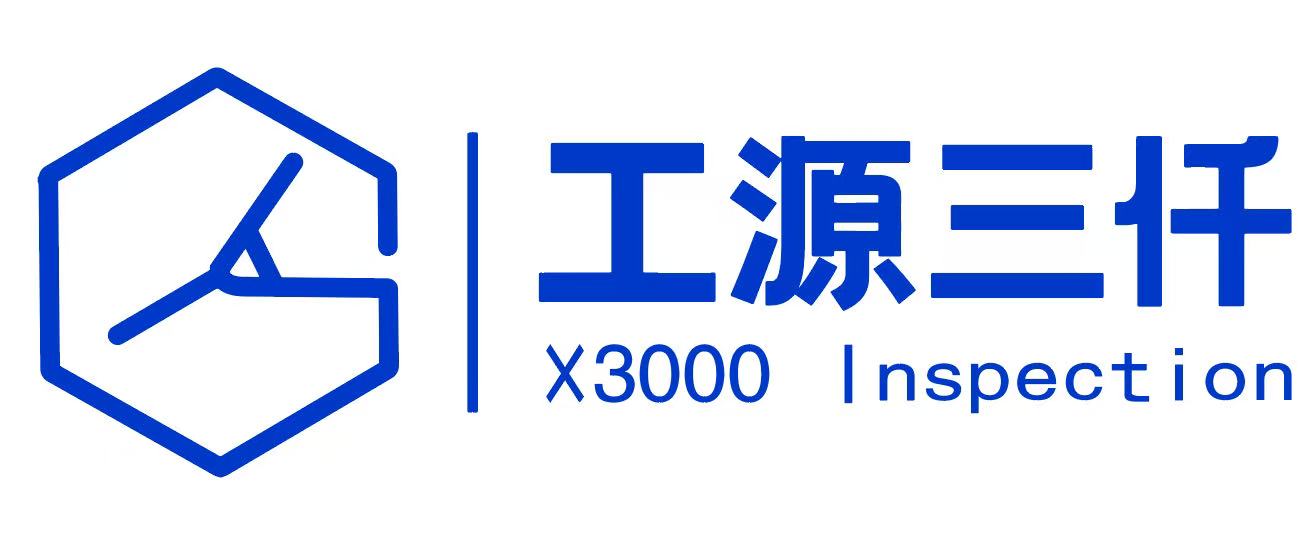}\hspace{0.1em}
		\end{minipage}
		\\[0.4em]
		\textcolor{academicblue}{\rule{\textwidth}{0.6pt}}%
	\end{minipage}%
}

\settitlepreskip{-10ex}
\settitlelogogap{3em}

\showlogointitle
\title{Seeing Through the Glare: A Multi-Source Benchmark and Ocular-Adaptive Pixel MeanFlow for Eyeglass Reflection Removal}
\shorttitle{OcuFlow}

\makeatletter

\newcommand{\manualtitlecontent}{%
	\begin{center}
	{\MA@author@font
		Tao Liu$^{1,5}$,
		Youwei Pang$^{1}$,
		Kailai Zhou$^{1}$, 
		Jiaming Zuo$^{2,5}$,
		Hanqi Liu$^{1,5}$,
		Wei Ji$^{3}$,
			\par\vspace{0.4em}
	    Peng-Tao Jiang$^{4}$,
		Xiaofeng Liu$^{3}$,
		Weisi Lin$^{1}$,
		Xiaoqi Zhao$^{1}$
		\par}
	
	\vspace{\MA@author@affiliation@skip}
	
	{\MA@affiliation@font
	$^{1}$AI4X Team, Nanyang Technological University, Singapore\par
	$^{2}$Tsinghua University, China\par
	$^{3}$Yale University, USA\par
	$^{4}$vivo BlueImage Lab, vivo Mobile Communication Co., Ltd.\par
	$^{5}$X3000 Inspection Co., Ltd.\par
	
	}
	
	\vspace{\MA@affiliation@email@skip}
	
\end{center}

	\vspace{\MA@author@abstract@preskip}
	{\centering\textcolor{abstractborder}{\rule{0.3\textwidth}{0.4pt}}\par}
	\vspace{\MA@author@abstract@postskip}

	\noindent
Eyeglass reflection removal is important across smartphone imaging, video conferencing, and other face-centric visual applications.
The task is challenging because reflections range from mild photometric contamination to severe ocular occlusion, requiring selective correction and plausible reconstruction without altering identity or natural appearance.
Existing datasets cover limited reflection conditions, constraining generalization to complex real-world scenes and systematic evaluation.
We introduce \textbf{OcuBench}, a multi-source benchmark comprising 10,280 controllable synthetic pairs, 732 real-input pseudo-pairs, and 458 independent real-world test images, supporting both paired evaluation and assessment beyond generated supervision.
We further propose \textbf{OcuFlow}, an ocular-adaptive pixel MeanFlow (pMF) framework for efficient, detail-preserving restoration.
It combines geometry-adaptive representation with one-step pMF to focus reconstruction on reflection-obscured ocular regions, together with native-resolution frequency-preserving synthesis to retain reliable observed details.
Experiments across diverse reflection conditions demonstrate that OcuFlow achieves consistent advantages in reflection removal quality, ocular fidelity, and efficiency. In a blind user study, it receives $67.32\%$ of selections, $6.2\times$ the next-best share.
The code and dataset will be released at \url{https://github.com/NTU-AI4X/OcuFlow}.
}

\renewcommand{\@abstractboxcontent}{\manualtitlecontent}

\makeatother
\begin{document}
\maketitle

\begin{figure}[h]
	\centering
	\includegraphics[width=\linewidth]{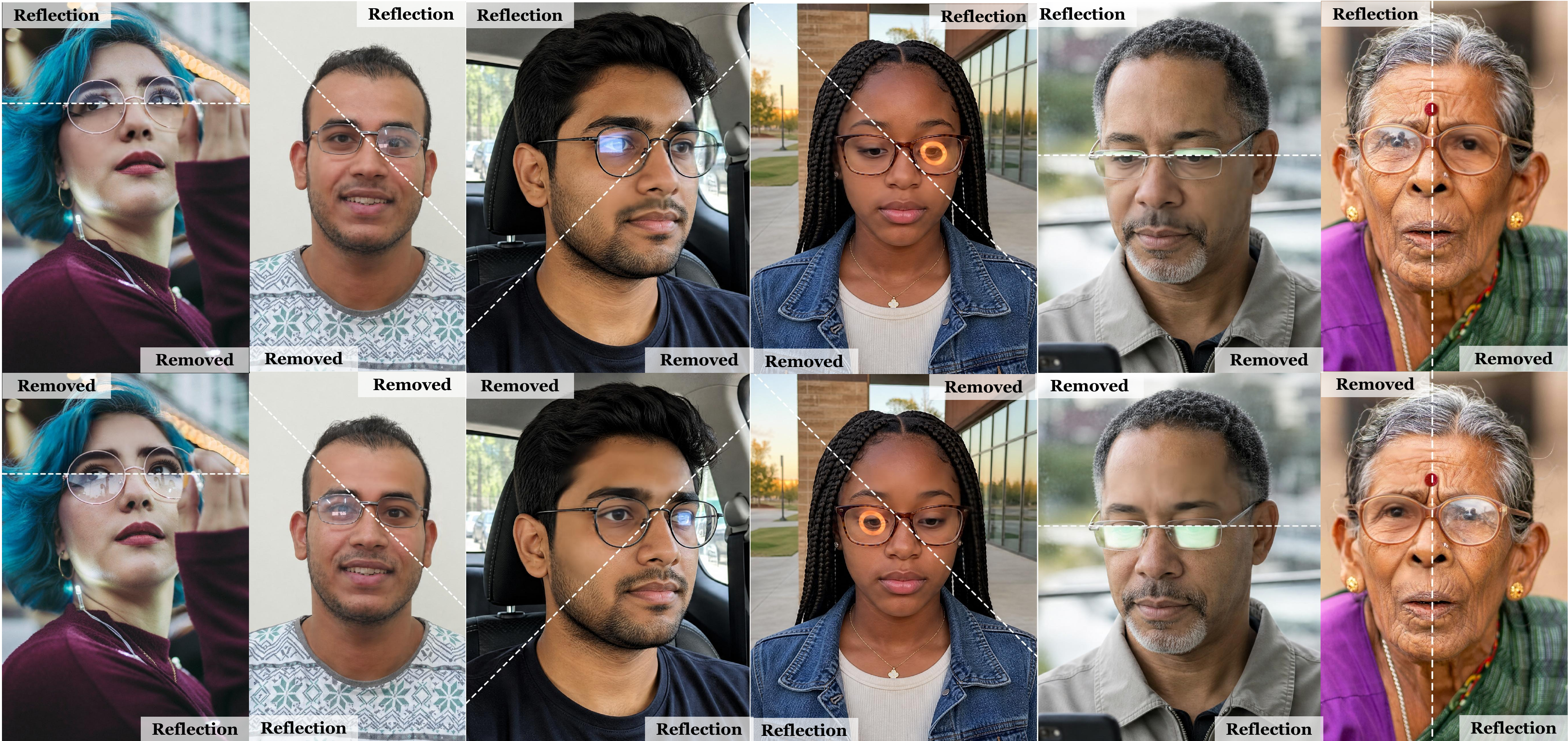}
	\caption{Eyeglass reflection removal on synthetic samples.
		Dashed lines separate input (Reflection) and restored (Removal) regions, with complementary comparisons across both rows.}
	\label{fig:teaser}
\end{figure}

\section{Introduction}
Reflections from windows or monitors on eyeglasses can obscure the eyes while leaving most of a portrait intact.
Although spatially confined, this interference can noticeably change facial appearance in photographs and video communication.
Its effects range from color shifts over visible structures to severe occlusion of the iris, eyelids, and eyelashes~\citep{zou2024eyeglass,yi2025dl2g}.
Fig.~\ref{fig:teaser} illustrates this diversity across synthetic portraits, including localized highlights, ring-shaped glare, and broad reflections covering substantial lens areas.
These patterns vary in color, intensity, and spatial extent, interacting with eyeglass geometry and head pose to obscure different amounts of ocular detail.
A single image may therefore require photometric correction in some regions, content reconstruction in others, and faithful preservation elsewhere.
Accommodating different restoration needs is essential to removing reflections without unnecessarily changing the portrait.

Despite progress in eyeglass reflection removal~\citep{sandhan2017anti,zou2024eyeglass,yi2025dl2g}, challenges remain in obtaining reliable supervision and evaluation data and balancing restoration accuracy with inference efficiency.
At the data level, limitations are threefold:
(1) head motion, blinking, and illumination changes compromise pixel-level correspondence in captured pairs~\citep{zou2024eyeglass};
(2) synthetic pairs provide controlled correspondence and degradation but cannot alone establish performance under natural reflections;
and (3) real-input pseudo-pairs retain natural reflections, but the true appearance of obscured ocular content cannot be verified from reflected inputs alone.
Therefore, reliable assessment requires complementary sources and real-world evaluation independent of generated targets.
At the model level, reconstructing obscured ocular structures requires expressive priors, while applications demand efficient inference.
Diffusion models~\citep{ho2020denoising}, DiT~\citep{DiT}, and SiT~\citep{SiT} demonstrate strong image synthesis capabilities.
These models typically rely on iterative sampling with multiple network evaluations.
MeanFlow~\citep{geng2026mean} learns average velocities over time intervals for single-evaluation generation, while pMF~\citep{lu2026onestep} combines image prediction with velocity-space learning for one-step generation without a latent autoencoder.

Although pMF provides an efficient foundation for one-step generation, applying it to eyeglass reflection removal requires aligning generative computation with spatially heterogeneous restoration demands.
Under a limited token budget, uniform tokenization allocates computation to largely unaffected regions while providing insufficient resolution for fine ocular structures.
Within the lenses, similar reflection intensities may leave different amounts of ocular structure visible, indicating that reflection strength alone cannot determine restoration requirements.
Therefore, effective reconstruction requires guidance from both local observations and ocular geometry.
In addition, compact patch representations can discard native photographic details, whereas directly preserving observed high-frequency content risks retaining reflection artifacts.

To address these challenges, we introduce \textbf{OcuBench}, a multi-source benchmark for eyeglass reflection removal.
OcuBench comprises 10,280 synthetic pairs, 732 real-input pseudo-pairs, and 458 independently collected real-world test images.
Synthetic pairs vary in subject, imaging, and reflection factors.
Real-input pseudo-pairs combine real reflected inputs with generated clean targets.
The real-world set evaluates performance beyond generated supervision, with clean references for a subset.
These sources support development and evaluation under controlled and real-world conditions.
Based on OcuBench, we develop \textbf{OcuFlow}, an efficient ocular-adaptive pMF framework for eyeglass reflection removal.
OcuFlow uses geometry-adaptive supports under a fixed token budget to concentrate resolution on ocular structures while preserving coarse context.
The supports retain position and scale information to preserve spatial correspondence.
A geometry-guided pMF integrates local reflected observations, ocular priors, and cross-scale features to restore pixel patches in one forward pass.
A native-resolution frequency-preserving renderer projects predictions to the supplied input resolution.
The renderer selectively blends predictions with observations across frequency bands to preserve photographic details and reduce unnecessary modifications.

Experiments show that OcuFlow achieves superior overall restoration quality on the held-out paired benchmark compared with existing methods.
OcuFlow runs in 56~ms per image on an RTX~4090 GPU, providing $4$--$66\times$ speedups over the compared methods under the same evaluation protocol.
Independent real-world evaluation further shows competitive no-reference performance, and OcuFlow achieves the highest user preference in the user study.
Beyond image-level restoration metrics, we evaluate downstream iris localization using a fixed detector without fine-tuning.
OcuFlow achieves the best both-eyes localization accuracy among the compared method, demonstrating that its restoration benefits extend to downstream ocular analysis.
These results demonstrate the potential of OcuFlow for efficient and detail-preserving portrait restoration.

\section{Related Work}

\textbf{Eyeglass Reflection Removal and Benchmarks.}
EwE~\citep{sandhan2017anti} overlays reflection patches on frontal portraits, while De-Glared~\citep{sadik2023glared} constructs 2,541 synthetic pairs from generated portraits.
ReyeR~\citep{zou2024eyeglass} provides filtered, color-corrected real pairs across varied conditions.
Synthetic realism and real-pair alignment remain challenges.
Methods span constrained optimization~\citep{sandhan2017anti}, supervised restoration~\citep{watanabe2021reflection,sadik2023glared}, and generative facial reconstruction~\citep{chandramouli2019blind}.
ER$^2$Net~\citep{zou2024eyeglass} fuses reflection removal and inpainting by reflection intensity, while DL2G~\citep{yi2025dl2g} combines degradation correction, local structure-aware diffusion, and global refinement.
We introduce a multi-source benchmark and efficient restoration for portrait editing and video communication.

\textbf{Spatially Adaptive Visual Representations.}
Adaptive visual representations improve efficiency through token aggregation, conditional computation, and spatially varying resolution.
TokenLearner~\citep{ryoo2021tokenlearner} aggregates dense features into fewer adaptive tokens, while A-ViT~\citep{yin2022vit} dynamically halts token processing at different network depths.
Mixed-Resolution Tokenization~\citep{ronen2023vision} constructs variable-sized patches through saliency-guided quadtree partitioning, whereas MSViT~\citep{havtorn2023msvit} uses learned gating to select patch scales with input-dependent token counts.
For eyeglass restoration, spatial allocation must resolve fine ocular structures while maintaining correspondence with the original image.
OcuFlow uses input-derived ocular geometry to optimize quadtree partitioning under a fixed token budget, concentrating resolution on critical eye structures.
Each token retains its position and scale for geometry-guided restoration and native-resolution reconstruction.

\textbf{Generative Restoration.}
Generative priors support restoration from incomplete observations.
Diffusion~\citep{ho2020denoising} and Flow Matching~\citep{lipman2023flow} typically require iterative sampling.
DifFace~\citep{yue2024difface} reconstructs faces with pretrained diffusion, while DiffBIR~\citep{lin2024diffbir} combines degradation removal with diffusion-based detail regeneration.
GFP-GAN~\citep{wang2021towards} enables single-pass restoration using pretrained facial GAN priors.
CodeFormer~\citep{zhou2022towards} predicts discrete codes with a controllable fidelity--quality trade-off.
Dereflection Any Image~\citep{hu2026dereflection} adopts one-step diffusion for reflection removal.
MeanFlow~\citep{geng2026mean} learns average velocities over time intervals for single-evaluation generation.
pMF~\citep{lu2026onestep} combines image prediction with velocity-space learning for one-step pixel-space generation.
We adapt it to conditional restoration over ocular-adaptive patches.
One-step restoration addresses \emph{efficiency}, but not \emph{what to restore and preserve}.
WaveFace~\citep{miao2024waveface} uses diffusion for low-frequency wavelet components and a feed-forward network for high-frequency recovery.
Our renderer blends predictions with native-resolution observations by frequency, selectively retaining reliable high-frequency details beyond compact patches.

\begin{table}[t]
	\centering
	\setlength{\abovecaptionskip}{3pt}
	\caption{Comparison of representative datasets for eyeglass reflection removal. ``Syn.'' denotes synthetic supervision, ``Real'' denotes real reflected inputs, and ``Real Test'' denotes an independent real-world evaluation set.}
	\label{tab:dataset_comparison}
	\setlength{\tabcolsep}{6.5pt} 
	\resizebox{\linewidth}{!}{%
		\begin{tabular}{lcccccl}
			\toprule
			Dataset & Year & Syn. & Real & Real Test & Scale & Access \\
			\midrule
			EwE~\citep{sandhan2017anti} & 2017 & \checkmark & -- & -- & 540 & Legacy (Link Unavailable) \\
			Watanabe et al.~\citep{watanabe2021reflection} & 2021 & -- & \checkmark & -- & 1,639 & No Public Link \\
			De-Glared~\citep{sadik2023glared} & 2023 & \checkmark & -- & -- & 2,541 & No Public Link \\
			ReyeR~\citep{zou2024eyeglass} & 2024 & -- & \checkmark & -- & 13,610  & No Public Link \\
			ReyeR+~\citep{yi2025dl2g} & 2025 & -- & \checkmark & -- & 14,328  & No Public Link \\
			\midrule
			\textbf{OcuBench} & 2026 & \textbf{\checkmark} & \textbf{\checkmark} & \textbf{\checkmark} & \textbf{10,280 / 732  / 458} & \textbf{Public / Verified} \\
			\bottomrule
		\end{tabular}%
	}
\end{table}

\section{OcuBench: A Multi-Source Benchmark}
\label{sec:dataset}

\begin{figure}[t]
	\centering
	\includegraphics[width=\linewidth]{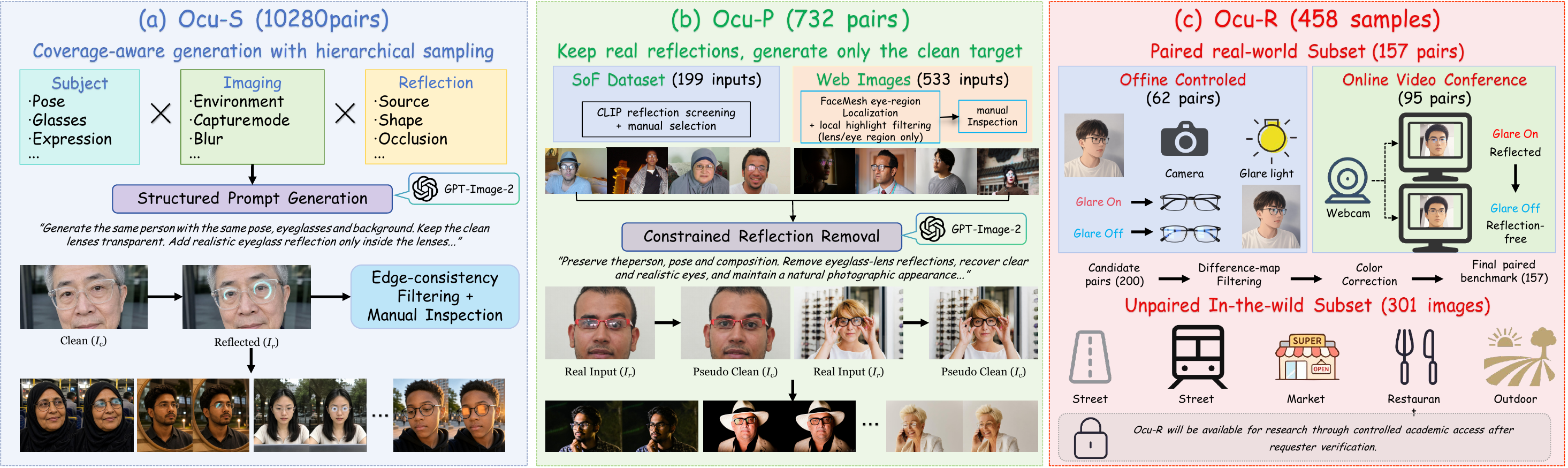}
	\caption{Construction of OcuBench. 
		Ocu-S is generated from hierarchically sampled subject, imaging, and reflection factors.
		Ocu-P retains original reflected images and generates only clean targets.
		Ocu-R combines captured reference pairs and unpaired images from unconstrained scenes.}
	\label{fig:dataset_pipeline}
\end{figure}

Tab.~\ref{tab:dataset_comparison} compares representative eyeglass reflection datasets by construction, pairing, scale, and access. 
Controlled real capture provides physically valid paired observations, while synthetic composition enables scalable and controllable variation in reflection patterns. However, large-scale real paired facial data are difficult to collect and release due to acquisition constraints and privacy concerns, whereas synthetic data alone cannot fully capture naturally occurring reflections. 
As shown in Fig.~\ref{fig:dataset_pipeline}, OcuBench therefore combines controllable synthetic pairs, real-input pseudo-pairs, and independent real-world samples, leveraging their complementary strengths for controlled degradation analysis, realistic reflection coverage, and independent real-world evaluation.
The following subsections describe data collection, generation, cleaning, and quality control, with further procedural details in \cref{app:dataset}.

\subsection{Data Collection and Generation}
\label{sec:dataset_collection}

\textbf{Controllable Synthetic Pairs (Ocu-S).}
We adopt \emph{structured hierarchical sampling} over subject factors $\mathbf{s}$ (pose, eyeglasses, expression, and gaze), imaging factors $\mathbf{i}$ (capture mode, scene mode, environment, background blur, and camera distance), and reflection factors $\mathbf{r}$ (source, shape, color, transparency, strength, and target eye occlusion), with their primary high-level values summarized in \cref{tab:synthetic_factors}:
\begin{equation}
p(\mathbf{s},\mathbf{i},\mathbf{r})
=p(\mathbf{s})\,p(\mathbf{i}\mid\mathbf{s})\,
p(\mathbf{r}\mid\mathbf{s},\mathbf{i}).
\label{eq:hierarchical_data_sampling}
\end{equation}
We sample subject factors first, then imaging factors conditioned on the subject, followed by reflection factors conditioned on both.
This hierarchy promotes compatible combinations spanning mild photometric contamination to severe ocular occlusion.
Structured prompts encode sampled configurations for GPT-Image-2~\citep{openai2026gptimage2} to generate reflected--clean candidates $(\widetilde I_r,\widetilde I_c)$, requesting consistent identity, pose, eyeglasses, background, and composition while varying only lens reflections.
These prompts are provided in \cref{app:syn_dataset}.

\textbf{Real-input Pseudo-pairs (Ocu-P).}
We collect naturally reflected images from SoF~\citep{afifi2019afif4} and Web search, with the latter yielding 11,520 initial candidates.
Following input screening, we preserve each selected reflected image and generate only its missing clean target as $\widetilde I_c=\mathcal{E}(I_r^{\mathrm{real}},\mathbf{p}_{\mathrm{remove}})$.
The operator $\mathcal{E}$ denotes GPT-Image-2 editing, and $\mathbf{p}_{\mathrm{remove}}$ requests reflection removal while preserving identity, pose, eyeglasses, composition, and photographic appearance.
The input $I_r^{\mathrm{real}}$ remains unchanged, retaining naturally observed reflection patterns.

\textbf{Independent Real-world Samples (Ocu-R).}
We acquire 200 candidate reflected--clean pairs through controlled offline photography and online video conferencing, with the reflection-inducing light enabled and disabled.
We collect 301 unpaired images from unconstrained environments, covering varied illumination, viewpoints, eyeglasses, and reflection conditions.
All reflected inputs and clean references are acquired without image generation and reserved for real-world evaluation.

\subsection{Data Cleaning and Quality Control}
\label{sec:dataset_cleaning}

\textbf{Ocu-P Input Screening.}
For the SoF dataset, CLIP-based reflection screening~\citep{radford2021learning} followed by manual inspection yields 199 images.
For the Web data, FaceMesh-guided local-highlight screening in estimated eye regions yields 1,740 candidates from 11,520 retrieved images, with manual inspection further reducing them to 533.
The resulting 732 images are used to generate pseudo-clean targets.

\textbf{Ocu-S and Ocu-P Validation.}
GPT-Image-2 generates synthetic pairs under shared scene constraints, whereas real-input editing must preserve the photograph while inferring obscured content.
Both risk \emph{pairwise structural drift} in facial or ocular geometry. 
An \emph{edge-consistency filter} is applied to detect contour changes unrelated to reflections.
Four reviewers assess realism, identity and ocular consistency, and preservation of unaffected content, accepting pairs by consensus.
We retain 10,280 samples in Ocu-S ($\mathcal{D}_{\mathrm{syn}}$) and 732 in Ocu-P ($\mathcal{D}_{\mathrm{pseudo}}$).
Pseudo-clean supervision remains approximate because obscured ocular appearance cannot be verified from reflected inputs alone.
Sampling configurations, prompt templates, and filtering criteria appear in \cref{app:dataset}.

\textbf{Ocu-R Reference-pair Processing.}
Following ReyeR~\citep{zou2024eyeglass}, we filter candidate captures with large spatial discrepancies using image differences and apply color correction to reduce capture-induced illumination shifts.
Of the 200 candidate pairs, 157 are retained, comprising 62 offline and 95 video-conferencing pairs.
Residual local misalignment remains, so the clean captures serve as visual references rather than pixel-accurate ground truth.
Together with the 301 unpaired images, these form the 458-image Ocu-R.

\begin{figure}[!t]
	\centering
	\setlength{\abovecaptionskip}{2pt}
	\includegraphics[width=\linewidth]{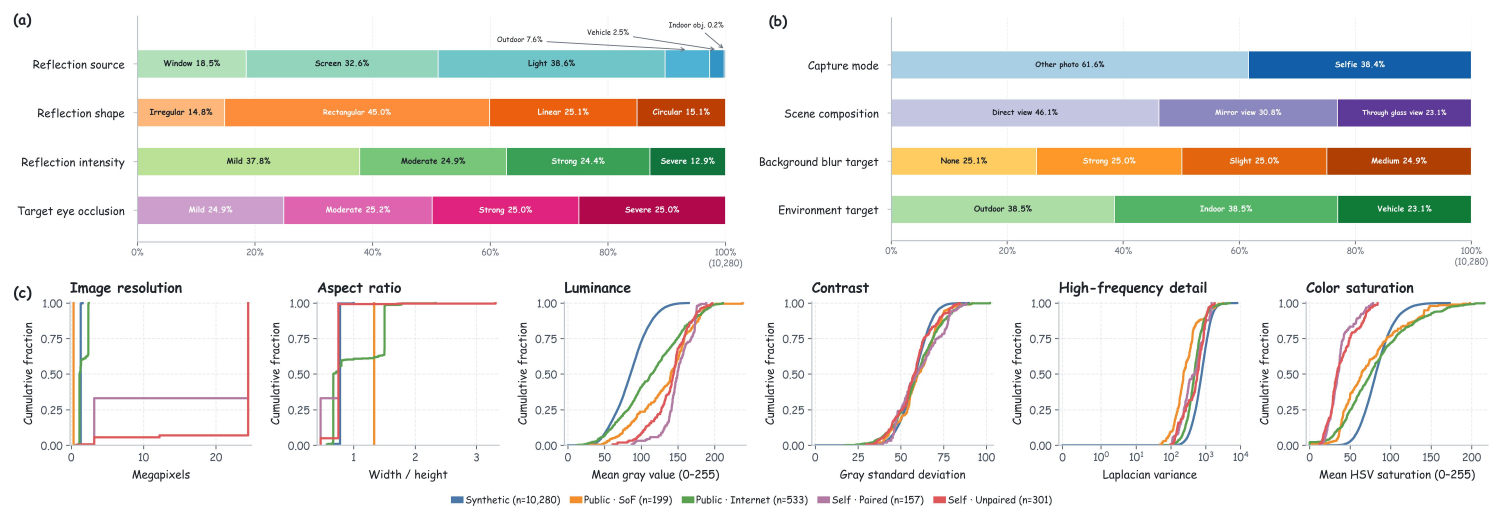}
	\caption{Dataset analysis of OcuBench.
		(a) Ocu-S reflection configurations.
		(b) Ocu-S imaging configurations.
		(c) Empirical cumulative distributions of image properties across OcuBench subsets.}
	\label{fig:combined_dataset_analysis}
\end{figure}

\subsection{Dataset Analysis}
\label{sec:dataset_analysis}

\textbf{Reflection and Occlusion Distributions.}
\cref{fig:combined_dataset_analysis}(a) summarizes requested reflection configurations in Ocu-S.
Screen and light reflections account for 32.6\% and 38.6\% of samples, respectively, while rectangular patterns are the most common shape at 45.0\%.
Target ocular occlusion is approximately balanced across mild, moderate, strong, and severe levels, each representing 24.9--25.2\% of samples.
Reflection strength follows a distinct distribution, with mild and severe cases accounting for 37.8\% and 12.9\%, respectively.
These statistics describe sampled generation settings.

\textbf{Imaging Conditions.}
\cref{fig:combined_dataset_analysis}(b) summarizes the imaging configurations.
Selfies account for 38.4\% of Ocu-S.
Scene compositions include direct, mirror, and through-glass views across indoor, outdoor, and vehicle settings.
The four target background-blur levels are approximately balanced, spanning sharp to defocused backgrounds.
These configurations vary the photographic context around reflected eye regions.

\textbf{Cross-source Image Statistics.}
\cref{fig:combined_dataset_analysis}(c) compares the data sources through empirical cumulative distributions of resolution, aspect ratio, luminance, contrast, Laplacian variance, and color saturation.
Ocu-R reaches substantially higher resolutions than Ocu-S and Ocu-P, while Ocu-S generally exhibits lower luminance.
The three sources also differ in aspect ratio, high-frequency detail, and saturation, but show more similar contrast distributions.
These differences highlight source-dependent variation and motivate separate evaluation on independently captured images.
\textbf{Dataset Splits, Evaluation, and Access.} 
We separately split Ocu-S ($\mathcal{D}_{\mathrm{syn}}$) and Ocu-P ($\mathcal{D}_{\mathrm{pseudo}}$) into 60\% training and 40\% test subsets, then merge the respective splits to form the final training and test sets. 
Held-out paired evaluation is performed on the merged test set.
All 458 Ocu-R inputs are evaluated using no-reference metrics, with 30 randomly selected for a blind user study, while the 157 clean captures support qualitative analysis.
We plan to release the synthetic and SoF components directly, release the Web-sourced component upon permission, and provide self-collected data through verified academic access (Tab.~\ref{tab:dataset_comparison}).

\section{Method}
\label{sec:method}

\begin{figure}[t]
	\centering
	\setlength{\abovecaptionskip}{2pt}
	\includegraphics[width=\linewidth]{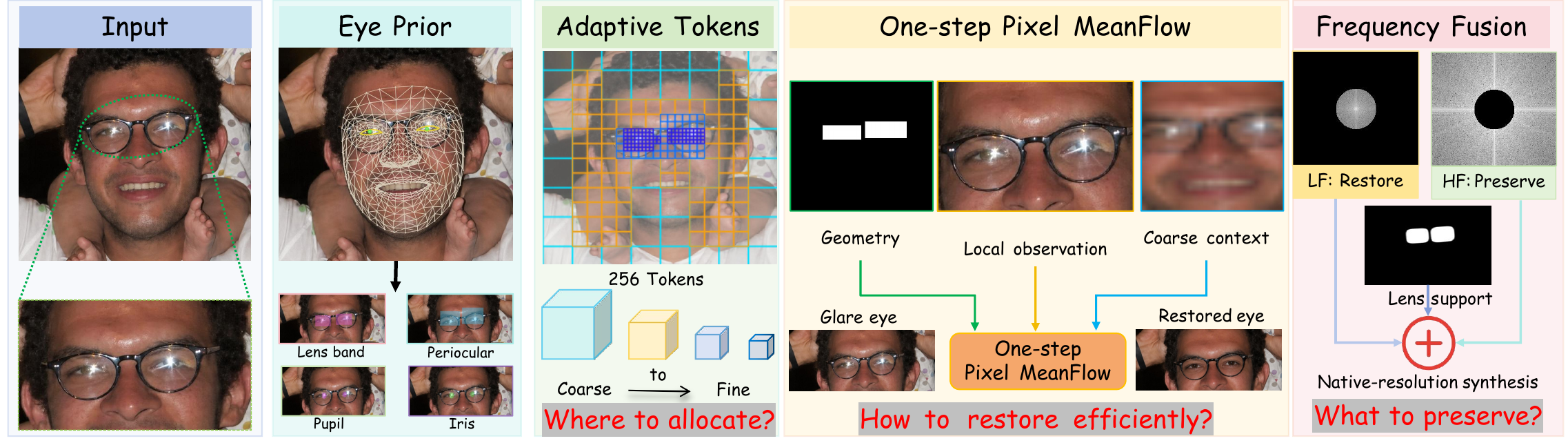}
	\caption{Overview of OcuFlow.
		Ocular priors guide allocation of 256 tokens.
		Geometry, reflected observations, and coarse context condition one-step patch restoration.
		Frequency-preserving synthesis blends predictions with native details within soft lens support.}
	\label{fig:framework}
\end{figure}

Given an arbitrary-resolution image with eyeglass reflections \(I_g\in\mathbb{R}^{H\times W\times3}\), OcuFlow estimates a reflection-free image \(I_{\mathrm{out}}\) at the same spatial resolution as the input, we denote the clean training target by $I_0$. 
As illustrated in \cref{fig:framework}, the method coordinates three decisions:
\emph{where to allocate capacity, how to restore efficiently, and what to preserve}.
Ocular geometry determines the spatial supports of a fixed number of pixel patches. 
A geometry-guided pMF restores the patches in one pass, while a frequency-preserving renderer fuses the predictions with native-resolution observations.

\subsection{Ocular-Prior-Guided Adaptive Representation}
\label{sec:adaptive_representation}

Ocular regions require finer detail than largely unaffected portrait regions. We therefore use finer patches around the eyes and coarser patches elsewhere, while keeping the sequence length fixed.
Facial landmarks from \(I_g\) define geometry-derived ocular priors on a \(256\times256\) logical domain to guide partitioning, while patches are sampled directly from the input-resolution image.
We realize this variable-resolution partition as a quadtree forest initialized by an \(8\times8\) grid of \(32\times32\) root cells.
Each leaf \(q_i=(\mathbf{c}_i,s_i)\) has center \(\mathbf{c}_i\), side length \(s_i\in\{4,8,16,32\}\), and support \(\Omega_i\).
We define the ocular-importance mass of each leaf as
\begin{equation}
M_i=\int_{\Omega_i} w(\mathbf{x})\,d\mathbf{x},
\end{equation}
where $w(\mathbf{x})\geq0$ assigns larger importance to the periocular, eye, iris, eyelid, and lens regions.
Under the fixed budget $N=256$, we select
\begin{equation}
\mathcal{Q}^{*}
=
\arg\min_{\substack{\mathcal{Q}\in\mathfrak{T}\\|\mathcal{Q}|=N}}
\sum_{q_i\in\mathcal{Q}}\rho(s_i)M_i,
\label{eq:grid_objective}
\end{equation}
where $\mathfrak{T}$ denotes the set of admissible quadtree leaf partitions and $\rho(s_i)$ is a scale penalty that increases with leaf size.
Thus, high-importance ocular regions are represented by smaller leaves, while largely unaffected regions retain coarser supports.
The resulting fixed-budget adaptive representation is used by the one-step restoration model described next. Implementation details are in \cref{app:adaptive_representation}.

\subsection{Geometry-Guided One-Step Pixel MeanFlow}
\label{sec:pmf}

The adaptive representation determines where the fixed token budget is concentrated; we next use these geometry-aware tokens to reconstruct clean ocular appearance from the reflected observation.
We combine reflected observations with ocular geometry and multi-scale context into a conditioned representation $\mathbf{h}$, with leaf-wise ocular priors supplied separately to the backbone.

\textbf{Observation-anchored Source.}
We construct $\widetilde{\mathbf{g}}$ by adding small ocular-prior-weighted Gaussian perturbations to the reflected adaptive patches $\mathbf{g}$.
This keeps the source close to the observation while emphasizing geometrically important ocular regions.
The modulation rule is given in \cref{app:pmf}.

\textbf{One-step Pixel MeanFlow.}
Following pMF~\citep{lu2026onestep}, we define a linear path from the clean adaptive patches $\mathbf{x}_0$ to the observation-anchored source:
\begin{equation}
\mathbf{z}_t
=
(1-t)\mathbf{x}_0+t\widetilde{\mathbf{g}},
\qquad t\in[0,1].
\label{eq:geometry_aware_source}
\end{equation}
The predictor $\widehat{\mathbf{x}}_{\theta}(\mathbf{z}_t,\Delta;\mathbf{h})$ estimates the clean endpoint conditioned on the current path state, interval length $\Delta$, and geometry-aware context $\mathbf{h}$.
At inference, setting $t=1$ and $\Delta=1$ yields the restored patches in one network evaluation:
\begin{equation}
\widehat{\mathbf{x}}
=
\widehat{\mathbf{x}}_{\bar{\theta}}
(\widetilde{\mathbf{g}},1;\mathbf{h}),
\label{eq:one_step_inference}
\end{equation}
where $\bar{\theta}$ denotes exponential-moving-average parameters.
The source modulation, MeanFlow parameterization, conditioning architecture, and time sampling are detailed in \cref{app:pmf}.

\subsection{Native-Resolution Frequency-Preserving Synthesis}
\label{sec:renderer}

Canonical patches provide a compact representation, but resampling can discard photographic details.
Conversely, retaining input detail indiscriminately can preserve reflection artifacts.
We therefore combine predicted and observed content through spatially varying frequency mixing.

\textbf{Absolute and Residual Reconstruction.}
For the fixed partition \(\mathcal{Q}^{*}\), let \(\mathcal{R}\) denote linear projection from canonical patches to the input-resolution canvas.
It resamples patches onto their native supports and combines them using normalized raised-cosine windows whose weights sum to one.
We construct two candidates:
\begin{equation}
I_{\mathrm{abs}}
=\mathcal{R}(\widehat{\mathbf{x}}),
\quad
I_{\mathrm{res}}
=I_g+\mathcal{R}(\widehat{\mathbf{x}}-\mathbf{g}).
\label{eq:render_candidates}
\end{equation}
The absolute candidate reconstructs the image from predicted patches, whereas the residual candidate applies their estimated correction to the native input.
By linearity, $I_{\mathrm{res}}-I_{\mathrm{abs}}=I_g-\mathcal{R}(\mathbf{g})$, which represents the input information not preserved during patch sampling and projection.
It may contain both useful photographic details and reflection artifacts, which motivates selective fusion.

\textbf{Frequency-dependent Fusion.}
Let $\mathcal{L}$ be a low-pass operator,
$\mathcal{H}(I)=I-\mathcal{L}(I)$ the high-frequency component,
and $A_{\mathrm{lens}}\in[0,1]^{H\times W}$ a soft lens-support map. 
Spatial maps $\alpha$ and $\beta$ provide location-dependent control over the mixing of low- and high-frequency components:
\begin{gather}
L_{\mathrm{out}} =
\alpha\odot\mathcal{L}(I_{\mathrm{abs}})
+(1-\alpha)\odot\mathcal{L}(I_{\mathrm{res}}),\\
H_{\mathrm{out}} =
(1-A_{\mathrm{lens}})\odot\mathcal{H}(I_g) 
+A_{\mathrm{lens}}\odot
\Big[
\beta\odot\mathcal{H}(I_g)
+(1-\beta)\odot\mathcal{H}(I_{\mathrm{abs}})
\Big],\\
I_{\mathrm{render}} =L_{\mathrm{out}}+H_{\mathrm{out}}.
\label{eq:frequency_render}
\end{gather}
All maps are broadcast over RGB channels.
$\alpha$ balances absolute and residual reconstructions, while larger $\beta$ retains more observed high-frequency detail.
A lightweight gate refines $\alpha$ and $\beta$ through lens-masked residuals, with $\alpha=0$ and $\beta=1$ where $A_{\mathrm{lens}}=0$.
Final compositing gives
\begin{equation}
I_{\mathrm{out}}
=
(1-A_{\mathrm{lens}})\odot I_g
+
A_{\mathrm{lens}}\odot I_{\mathrm{render}}.
\label{eq:final_composite}
\end{equation}
Details of native projection, lens-support construction, and spatial gate construction are in \cref{app:renderer}.

\section{Experiments}
\label{sec:experiments}

\subsection{Metrics and Implementation Details}
\label{sec:exp_setup}

Paired evaluation uses Mask-PSNR/SSIM, Mixed-PSNR/SSIM, Eye-ROI LPIPS, and Eye-NME/Iris-NME. Eyeglass masks are initialized with Grounded-SAM-2~\citep{ren2024grounded} and manually corrected, while NME is computed from MediaPipe landmarks~\citep{lugaresi2019mediapipe}. Evaluation on Ocu-R comprises NIQE~\citep{mittal2012making}, CLIP-IQA+~\citep{wang2023exploring}, and a blind user study. All methods use $1008\times1296$ inputs for paired inference and efficiency evaluation on the same RTX~4090. Runtime is measured as the steady-state average wall-clock processing time per image under sustained inference, with each method evaluated using the same throughput-oriented protocol. Paired metrics are computed at the original paired resolutions, while Ocu-R is evaluated at its original resolution. Metric definitions, baseline configurations, training objectives, and efficiency measurement protocols are provided in \cref{app:implementation}.

\begin{table}[t]
	\centering
	\setlength{\abovecaptionskip}{3pt}
	\caption{Quantitative comparison on the held-out paired OcuBench test set.
		Best and second-best results are \textbf{bold}
		and \underline{underlined}, respectively.}
	\label{tab:comparison_all}
	
	\setlength{\aboverulesep}{1pt}
	\setlength{\belowrulesep}{1pt}
	
	\begin{adjustbox}{max width=\linewidth}
		\begin{tabular}{l*{7}{c}rr}
			\toprule
			& \multicolumn{5}{c}{Restoration Quality}
			& \multicolumn{2}{c}{Ocular Geometry}
			& \multicolumn{2}{c}{Efficiency} \\
			\cmidrule(lr){2-6}
			\cmidrule(lr){7-8}
			\cmidrule(l){9-10}
			Method
			& \shortstack{Mask\\PSNR $\uparrow$}
			& \shortstack{Mixed\\PSNR $\uparrow$}
			& \shortstack{Mask\\SSIM $\uparrow$}
			& \shortstack{Mixed\\SSIM $\uparrow$}
			& \shortstack{Eye-ROI\\LPIPS $\downarrow$}
			& \shortstack{Eye-NME\\(\%) $\downarrow$}
			& \shortstack{Iris-NME\\(\%) $\downarrow$}
			& \shortstack{Memory\\(GB) $\downarrow$}
			& \shortstack{Runtime\\(ms/image) $\downarrow$} \\
			\midrule
			
			Stable Diffusion 1.5~\citep{rombach2022high}
			& 19.1926 & 21.2009 & 0.5446 & 0.6375
			& 0.1187 & 1.7369 & 2.1161 & 10.31 & 2709 \\
			
			DSRNet~\citep{hu2023single}
			& 20.1502 & 21.8836 & 0.6256 & 0.6845
			& 0.1831 & 2.0755 & 2.4358 & 6.55 & 3430 \\
			
			ER$^2$Net~\citep{zou2024eyeglass}
			& 19.9652 & 22.1108 & 0.5761 & 0.6742
			& 0.1520 & 1.7455 & 2.1130 & 18.20
			& \underline{240} \\

			Stable Diffusion 3.5~\citep{esser2024scaling}
			& 18.4196 & 20.6247 & 0.5249 & 0.6346
			& 0.2029 & 2.0579 & 2.5207 & 20.06 & 1454 \\
			
			DSIT~\citep{hu2024single}
			& \underline{21.5400} & \underline{23.2821}
			& \underline{0.6478} & \textbf{0.7078}
			& 0.1461 & \underline{1.5363} & \underline{1.6998}
			& 2.96 & 3094 \\
			
			RDNet~\citep{zhao2025reversible}
			& 21.0688 & 22.8433 & 0.6300 & 0.6931
			& \underline{0.1161} & 1.5624 & 1.7536
			& \underline{1.49} & 2707 \\
			
			DL2G~\citep{yi2025dl2g}
			& 19.8145 & 21.7636 & 0.5747 & 0.6599
			& 0.1788 & 1.7772 & 2.1993 & 1.62 & 3747 \\
			
			ReflexSplit~\citep{Lee_2026_CVPR}
			& 20.7927 & 22.4125 & 0.6275 & 0.6838
			& 0.1350 & 1.7792 & 1.9592 & 13.15 & 1840 \\

			\rowcolor{oursgreen}
			\textbf{Ours}
			& \textbf{22.1054} & \textbf{23.4710}
			& \textbf{0.6633} & \underline{0.7074}
			& \textbf{0.1035} & \textbf{1.4030} & \textbf{1.5863}
			& \textbf{1.04} & \textbf{56} \\
			\bottomrule
		\end{tabular}
	\end{adjustbox}
	
\end{table}

\begin{figure}[t]
	\centering
	\setlength{\abovecaptionskip}{3pt}
	\includegraphics[width=\linewidth]{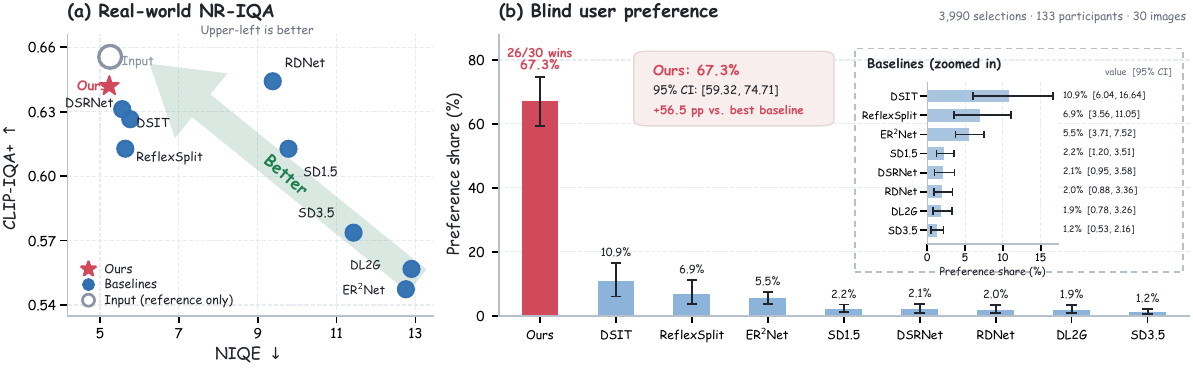}
	\caption{Real-world evaluation on OcuBench.
		\textbf{(a)} NR-IQA results on 458 real-world images, where the arrow indicates the direction of better performance.
		\textbf{(b)} Blind user preference comparison; error bars denote 95\% cluster-bootstrap confidence intervals.}
	\label{fig:realworld_evaluation}
\end{figure}

\subsection{Evaluation}
\label{sec:sota}

\textbf{Quantitative Results.} 
In Tab.~\ref{tab:comparison_all}, OcuFlow shows clear advantages in restoration accuracy and efficiency.
It ranks first on six of the seven restoration and ocular geometry metrics and achieves a $4.3\times$ speedup in steady-state processing over ER$^2$Net, the second-fastest method under the same evaluation protocol.
On Ocu-R (\cref{fig:realworld_evaluation}(a)), OcuFlow achieves the lowest NIQE (5.2369) and second-highest CLIP-IQA+ (0.6419) among restoration methods.
Since the reflected input itself obtains a higher CLIP-IQA+ score (0.6556), we treat no-reference metrics as auxiliary indicators.
In the blind user study (\cref{fig:realworld_evaluation}(b)), OcuFlow receives 67.32\% of valid selections and ranks first on 26 of the 30 images. The 95\% cluster-bootstrap confidence interval for its preference share is $[59.32\%,74.71\%]$.
Beyond image-quality evaluation, downstream iris localization further shows that reflection removal improves the reliability of ocular analysis, with OcuFlow achieving strong bilateral iris-localization performance.
Additional analyses and downstream task results are provided in \cref{app:additional_qutitity_results,app:downstream_iris}.

\begin{figure}[t]
	\centering
	\setlength{\abovecaptionskip}{3pt}
	\includegraphics[width=\linewidth]{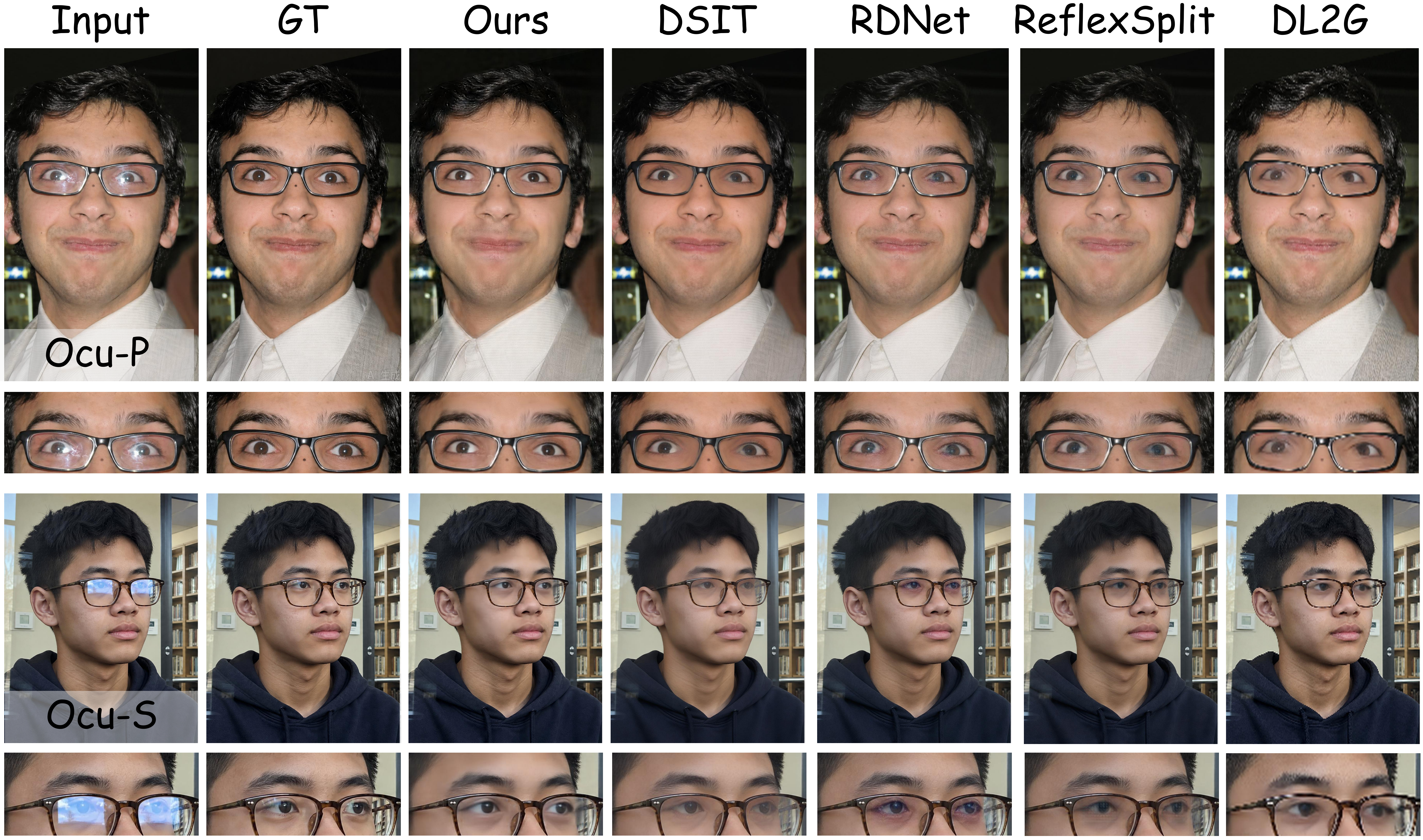}
	\caption{Qualitative comparison.
		Close-ups highlight reflection removal, ocular reconstruction, and detail preservation.}
	\label{fig:qualitative_comparison}
\end{figure}

\textbf{Qualitative Results.}
\cref{fig:qualitative_comparison} compares results under localized glare and broad blue reflections.
RDNet and ReflexSplit retain lens haze in the first example.
In the second example, DSIT and ReflexSplit produce softer, lower-contrast eyes, while RDNet retains a blue-purple cast.
DL2G introduces irregular details around the rims and eyelids.
OcuFlow suppresses dominant reflections in both cases, recovering clearer iris and eyelid contours closer to the references while preserving reliable facial details.
Additional results appear in \cref{app:additional_qualitative_results}.

\subsection{Ablation Study}

We evaluate the three components of OcuFlow in Tab.~\ref{tab:ablation_core} and the effect of training data composition in Tab.~\ref{tab:data_supervision}. See \cref{app:additional_ablation} for more results.

\begin{table}
	\centering
	\setlength{\abovecaptionskip}{3pt}
	\caption{Ablation studies of OcuFlow. 
		Out-PSNR and Out-HF-PSNR measure RGB and high-frequency agreement outside the evaluation mask.
		Detailed definitions are provided in \cref{app:evaluation}.}
	\setlength{\aboverulesep}{1pt}
	\setlength{\belowrulesep}{1pt}
	
	\begin{adjustbox}{max width=\linewidth}
		\begin{tabular}{l*{7}{c}r}
			\toprule
			Setting
			& \shortstack{Mask\\PSNR $\uparrow$}
			& \shortstack{Mask\\SSIM $\uparrow$}
			& \shortstack{Eye-ROI\\LPIPS $\downarrow$}
			& \shortstack{Eye-NME\\(\%) $\downarrow$}
			& \shortstack{Iris-NME\\(\%) $\downarrow$}
			& \shortstack{Out\\PSNR $\uparrow$}
			& \shortstack{Out-HF\\PSNR $\uparrow$}
			& \shortstack{Time\\(ms) $\downarrow$} \\
			\midrule
			
			\multicolumn{9}{c}{\emph{(a) Where to allocate capacity}} \\
			\addlinespace[1pt]
			Uniform Grid
			& 20.8082 & 0.6171 & 0.1811
			& 2.0670 & 2.5891 & 26.4132 & 31.2027
			& \textbf{49} \\
			w/o Geometry Embedding
			& 21.9823 & 0.6586 & 0.1129
			& 1.4367 & 1.6297 & 26.4076 & 31.2065
			& +8 \\ 
			\rowcolor{oursgreen}
			\textbf{Ocular-Adaptive Grid (Ours)}
			& \textbf{22.1054} & \textbf{0.6633} & \textbf{0.1035}
			& \textbf{1.4030} & \textbf{1.5863}
			& \textbf{26.4147} & \textbf{31.2118}
			& +7 \\ 
			\midrule
			
			\multicolumn{9}{c}{\emph{(b) How to restore efficiently}} \\
			\addlinespace[1pt]
			w/o Geometry-Aware Source
			& 21.9896 & 0.6592 & 0.1127
			& 1.4449 & 1.6336 & 26.4040 & 31.2059
			& \textbf{48} \\
			w/o Coarse-to-Fine
			& 21.9891 & 0.6591 & 0.1133
			& 1.4384 & 1.6250 & 26.4079 & 31.2076
			& +10 \\ 
			2-NFE
			& 21.4412 & 0.6329 & \textbf{0.0826}
			& 1.4695 & 1.7152 & 26.3898 & 31.1961
			& +12 \\ 
			\rowcolor{oursgreen}
			\textbf{1-NFE (Ours)}
			& \textbf{22.1054} & \textbf{0.6633} & 0.1035
			& \textbf{1.4030} & \textbf{1.5863}
			& \textbf{26.4147} & \textbf{31.2118}
			& +8 \\ 
			\midrule
			
			\multicolumn{9}{c}{\emph{(c) What to preserve}} \\
			\addlinespace[1pt]
			Direct PMF
			& 21.2437 & 0.6287 & 0.1107
			& 1.4352 & 1.5988 & 25.9394 & 30.8842
			& 59 \\
			Residual Rendering
			& 21.7838 & 0.6461 & 0.1076
			& 1.4389 & 1.6252 & 26.2933 & 31.0292
			& $<$3 \\ 
			\rowcolor{oursgreen}
			\textbf{Frequency Renderer (Ours)}
			& \textbf{22.1054} & \textbf{0.6633} & \textbf{0.1035}
			& \textbf{1.4030} & \textbf{1.5863}
			& \textbf{26.4147} & \textbf{31.2118}
			& $<$3 \\ 
			\bottomrule
		\end{tabular}
	\end{adjustbox}
	\label{tab:ablation_core}
\end{table}

\textbf{Where to Allocate Capacity.}
Tab.~\ref{tab:ablation_core}(a) shows that ocular-adaptive allocation improves masked reconstruction, Eye-ROI LPIPS, and both ocular NME measures over a uniform grid. Out-PSNR and Out-HF-PSNR remain nearly unchanged, indicating that the gains primarily concern ocular restoration. Removing geometry embedding reduces reconstruction and ocular fidelity, supporting explicit encoding of token position and scale. The uniform grid remains faster, so adaptive allocation trades additional processing time for improved restoration quality.

\textbf{How to Restore Efficiently.}
Removing either the geometry-aware source or coarse-to-fine conditioning worsens masked reconstruction and all eye-specific metrics in Tab.~\ref{tab:ablation_core}(b). These results support combining local observations with geometric and contextual guidance. A single network function evaluation (1-NFE) achieves better masked reconstruction and ocular geometry than two evaluations (2-NFE), while requiring less runtime. Although 2-NFE achieves lower Eye-ROI LPIPS, its additional computation does not consistently improve restoration. The 1-NFE setting therefore provides a favorable balance between reconstruction accuracy, ocular fidelity, and efficiency.

\textbf{What to Preserve.}
Residual rendering improves masked reconstruction and outside-mask agreement over Direct PMF but slightly worsens ocular geometry. The frequency-preserving renderer performs best across this block, with higher Out-PSNR and Out-HF-PSNR indicating closer outside-mask agreement, including in high-frequency content. These results support selective frequency fusion for balancing restoration and preservation.

\begin{table}
	\centering
	\setlength{\abovecaptionskip}{3pt}
	\caption{Ablation studies of data scale and composition.}
	\label{tab:multi_source}
	
	\setlength{\aboverulesep}{1pt}
	\setlength{\belowrulesep}{1pt}
	
	\begin{adjustbox}{max width=\linewidth}
		\begin{tabular}{c*{8}{c}}
			\toprule
			& &
			\multicolumn{6}{c}{Paired Benchmark (Ocu-S$_\text{test}$ + Ocu-P$_\text{test}$)} &
			Real-World (Ocu-R) \\
			\cmidrule(lr){3-8}
			\cmidrule(l){9-9}
			\shortstack{Ocu-S$_\text{train}$\\used (\%)}
			& \shortstack{Ocu-P$_\text{train}$}
			& \shortstack{Mask\\PSNR $\uparrow$}
			& \shortstack{Mixed\\PSNR $\uparrow$}
			& \shortstack{Mask\\SSIM $\uparrow$}
			& \shortstack{Mixed\\SSIM $\uparrow$}
			& \shortstack{Eye-NME\\(\%) $\downarrow$}
			& \shortstack{Iris-NME\\(\%) $\downarrow$}
			& \shortstack{CLIP-IQA+\\$\uparrow$} \\
			\midrule
			
			10 & --
			& 21.1233 & 22.7431 & 0.6375 & 0.6944
			& 1.6321 & 1.9898 & 0.6410 \\
			
			30 & --
			& 21.6538 & 23.1349 & 0.6546 & 0.7030
			& 1.4955 & 1.7323 & 0.6369 \\
			
			70 & --
			& 21.9224 & 23.3336 & 0.6597 & 0.7055
			& 1.4264 & 1.6076 & 0.6313 \\
			
			100 & --
			& 21.9834 & 23.3767 & 0.6617 & 0.7065
			& 1.4119 & \textbf{1.5842} & 0.6356 \\
			
			\rowcolor{oursgreen}
			100 & \checkmark
			& \textbf{22.1054} & \textbf{23.4710}
			& \textbf{0.6633} & \textbf{0.7074}
			& \textbf{1.4030} & 1.5863 & \textbf{0.6419} \\
			\bottomrule
		\end{tabular}
	\end{adjustbox}
	\label{tab:data_supervision}
\end{table}

\textbf{Effect of Multi-source Supervision.}
\cref{tab:data_supervision} shows that increasing the synthetic training fraction consistently improves paired reconstruction and ocular geometry. 
However, real-world CLIP-IQA+ does not improve monotonically with synthetic data scale.
Adding real-input pseudo-pairs (\textit{i.e.}, Ocu-P$_\text{train}$) further improves the paired
PSNR/SSIM metrics and Eye-NME, and yields the highest CLIP-IQA+ on the independent real-world set.
Iris-NME increases slightly relative to the 100\% Ocu-S-only setting.
The combined supervision thus provides broader benefits across paired and real-world evaluation, although these gains do not extend to every metric.

\section{Conclusion}

We present OcuBench and OcuFlow, a multi-source benchmark and an ocular-adaptive pMF framework for eyeglass reflection removal.
OcuBench combines synthetic pairs, real-input pseudo-pairs, and independent real-world evaluation beyond generated targets.
OcuFlow aligns a fixed token budget with ocular geometry and couples one-step restoration with frequency-preserving synthesis to reconstruct obscured eyes while retaining reliable native details. 
Experiments demonstrate improved reflection removal and ocular-structure reconstruction, leading to more accurate downstream iris localization, while achieving favorable human perceptual preferences on independent real-world images at substantially lower inference cost. 
Our findings suggest that adapting representation to local needs and selectively preserving observations support effective one-step portrait restoration.
Future work could extend this approach to temporally consistent video restoration.

\clearpage

\FloatBarrier
\bibliography{main}

@article{zou2024eyeglass,
  title={Eyeglass reflection removal with joint learning of reflection elimination and content inpainting},
  author={Zou, Wentao and Lu, Xiao and Yi, Zhilv and Zhang, Ling and Fu, Gang and Li, Ping and Xiao, Chunxia},
  journal={IEEE TCSVT},
  volume={34},
  number={10},
  pages={10266--10280},
  year={2024}
}

@inproceedings{yi2025dl2g,
  title={DL2G: Degradation-guided Local-to-Global Restoration for Eyeglass Reflection Removal},
  author={Yi, Zhilv and Lu, Xiao and Ding, Hong and Hu, Jingbo and Jiang, Zhi and Xiao, Chunxia},
  booktitle={CVPR},
  pages={16061--16070},
  year={2025}
}

@inproceedings{hu2023single,
  title={Single image reflection separation via component synergy},
  author={Hu, Qiming and Guo, Xiaojie},
  booktitle={ICCV},
  pages={13092--13101},
  year={2023}
}

@inproceedings{zhao2025reversible,
  title={Reversible decoupling network for single image reflection removal},
  author={Zhao, Hao and Li, Mingjia and Hu, Qiming and Guo, Xiaojie},
  booktitle={CVPR},
  pages={26430--26439},
  year={2025}
}

@inproceedings{sandhan2017anti,
  title={Anti-glare: Tightly constrained optimization for eyeglass reflection removal},
  author={Sandhan, Tushar and Young Choi, Jin},
  booktitle={CVPR},
  pages={1241--1250},
  year={2017}
}

@inproceedings{watanabe2021reflection,
  title={Reflection removal on eyeglasses using deep learning},
  author={Watanabe, Sota and Hasegawa, Makoto},
  booktitle={International Technical Conference on Circuits/Systems, Computers and Communications},
  pages={1--4},
  year={2021}
}

@inproceedings{sadik2023glared,
  title={De-Glared: Eyeglasses Glare and Reflection Removal Using Deep Neural Networks},
  author={Sadik-E-Tawheed, Md and Ahmed, Tasnia and Bhuiyan, Masud and Hossain, Abir and Shatabda, Swakkhar},
  booktitle={International Conference on Computer and Information Technology},
  pages={1--6},
  year={2023}
}

@inproceedings{chandramouli2019blind,
  title={Blind single image reflection suppression for face images using deep generative priors},
  author={Chandramouli, Paramanand and Gandikota, Kanchana Vaishnavi},
  booktitle={ICCV Workshops},
  pages={3315--3323},
  year={2019},
}

@inproceedings{ryoo2021tokenlearner,
  title={Tokenlearner: Adaptive space-time tokenization for videos},
  author={Ryoo, Michael and Piergiovanni, AJ and Arnab, Anurag and Dehghani, Mostafa and Angelova, Anelia},
  booktitle={NeurIPS},
  pages={12786--12797},
  year={2021}
}

@inproceedings{yin2022vit,
  title={A-vit: Adaptive tokens for efficient vision transformer},
  author={Yin, Hongxu and Vahdat, Arash and Alvarez, Jose M and Mallya, Arun and Kautz, Jan and Molchanov, Pavlo},
  booktitle={CVPR},
  pages={10799--10808},
  year={2022}
}

@inproceedings{ronen2023vision,
  title={Vision transformers with mixed-resolution tokenization},
  author={Ronen, Tomer and Levy, Omer and Golbert, Avram},
  booktitle={CVPR Workshops},
  pages={4613--4622},
  year={2023}
}

@inproceedings{havtorn2023msvit,
  title={Msvit: Dynamic mixed-scale tokenization for vision transformers},
  author={Havtorn, Jakob Drachmann and Royer, Am{\'e}lie and Blankevoort, Tijmen and Bejnordi, Babak Ehteshami},
  booktitle={ICCV Workshops},
  pages={838--848},
  year={2023}
}

@inproceedings{wang2021towards,
  title={Towards real-world blind face restoration with generative facial prior},
  author={Wang, Xintao and Li, Yu and Zhang, Honglun and Shan, Ying},
  booktitle={CVPR},
  pages={9164--9174},
  year={2021}
}

@inproceedings{zhou2022towards,
  title={Towards robust blind face restoration with codebook lookup transformer},
  author={Zhou, Shangchen and Chan, Kelvin and Li, Chongyi and Loy, Chen Change},
  booktitle={NeurIPS},
  pages={30599--30611},
  year={2022}
}

@inproceedings{ho2020denoising,
  title={Denoising diffusion probabilistic models},
  author={Ho, Jonathan and Jain, Ajay and Abbeel, Pieter},
  booktitle={NeurIPS},
  pages={6840--6851},
  year={2020}
}

@inproceedings{lipman2023flow,
  title={Flow Matching for Generative Modeling},
  author={Yaron Lipman and Ricky T. Q. Chen and Heli Ben-Hamu and Maximilian Nickel and Matthew Le},
  booktitle={ICLR},
  year={2023}
}

@inproceedings{geng2026mean,
  title={Mean flows for one-step generative modeling},
  author={Geng, Zhengyang and Deng, Mingyang and Bai, Xingjian and Kolter, Zico and He, Kaiming},
  booktitle={NeurIPS},
  pages={75460--75482},
  year={2026}
}

@inproceedings{lu2026onestep,
  title={One-step Latent-free Image Generation with Pixel Mean Flows},
  author={Yiyang Lu and Susie Lu and Qiao Sun and Hanhong Zhao and Zhicheng Jiang and Xianbang Wang and Tianhong Li and Zhengyang Geng and Kaiming He},
  booktitle={ICML},
  year={2026}
}

@inproceedings{miao2024waveface,
  title={Waveface: Authentic face restoration with efficient frequency recovery},
  author={Miao, Yunqi and Deng, Jiankang and Han, Jungong},
  booktitle={CVPR},
  pages={6583--6592},
  year={2024}
}

@inproceedings{rombach2022high,
  title={High-resolution image synthesis with latent diffusion models},
  author={Rombach, Robin and Blattmann, Andreas and Lorenz, Dominik and Esser, Patrick and Ommer, Bj{\"o}rn},
  booktitle={CVPR},
  pages={10674--10685},
  year={2022}
}

@inproceedings{esser2024scaling,
  title={Scaling rectified flow transformers for high-resolution image synthesis},
  author={Esser, Patrick and Kulal, Sumith and Blattmann, Andreas and Entezari, Rahim and M{\"u}ller, Jonas and Saini, Harry and Levi, Yam and Lorenz, Dominik and Sauer, Axel and Boesel, Frederic and others},
  booktitle={ICML},
  year={2024}
}

@inproceedings{hu2024single,
  title={Single image reflection separation via dual-stream interactive transformers},
  author={Hu, Qiming and Wang, Hainuo and Guo, Xiaojie},
  booktitle={NeurIPS},
  pages={55228--55248},
  year={2024}
}

@inproceedings{Lee_2026_CVPR,
    author    = {Lee, Chia-Ming and Lin, Yu-Fan and Jiang, Jin-Hui and Hsiao, Yu-Jou and Hsu, Chih-Chung and Liu, Yu-Lun},
    title     = {ReflexSplit: Single Image Reflection Separation via Layer Fusion-Separation},
    booktitle = {CVPR},
    pages     = {1300-1309},
    year      = {2026}
}

@misc{ren2024grounded,
  title={Grounded sam 2: Ground and track anything in videos},
  author={Ren, Tianhe and Liu, Shilong and Zeng, Ailing and Lin, Jing and Li, Kunchang and Cao, He and Chen, Jiayu and Huang, Xinyu and Chen, Yukang and Yan, Feng and others},
  year={2024}
}

@article{lugaresi2019mediapipe,
  title={Mediapipe: A framework for building perception pipelines},
  author={Lugaresi, Camillo and Tang, Jiuqiang and Nash, Hadon and McClanahan, Chris and Uboweja, Esha and Hays, Michael and Zhang, Fan and Chang, Chuo-Ling and Yong, Ming Guang and Lee, Juhyun and others},
  journal={arXiv preprint arXiv:1906.08172},
  year={2019}
}

@article{mittal2012making,
  title={Making a “completely blind” image quality analyzer},
  author={Mittal, Anish and Soundararajan, Rajiv and Bovik, Alan C},
  journal={IEEE SPL},
  volume={20},
  number={3},
  pages={209--212},
  year={2012}
}

@inproceedings{wang2023exploring,
  title={Exploring clip for assessing the look and feel of images},
  author={Wang, Jianyi and Chan, Kelvin CK and Loy, Chen Change},
  booktitle={AAAI},
  pages={2555--2563},
  year={2023}
}

@article{afifi2019afif4,
title={Afif4: Deep gender classification based on adaboost-based fusion of isolated facial features and foggy faces},
author={Afifi, Mahmoud and Abdelhamed, Abdelrahman},
journal={Journal of Visual Communication and Image Representation},
volume={62},
pages={77--86},
year={2019}
}

@inproceedings{radford2021learning,
  title={Learning transferable visual models from natural language supervision},
  author={Radford, Alec and Kim, Jong Wook and Hallacy, Chris and Ramesh, Aditya and Goh, Gabriel and Agarwal, Sandhini and Sastry, Girish and Askell, Amanda and Mishkin, Pamela and Clark, Jack and others},
  booktitle={ICML},
  pages={8748--8763},
  year={2021}
}

@inproceedings{DiT,
  title={Scalable diffusion models with transformers},
  author={Peebles, William and Xie, Saining},
  booktitle={ICCV},
  pages={4172--4182},
  year={2023}
}

@inproceedings{SiT,
  title={Sit: Exploring flow and diffusion-based generative models with scalable interpolant transformers},
  author={Ma, Nanye and Goldstein, Mark and Albergo, Michael S and Boffi, Nicholas M and Vanden-Eijnden, Eric and Xie, Saining},
  booktitle={ECCV},
  pages={23--40},
  year={2024}
}

@article{yue2024difface,
  title={Difface: Blind face restoration with diffused error contraction},
  author={Yue, Zongsheng and Loy, Chen Change},
  journal={IEEE TPAMI},
  volume={46},
  number={12},
  pages={9991--10004},
  year={2024}
}

@inproceedings{lin2024diffbir,
  title={Diffbir: Toward blind image restoration with generative diffusion prior},
  author={Lin, Xinqi and He, Jingwen and Chen, Ziyan and Lyu, Zhaoyang and Dai, Bo and Yu, Fanghua and Qiao, Yu and Ouyang, Wanli and Dong, Chao},
  booktitle={ECCV},
  pages={430--448},
  year={2024}
}

@inproceedings{hu2026dereflection,
  title={Dereflection any image with diffusion priors and diversified data},
  author={Hu, Jichen and Yang, Chen and Zhou, Zanwei and Fang, Jiemin and Tian, Qi and Shen, Wei},
  booktitle={AAAI},
  pages={4860--4868},
  year={2026}
}

@inproceedings{huang2023t2i,
  title={T2i-compbench: A comprehensive benchmark for open-world compositional text-to-image generation},
  author={Huang, Kaiyi and Sun, Kaiyue and Xie, Enze and Li, Zhenguo and Liu, Xihui},
  booktitle={NeurIPS},
  pages={78723--78747},
  year={2023}
}

@inproceedings{ghosh2023geneval,
  title={Geneval: An object-focused framework for evaluating text-to-image alignment},
  author={Ghosh, Dhruba and Hajishirzi, Hannaneh and Schmidt, Ludwig},
  booktitle={NeurIPS},
  pages={52132--52152},
  year={2023}
}

@misc{openai2026gptimage2,
  author       = {{OpenAI}},
  title        = {{GPT-Image-2}},
  year         = {2026},
  howpublished = {\url{https://developers.openai.com/api/docs/models/gpt-image-2}},
  note         = {OpenAI API documentation, released April 21, 2026; accessed September 20, 2026}
}

@inproceedings{ravi2025sam,
  title={Sam 2: Segment anything in images and videos},
  author={Ravi, Nikhila and Gabeur, Valentin and Hu, Yuan-Ting and Hu, Ronghang and Ryali, Chaitanya and Ma, Tengyu and Khedr, Haitham and R{\"a}dle, Roman and Rolland, Chloe and Gustafson, Laura and others},
  booktitle={ICLR},
  pages={28085--28128},
  year={2025}
}

@article{wang2009mean,
  title={Mean squared error: Love it or leave it? A new look at signal fidelity measures},
  author={Wang, Zhou and Bovik, Alan C},
  journal={IEEE Signal Processing Magazine},
  volume={26},
  number={1},
  pages={98--117},
  year={2009}
}

@inproceedings{zhang2018unreasonable,
  title={The unreasonable effectiveness of deep features as a perceptual metric},
  author={Zhang, Richard and Isola, Phillip and Efros, Alexei A and Shechtman, Eli and Wang, Oliver},
  booktitle={CVPR},
  pages={586--595},
  year={2018}
}

@inproceedings{kumar2020luvli,
  title={Luvli face alignment: Estimating landmarks’ location, uncertainty, and visibility likelihood},
  author={Kumar, Abhinav and Marks, Tim K and Mou, Wenxuan and Wang, Ye and Jones, Michael and Cherian, Anoop and Koike-Akino, Toshiaki and Liu, Xiaoming and Feng, Chen},
  booktitle={CVPR},
  pages={8233--8243},
  year={2020}
}

\clearpage
\appendix
\startcontents[appendix]
\section*{Appendix Contents}
\begingroup
\small
\setlength{\parskip}{0pt}
\printcontents[appendix]{}{1}[3]{}
\endgroup
\clearpage

\section{Dataset Construction Details}
\label{app:dataset}

This section provides additional details of the multi-source eyeglass-reflection dataset, including the controllable synthetic pairs (Ocu-S), real-input pseudo-pairs (Ocu-P), and independent real-world test set (Ocu-R).

\subsection{Ocu-S Construction Details}
\label{app:syn_dataset}

\subsubsection{Hierarchical Factor Sampling}

\begin{table}
	\centering
	\caption{Primary high-level factor space used for Ocu-S generation.}
	\label{tab:synthetic_factors}
	\resizebox{\linewidth}{!}{%
		\begin{tabular}{lll}
			\toprule
			Group & Factor & Values / Range \\
			\midrule
			
			Subject $\mathbf{s}$
			& Pose
			& frontal; yaw $\pm15^\circ,\pm30^\circ$; pitch/roll $\pm10^\circ$ \\
			
			& Eyeglasses
			& rectangular; round; square; oval; \newline
			semi-rimless; browline; acetate; frameless \\
			
			& Expression
			& neutral; slight smile; calm; serious; \newline
			natural; relaxed; focused; slightly tired \\
			
			& Gaze
			& camera/center; left; right; up; down; \newline
			away; phone-directed; screen-directed \\
			
			\midrule
			
			Imaging $\mathbf{i}$
			& Capture mode
			& selfie; photographed by another person \\
			
			& Scene mode
			& direct; through-glass; mirror view \\
			
			& Environment
			& indoor; vehicle; outdoor \\
			
			& Background blur
			& none; slight; medium; strong \\
			
			& Camera distance
			& 0.35--2.5 m \\
			
			\midrule
			
			Reflection $\mathbf{r}$
			& Source
			& window; screen; light; outdoor; vehicle \\
			
			& Shape
			& irregular; rectangular; linear; circular \\
			
			& Color
			& neutral; warm; cool; mixed \\
			
			& Transparency
			& 15\%--95\% \\
			
			& Strength
			& mild; moderate; strong; severe \\
			
			& Target eye occlusion
			& mild; moderate; strong; severe \\
			
			\bottomrule
		\end{tabular}
	}
\end{table}

We instantiate the hierarchical sampling in \cref{eq:hierarchical_data_sampling} using the primary factor space summarized in \cref{tab:synthetic_factors}. Subject factors are sampled first, after which imaging conditions are selected through compatibility-aware conditional rules. In particular, capture mode conditions the scene configuration, environment, background blur, and camera distance, while the reflection source is further conditioned on the sampled imaging setting. Reflection shape, color, strength, transparency, and target ocular-occlusion level are then sampled according to the selected source and imaging configuration. Eyeglass styles are grouped into eight high-level frame families, with color, material, size, and related appearance variants instantiated within each family during prompt generation. Additional prompt attributes, including time of day, lighting, background complexity, camera style, reflection location and edge character, eye-area occlusion percentage, and eye-visibility targets, are instantiated from the sampled factors using deterministic mappings and factor-conditioned sampling rules.

The target ocular-occlusion levels are explicitly scheduled to be approximately balanced across mild, moderate, strong, and severe cases, accounting for about 25\% of the samples at each level. In contrast, the sampled reflection-strength levels are intentionally non-uniform. 
We distinguish this prompt-level strength control from the realized visual intensity in the generated image, which may additionally depend on reflection source, transparency, spatial extent, and other scene-dependent factors, consistent with the imperfect attribute binding and compositional control observed in text-to-image generation~\citep{huang2023t2i,ghosh2023geneval}.

\subsubsection{Pair Generation}

For each sampled factor tuple, we instantiate two structured prompts: a clean-image prompt and a reflection-edit prompt. All prompts are generated offline in advance using a fixed random seed of 2026 and stored before image
synthesis. This decouples factor sampling from image generation and allows interrupted runs to resume from the stored prompt list without resampling or duplicating previously generated factor combinations.

GPT-Image-2 first generates a clean candidate $\widetilde I_c$ from the clean-image prompt. The resulting image is then used as \emph{Image A} and edited with the corresponding reflection prompt to produce the matched reflection candidate $\widetilde I_r$. The two prompt templates are shown below, where green bracketed terms denote values instantiated from the sampled subject, imaging, and reflection factors.

\begin{promptbox}{Clean-image prompt template}
	\fontsize{8}{10}\selectfont\ttfamily
	Generate a realistic high-detail clean portrait candidate for a paired eyeglass-reflection dataset.
	
	\vspace{3pt}
	\textbf{SUBJECT / IDENTITY}
	
	Subject: \pvar{subject}\\
	Eyeglasses: \pvar{eyeglass type}\\
	Expression: \pvar{expression}\\
	Face pose: \pvar{face pose}\\
	Gaze: \pvar{gaze direction}
	
	\vspace{3pt}
	\textbf{IMAGING CONDITION}
	
	Capture mode: \pvar{capture mode}\\
	Scene mode: \pvar{scene mode}\\
	Environment: \pvar{environment} (\pvar{environment category})\\
	Time of day: \pvar{time of day}\\
	Camera-to-subject distance: \pvar{distance}\\
	Lighting: \pvar{lighting type}\\
	Background complexity: \pvar{background complexity}\\
	Background blur: \pvar{background blur}\\
	Camera style/composition: \pvar{camera style}
	
	\vspace{3pt}
	\pvar{scene-mode instruction}
	
	\vspace{3pt}
	\textbf{PAIRING CONSTRAINTS}
	
	This image will be used as the clean member of a matched clean--reflection pair. The eyeglass lenses must be transparent and optically clean, with no lens glare or lens reflection. Both eyes should be visible as allowed by the pose and gaze. Do not add text, borders, watermarks, collages, or extra panels.
	
	\vspace{3pt}
	Render at \pvar{image resolution} and keep a realistic camera-photo
	appearance with high detail.
\end{promptbox}

\begin{promptbox}{Reflection-edit prompt template}
	\fontsize{8}{10}\selectfont\ttfamily
	Edit Image A (the clean image) into the reflected member of the SAME
	matched pair.
	
	\vspace{3pt}
	\textbf{STRICT PAIRWISE CONSISTENCY}
	
	Preserve the same identity, face geometry, ocular geometry, expression, face pose, gaze direction, hairstyle, skin texture, clothing, eyeglass frame geometry, lens geometry, environment, background, camera viewpoint, framing, crop, depth of field, and overall composition.
	
	Do not change any facial or ocular contours. The intended change is ONLY the addition of realistic optical reflection inside the eyeglass lenses.
	
	\vspace{3pt}
	\textbf{REFLECTION CONTROL}
	
	Source: \pvar{reflection source}\\
	Shape: \pvar{reflection shape}\\
	Color: \pvar{reflection color}\\
	Strength: \pvar{reflection strength}\\
	Approximate opacity / reflection visibility:
	\pvar{reflection transparency}\\
	Spatial location: \pvar{reflection position}\\
	Edge character: \pvar{reflection edge}\\
	Target ocular-occlusion level: \pvar{occlusion level}\\
	Target eye-area occlusion:
	approximately \pvar{eye occlusion percentage}\%\\
	Eye visibility target: \pvar{eye visibility}
	
	\vspace{3pt}
	\textbf{PHYSICAL / LOCALITY CONSTRAINTS}
	
	The reflection must look optically plausible and conform to the lens perspective and curvature. Keep the added reflection strictly inside the transparent lens regions. Do not paint reflection onto the frame, skin, eyes themselves, hair, clothing, or background. Do not introduce new objects or alter the environment to explain the reflection. Do not change scene lighting outside the lens reflections. Do not add text, borders, watermarks, collages, or extra panels.
	
	\vspace{3pt}
	Return only the edited reflected portrait, aligned to Image A.
\end{promptbox}

Each prompt produces a single $1008\times1296$ image using GPT-Image-2. If a generation call fails, the request is retried after a 20-second delay, with at most five total attempts per prompt. If all five attempts fail, the corresponding sample is discarded and generation proceeds to the next prompt. In practice, no prompt reached the maximum number of attempts, and all scheduled samples were successfully generated. Generating the complete synthetic set required approximately 17 days.

\subsubsection{Pair Filtering}

We apply a two-stage quality-control procedure consisting of automatic edge-consistency filtering followed by manual inspection.
\begin{figure}[t!]
	\centering
	\includegraphics[width=\textwidth]{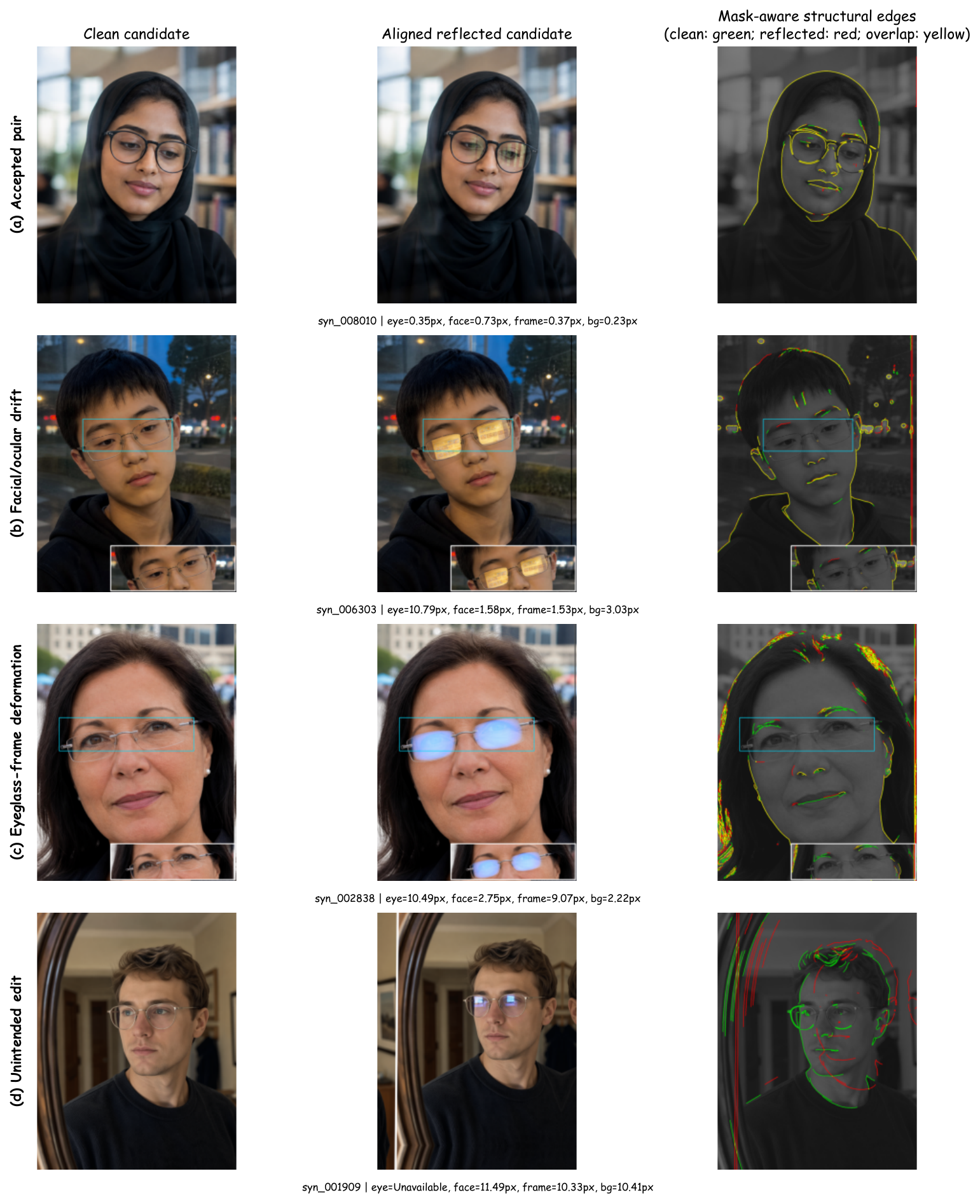}
	\caption{Representative examples of mask-aware edge-consistency filtering. Each row shows a clean candidate, its similarity-aligned reflected counterpart, and the corresponding structural-edge overlay. The examples include one accepted pair and three typical failure modes: facial/ocular drift, eyeglass-frame deformation, and unintended edits beyond the lens-reflection region.}
	\label{fig:synthetic_edge_filter}
\end{figure}

\textbf{Edge-consistency Filtering}
\label{app:edge_filter}
To tolerate small generation-induced pixel shifts while screening for pairwise structural drift, we first register the reflected candidate $\widetilde{I}_r$ to its clean counterpart $\widetilde{I}_c$ using a similarity transformation restricted to translation, rotation, and isotropic scaling. We allow at most $8$ pixels of translation, $1.5^\circ$ of rotation, and an isotropic scale within $[0.985, 1.015]$. Local non-rigid deformations are assessed during subsequent manual inspection.

After alignment, structural consistency is evaluated over three regions: facial structure outside the eyeglass area, visible periocular structure, and the approximate eyeglass-frame/bridge region. Pixels dominated by the intended lens-reflection change are excluded from the corresponding regional comparisons. For each available region $k \in \{\mathrm{face}, \mathrm{eye}, \mathrm{frame}\}$, we compute a symmetric distance-transform-based edge distance
\begin{equation}
d_k
=
\frac{1}{2}
\left[
\frac{1}{|E_c^k|}
\sum_{p \in E_c^k} D_{E_r^k}(p)
+
\frac{1}{|E_r^k|}
\sum_{q \in E_r^k} D_{E_c^k}(q)
\right],
\label{eq:edge_consistency_distance}
\end{equation}
where $E_c^k$ and $E_r^k$ denote the valid clean and aligned-reflection edge sets within region $k$, respectively, and $D_E(\cdot)$ denotes the Euclidean distance to the nearest edge in $E$. A regional distance is treated as unavailable when insufficient valid edge support remains after excluding the intended reflection-affected area.

A pair is rejected automatically if any available regional edge distance exceeds $3$ pixels. When the periocular or frame metric is unavailable because too little valid structure remains after masking strong reflections, the pair is forwarded to subsequent manual inspection. Pairs passing the automatic edge-consistency filter are manually inspected for local non-rigid deformation and other structural changes.

\cref{fig:synthetic_edge_filter} illustrates representative outcomes of the filtering procedure. The accepted example in \cref{fig:synthetic_edge_filter}(a) exhibits closely overlapping facial, ocular, frame, and background structures, indicating that the reflected image differs primarily by the intended lens reflection. In contrast, \cref{fig:synthetic_edge_filter}(b) shows a predominantly periocular failure, where the eye-region edges become spatially displaced despite relatively stable facial and frame structure. \cref{fig:synthetic_edge_filter}(c) illustrates eyeglass-frame deformation, characterized by pronounced disagreement along the frame and lens boundaries, while \cref{fig:synthetic_edge_filter}(d) exhibits widespread edge discrepancies extending beyond the eyeglass region, indicating an unintended edit to the face, subject placement, or background.

In the structural-edge overlays, yellow indicates edges shared by the clean and aligned reflected candidates, whereas green and red indicate edges appearing only in the clean and reflected images, respectively. Strong yellow overlap indicates structural consistency, while separated red--green responses indicate spatial drift or deformation. Cyan boxes mark the primary periocular/eyeglass inspection region, with enlarged local crops shown below. The displayed distances are measured in pixels, with lower values indicating stronger structural agreement. For visualization, we additionally report a background edge distance outside the expanded face region as a diagnostic measure; it is not used as a hard acceptance criterion. An unavailable eye-region value indicates that insufficient valid periocular edge support remains after excluding the intended reflection-affected area.

\textbf{Manual Inspection}

Pairs passing the automatic filter are further reviewed by four reviewers. The reviewers assess realism, identity and ocular consistency, eyeglass geometry, and preservation of unaffected image content. Samples are rejected if they exhibit identity drift, facial or ocular deformation, eyeglass-frame distortion, changes in pose, gaze, expression, background, crop, or composition, unintended modifications outside the lens-reflection regions, or other generation artifacts that compromise pairwise correspondence. A pair is retained only when accepted by reviewer consensus. The complete quality-control procedure retains 10,280 clean--reflection pairs for Ocu-S.


\subsection{Ocu-P Construction Details}
\label{app:pseudo_data}

\begin{figure}[t!]
	\centering
	\includegraphics[width=\textwidth]{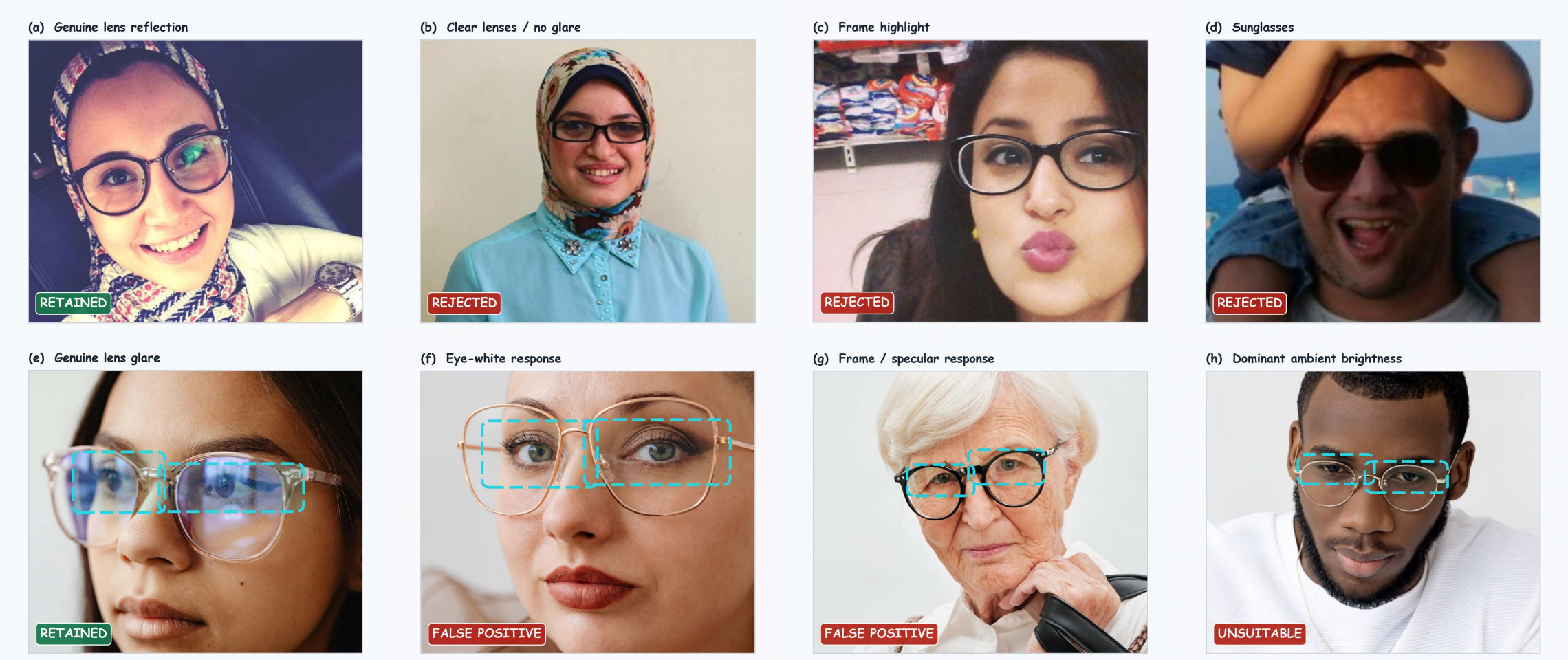}
	\caption{Representative examples of real reflected-input screening. The top row illustrates SoF samples under CLIP-based screening, including a genuine lens reflection and typical negatives such as clear lenses, frame highlights, and sunglasses. The bottom row illustrates Web samples under FaceMesh-guided local screening, including a retained lens-glare
		example and representative false positives caused by eye whites, frame/specular highlights, and dominant ambient brightness. Cyan boxes indicate the estimated lens regions used for local glare analysis.}
	\label{fig:real_input_screening}
\end{figure}

\subsubsection{Real Reflected-input Collection}
For the SoF subset, we start from 2,662 images and perform CLIP-based~\citep{radford2021learning} reflection screening using CLIP ViT-B/32. To emphasize localized lens reflections while retaining global semantic context, we combine the full-image and periocular-region CLIP features with weights of $0.4$ and $0.6$, respectively. The text representation is constructed from a multi-prompt ensemble covering five semantic categories: lens reflection, clear eyeglasses, frame highlight, sunglasses, and no glasses. For each image, we compute a reflection margin as the cosine similarity to the reflection prototype minus the highest similarity among the four negative categories, and rank all images accordingly. The top 400 candidates are manually reviewed, yielding 199 images with genuine eyeglass-lens reflections. These 199 images are used for clean-target generation and retained in the final SoF pseudo-pair subset.

\begin{wraptable}{r}{0.35\textwidth}
	\centering
	\setlength{\abovecaptionskip}{3pt}
	\setlength{\belowcaptionskip}{3pt}
	\caption{Construction of the real reflected-input subset.
		Automatic screening uses CLIP ranking for SoF and
		FaceMesh-guided reflection screening for Web images.}
	\label{tab:real_input_collection}
	\resizebox{\linewidth}{!}{%
		\begin{tabular}{lrrrr}
			\toprule
			Source & Initial & Automatic & Manual & Final \\
			\midrule
			SoF   & 2,662  & 400   & 199 & 199 \\
			Web   & 11,520 & 1,740 & 533 & 533 \\
			\midrule
			Total & 14,182 & 2,140 & 732 & 732 \\
			\bottomrule
		\end{tabular}%
	}
\end{wraptable}

For the Web subset, we retrieve 11,520 images from Pexels using eyeglass-reflection-related portrait queries. MediaPipe FaceMesh~\citep{lugaresi2019mediapipe} is used to localize the face and estimate the left and right lens regions from periocular landmarks. Within the estimated lens regions, we compute a local glare score that combines the proportions of low-saturation bright pixels and highly bright pixels, the number and area of valid highlight connected components, and the peak intensity. We retain candidates with a glare score of at least $0.58$, a white-highlight ratio of at least $0.006$, a bright-pixel ratio of at least $0.010$, and at least one valid highlight component. This automatic screening reduces the pool from 11,520 to 1,740 images. We then manually remove false positives caused by eye whites, skin or eyeglass-frame highlights, and other non-lens bright regions, as well as images without clear lens reflections, with opaque or heavily tinted lenses, or with insufficient facial or eyeglass detail. This results in 533 Web images, all of which are retained in the final pseudo-pair set after target generation and pair-level quality control. \cref{tab:real_input_collection} summarizes the screening statistics,
while \cref{fig:real_input_screening} shows representative retained examples and typical false positives from the two screening pipelines.

\subsubsection{Clean-target Generation}
For each retained real reflected input $I_r^{\mathrm{real}}$, we keep the original image unchanged and use GPT-Image-2 to generate the missing reflection-free target $\widetilde{I}_c$ at the same resolution as the input. If an editing call fails, the request is retried after a 20-second delay, with at most five total attempts per input. If all five attempts fail, the corresponding sample is discarded and processing proceeds to the next input. In practice, no input exhausted the five-attempt budget, and all retained real reflected inputs were successfully paired with generated clean targets. The original reflected image is used directly as the input of each pseudo-pair, while GPT-Image-2 is applied only to obtain its missing clean counterpart.

The editing instruction removes lens reflections while preserving identity, facial and ocular structure, pose, eyeglass geometry, background, composition, and overall photographic appearance. A representative prompt template is shown below.

\begin{promptbox}{Reflection-removal prompt template}
	\fontsize{8}{10}\selectfont\ttfamily
	Preserve the person, pose, eyeglasses, background, and composition.
	
	Remove reflections only from the eyeglass lenses, recover clear and realistic eyes, and maintain a natural photographic appearance.
	
	Do not modify identity, facial geometry, eyeglass geometry, or any non-reflection content.
\end{promptbox}

The generated image is used exclusively as the pseudo-clean target $\widetilde I_c$; the reflected member of the pair remains the original real image $I_r^{\mathrm{real}}$ without generative modification. This asymmetric construction ensures that the reflection distribution in the pseudo-pairs is inherited directly from real observations rather than from the image generator.

\subsubsection{Pseudo-pair Filtering}

We apply the same registration-aware, mask-aware edge-consistency filtering procedure described in \cref{app:edge_filter} to assess structural agreement between each real reflected input and its generated pseudo-clean target. Pairs are subsequently reviewed by four reviewers for identity drift, facial or ocular deformation, eyeglass-frame distortion, changes in pose or gaze, background or composition changes, incomplete reflection removal, and unintended modifications outside the lens regions. A pseudo-pair is retained only with unanimous agreement among all four reviewers.

The 732 reflected inputs retained after input screening, comprising 199 images from SoF and 533 from the Web subset, are used for clean-target generation. In practice, none of the resulting pseudo-pairs were rejected during pair-level quality control, and all 732 were retained in the final Ocu-P set. 

\subsection{Ocu-R Construction Details}
\label{app:real_test}

\subsubsection{Reference-pair Capture and Processing}

We collect 200 candidate reflected--clean pairs through two complementary acquisition protocols, comprising 100 offline pairs captured under controlled indoor conditions and 100 online pairs collected through remote video sessions.

For the offline subset, images are captured with an iPhone in a dim indoor environment under relatively soft ambient illumination. A clean image is first captured without an active reflection source. Controlled reflection sources, including a candle, flashlight, desk lamp, smartphone screen, and LED light stick, are then introduced while keeping the camera and subject approximately fixed. The reflection source is repositioned around the subject to produce spatially diverse lens reflections. Subject pose is also varied using near-frontal views, left/right head turns, and upward/downward head rotations. These controlled combinations yield 100 candidate reflected--clean pairs.

For the online subset, data are collected through remote WeChat video sessions. Participants prepare common controllable light sources, such as a computer screen and desk lamp, and switch or reposition them to produce visible eyeglass reflections. Clean and reflected frames are captured from the video stream under approximately unchanged subject pose and camera viewpoint, yielding another 100 candidate reflected--clean pairs.

Following the paired-data processing strategy of ReyeR~\citep{zou2024eyeglass}, we screen the candidate pairs using geometric and appearance-based consistency checks followed by manual inspection. Samples with noticeable head motion, eye or gaze changes, eyeglass-frame displacement, background shift, camera shake, or other local structural inconsistencies are removed. We then apply color-consistency correction using white-balance adjustment, histogram matching, and color transfer. After filtering and correction, 62 offline pairs and 95 online pairs are retained, resulting in 157 reference pairs.

The retained clean images are used as visual references for qualitative analysis rather than as pixel-level ground truth. Their corresponding 157 reflected images are included in Ocu-R together with 301 independently collected unpaired reflected images, forming the full 458-image real-world evaluation set. From these 458 reflected inputs, 30 images are randomly sampled for the blind user study. The clean reference images are not shown to participants and are not used in the user-study evaluation.

\subsubsection{In-the-wild Test Images}

We additionally collect 301 unpaired reflected images in unconstrained real-world environments. The images are self-captured across diverse everyday scenes, including shopping malls, subway stations, streets and intersections, supermarkets, and other outdoor locations. We use opportunistic capture: subjects wear their own eyeglasses, and each image is acquired after visually confirming the presence of lens reflection at capture time. Reflections arise naturally from surrounding illumination, including nearby artificial light sources and outdoor natural light such as low-angle sunlight around sunset.

No corresponding clean image is captured for this subset. The resulting 301 images form the unpaired real-world subset of Ocu-R. The subset follows the naturally occurring variation induced by opportunistic capture, including differences in illumination, viewpoint, head pose, eyeglass appearance, reflection source, and reflection severity. These images contribute to the real-world evaluation on Ocu-R, including no-reference assessment and blind user evaluation. Representative self-collected face images are omitted from the dataset overview figure.

\subsubsection{Real-world Test-set Characteristics}

\begin{figure}[t!]
	\centering
	\includegraphics[width=\textwidth]{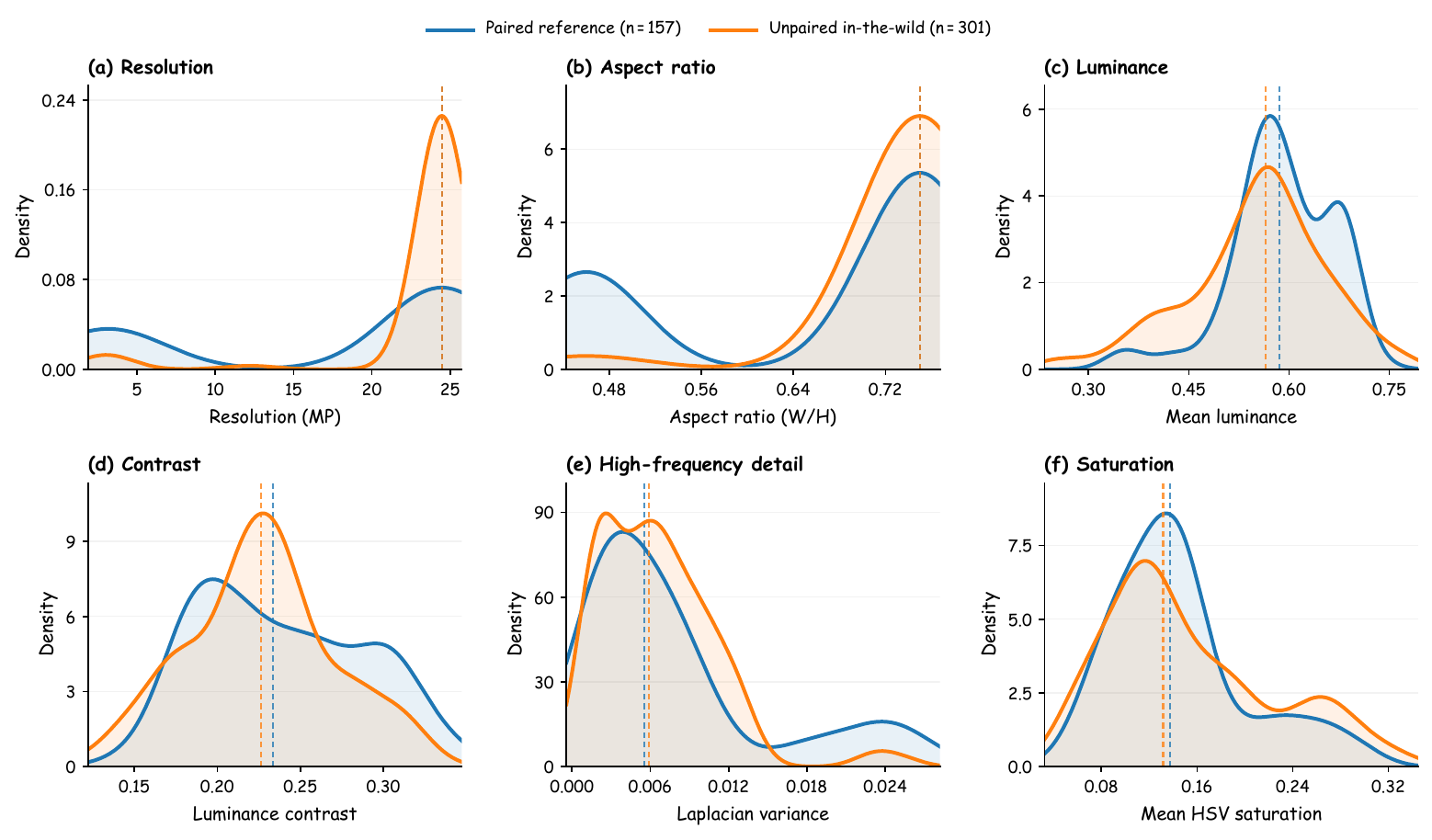}
	\caption{Image-level characteristics of the independently collected real-world test set. Smoothed distributions are shown separately for the 157 paired-reference reflected inputs and the 301 unpaired in-the-wild inputs in terms of (a) resolution, (b) aspect ratio, (c) luminance, (d) contrast, (e) high-frequency detail, and (f) color saturation. Dashed vertical lines indicate the median of each subset.}
	\label{fig:real_test_statistics}
\end{figure}

We characterize the independently collected real-world test set using aggregate image statistics over all 458 reflected inputs, comprising 157 paired-reference reflected images and 301 unpaired in-the-wild images. The corresponding clean captures from the paired-reference subset serve as visual references and are not included in this analysis.

We measure image resolution, aspect ratio, luminance, contrast, high-frequency detail, and color saturation. For an image of width $W$ and height $H$, resolution is measured in megapixels as $HW/10^6$, and aspect ratio is defined as $W/H$. The RGB image is converted to luminance using
\begin{equation}
Y = 0.299R + 0.587G + 0.114B,
\end{equation}
from which luminance and contrast are computed as the mean and standard deviation of $Y$, respectively. Color saturation is measured as the mean saturation value in HSV space.

High-frequency detail is quantified by the mean absolute Laplacian response of the luminance image,
\begin{equation}
H_{\mathrm{freq}}
=
\frac{1}{|\Omega|}
\sum_{p \in \Omega}
\left|L(p)\right|,
\end{equation}
where $L$ denotes the Laplacian response of the luminance image and $\Omega$ denotes the image domain.

To standardize the spatial scale used for photometric and high-frequency measurements, images whose longer side exceeds 512 pixels are resized with preserved aspect ratio such that the longer side is 512 pixels. Resolution and aspect ratio are always computed from the original images.

\cref{fig:real_test_statistics} summarizes the resulting image-level characteristics. The paired-reference and in-the-wild subsets exhibit broad and distinct distributions across resolution, aspect ratio, luminance, contrast, high-frequency detail, and color saturation, characterizing the range of capture conditions represented in Ocu-R.

\section{Method Details}
\label{app:method_details}

\subsection{Ocular-Prior-Guided Adaptive Representation}
\label{app:adaptive_representation}

\begin{figure}[t!]
	\centering
	\includegraphics[width=\linewidth]{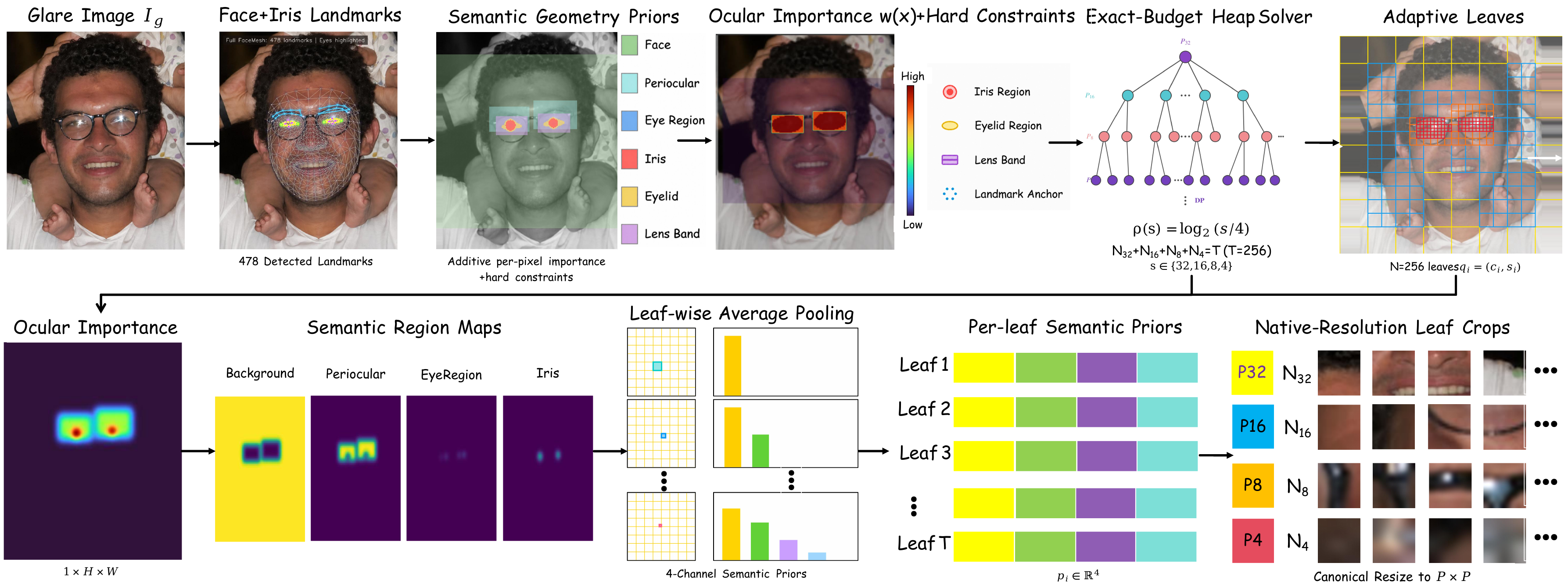}
	\caption{
		Construction of the ocular-prior-guided adaptive representation.
		Face and iris landmarks extracted from the reflected input are used to construct geometry-derived ocular supports, which define the spatial importance field and structural constraints.
		An exact-budget heap solver allocates $N=256$ quadtree leaves across four spatial scales, concentrating fine supports on ocular and lens regions while retaining coarse context elsewhere.
		The resulting leaf supports are mapped back to the native-resolution image for canonical patch sampling, while leaf geometry and pooled semantic priors provide conditioning for subsequent restoration.
	}
	\label{fig:adaptive_representation}
\end{figure}

We provide additional details of the ocular-prior-guided representation introduced in \cref{sec:adaptive_representation}. As illustrated in \cref{fig:adaptive_representation}, the reflected input is first converted into a geometry-aware importance field, from which an exact-budget quadtree determines the spatial support of each token. The resulting leaves retain both native image content and continuous geometric information for subsequent restoration.

\subsubsection{Ocular Prior and Structural Support Construction}
\label{app:ocular_prior}

All geometric cues are extracted exclusively from the reflected observation $I_g$; the clean target $I_0$ is never used for prior construction. We detect facial landmarks with MediaPipe and retain the eye contours, eyebrow landmarks, and iris landmarks of both eyes. These cues are mapped to a common $256\times256$ logical domain and used to construct a spatial importance field together with structural supports for the constrained quadtree solver.

\textbf{Logical-coordinate mapping.}
Let the original image size be $W\times H$, $S=\max(W,H)$, and $\delta_x=(S-W)/2$, $\delta_y=(S-H)/2$. A landmark with normalized image coordinates $\boldsymbol{\ell}=(u,v)$ is mapped to
\begin{equation}
T(\boldsymbol{\ell})
=
\frac{256}{S}
\left(
\delta_x+uW,\,
\delta_y+vH
\right).
\label{eq:logical_landmark_mapping}
\end{equation}
Rasterized masks are aligned to the corresponding padded square canvas and resized to the logical domain using nearest-neighbor resampling, keeping all geometry aligned with the quadtree independently of input aspect ratio.

\textbf{Primitive ocular supports.}
For each eye, 16 eye-contour landmarks together with the corresponding eyebrow landmarks define an axis-aligned periocular box with spans $d_x^k$ and $d_y^k$. We expand it by $0.22d_x^k$ horizontally, $0.35d_y^k$ upward, and $0.245d_y^k$ downward; the union of both boxes forms $M_{\mathrm{peri}}$. The filled eye-contour polygons define $M_{\mathrm{eye}}$. For four iris-contour landmarks $\{\mathbf r_j\}_{j=1}^{4}$, we compute
\begin{equation}
\mathbf c_{\mathrm{iris}}
=
\frac{1}{4}\sum_{j=1}^{4}\mathbf r_j,
\quad
r_{\mathrm{iris}}
=
1.35\,
\frac{1}{4}\sum_{j=1}^{4}
\left\|
\mathbf r_j-\mathbf c_{\mathrm{iris}}
\right\|_2 .
\label{eq:iris_support}
\end{equation}
A filled disk with center $\mathbf c_{\mathrm{iris}}$ and radius $r_{\mathrm{iris}}$ is rasterized for each eye and dilated by one logical pixel with an elliptical structuring element to obtain $M_{\mathrm{iris}}$. The eyelid support is derived directly from the eye mask as
$M_{\mathrm{lid}}
=
\operatorname{Grad}_{r=2}
\left(M_{\mathrm{eye}}\right)$,
where $\operatorname{Grad}_{r=2}$ is a binary morphological gradient using a $5\times5$ elliptical structuring element.

\textbf{Face-context and spectacle-lens supports.}
A weak face-context support is obtained from the bounding box covering both periocular regions. If its spans are $d_x$ and $d_y$, we expand it by $0.65d_x$ to both sides, $1.40d_y$ upward, and $3.00d_y$ downward, followed by clipping to the logical canvas, yielding $M_{\mathrm{face}}$. For each eye contour, let $(c_x^k,c_y^k)$ denote its bounding-box center and $(d_x^k,d_y^k)$ its spans. We estimate a spectacle-lens support with
\begin{equation}
w_{\mathrm{lens}}^k
=
(1+2\eta_x)d_x^k,
\quad
h_{\mathrm{lens}}^k
=
\max\!\left(
(1+2\eta_y)d_y^k,\,
0.55w_{\mathrm{lens}}^k
\right),
\label{eq:lens_support_size}
\end{equation}
where $\eta_x=0.45$ and $\eta_y=1.00$. The corresponding axis-aligned rectangles are united to form $M_{\mathrm{lens}}$; if detailed eye contours are unavailable, $M_{\mathrm{peri}}$ is used as a conservative fallback. This binary support is used only for adaptive allocation and is distinct from the soft renderer support $A_{\mathrm{lens}}$ in \cref{sec:renderer}.

\textbf{Sparse structural anchors.}
We further preserve structurally critical ocular locations through sparse anchors. These include the two iris centers, the inner and outer corner of each eye, and regularly subsampled eye-contour landmarks. With the 16-point contours used in our implementation, every second point is selected, producing eight contour anchors per eye. Each anchor is rasterized as a disk of radius one logical pixel, and their union defines $M_{\mathrm{anc}}$.

\textbf{Spatial importance field.}
The geometry-derived supports are combined additively into the non-negative importance field used in \cref{eq:grid_objective}:
\begin{equation}
\begin{aligned}
w(\mathbf{x})
={}&
\lambda_{\mathrm{c}}
+\lambda_{\mathrm{f}}M_{\mathrm{face}}(\mathbf{x})
+\lambda_{\mathrm{p}}M_{\mathrm{peri}}(\mathbf{x})
+\lambda_{\mathrm{e}}M_{\mathrm{eye}}(\mathbf{x})\\
&+
\lambda_{\mathrm{i}}M_{\mathrm{iris}}(\mathbf{x})
+\lambda_{\mathrm{l}}M_{\mathrm{lid}}(\mathbf{x})
+\lambda_{\mathrm{g}}M_{\mathrm{lens}}(\mathbf{x}),
\end{aligned}
\label{eq:ocular_importance}
\end{equation}
with
$(\lambda_{\mathrm{c}},
\lambda_{\mathrm{f}},
\lambda_{\mathrm{p}},
\lambda_{\mathrm{e}},
\lambda_{\mathrm{i}},
\lambda_{\mathrm{l}},
\lambda_{\mathrm{g}})
=
(0.02,\,0.20,\,1.00,\,3.00,\,8.00,\,5.00,\,4.00)$.
These weights are designed as relative importance levels rather than independently optimized parameters. We use the periocular region as a reference level and progressively increase the weight for finer ocular structures, with the iris receiving the highest priority, followed by the eyelid and spectacle-lens regions. The face and canvas terms provide weaker global coverage, with a small positive canvas weight preventing zero-cost regions. The relative scale was chosen by coarse empirical tuning on training data and then fixed for all experiments.
The supports intentionally overlap, so geometrically important locations accumulate contributions from multiple terms.
The positive canvas weight avoids zero-cost regions, while the larger ocular weights progressively bias capacity toward fine eye structures and likely spectacle regions.
The field $w$ determines the soft refinement preference.
For structural constraints, we define
$S_{\mathrm{iris}}=\operatorname{supp}(M_{\mathrm{iris}})$,
$S_{\mathrm{lid}}=\operatorname{supp}(M_{\mathrm{lid}})$,
$S_{\mathrm{lens}}=\operatorname{supp}(M_{\mathrm{lens}})$, and
$S_{\mathrm{anc}}=\operatorname{supp}(M_{\mathrm{anc}})$.

\subsubsection{Exact-Budget Quadtree Allocation}
\label{app:quadtree_allocation}

The $256\times256$ logical domain is initialized by an $8\times8$ forest of $32\times32$ roots, giving $R=64$ independent quadtrees. Each node can remain a leaf or be divided into four equal children down to the minimum size $s_{\min}=4$. We use
$\rho(s)=\log_2\frac{s}{4}$,
and define the importance mass and leaf cost of a node $q$ as
$M(q)=\int_{\Omega_q}w(\mathbf{x})\,d\mathbf{x}$ and
$C(q)=\rho(s_q)M(q)$.
For a selected leaf $q_i$, this notation coincides with the main-text definition
$M_i=M(q_i)$.
All masses are queried in constant time from an integral image of $w$.

\textbf{Structural admissibility.}
The importance field softly favors refinement, while selected structural supports impose a maximum admissible leaf size.
For a hard-support family
$\mathcal H\subseteq
\{S_{\mathrm{iris}},S_{\mathrm{lid}},S_{\mathrm{lens}},S_{\mathrm{anc}}\}$,
a leaf is admissible iff
\begin{equation}
\operatorname{adm}(q;\mathcal H)
=
\bigwedge_{S\in\mathcal H}
\left[
\Omega_q\cap S=\varnothing
\;\lor\;
s_q\leq 8
\right].
\label{eq:leaf_admissibility}
\end{equation}
Thus, any leaf larger than $8\times8$ that intersects an active iris, eyelid, lens, or anchor support must be split.

\textbf{Exact dynamic-programming formulation.}
A generic exact solution can be obtained by defining $D_q(k)$ as the minimum cost of representing the subtree rooted at $q$ with exactly $k$ admissible leaves:
\begin{equation}
D_q(k)
=
\begin{cases}
C(q),
& k=1\ \text{and}\ \operatorname{adm}(q;\mathcal H),\\[1mm]
\displaystyle
\min_{\substack{k_1+\cdots+k_4=k\\k_j\geq1}}
\sum_{j=1}^{4}D_{q_j}(k_j),
& s_q>4,\\[3mm]
+\infty,
& \text{otherwise},
\end{cases}
\label{eq:quadtree_dp}
\end{equation}
where $\{q_j\}_{j=1}^{4}$ are the four children of $q$. 
The resulting root budget curves can be combined by a second knapsack dynamic program (DP) under the global constraint $\sum_{r=1}^{R}k_r=N$. 
This formulation is exact but requires maintaining and convolving budget tables throughout the forest.

\textbf{Exact heap specialization.}
Our choice of $\rho$ admits a substantially simpler exact solver. Splitting a node $q$ of size $s$ into four children of size $s/2$ decreases the objective by
\begin{align}
\Delta(q)
= C(q)-\sum_{q'\in\operatorname{ch}(q)}C(q')
& = \rho(s)M(q) -\rho(s/2) \sum_{q'\in\operatorname{ch}(q)}M(q') \\
& = \big[\rho(s)-\rho(s/2)\big]M(q)
= M(q),
\label{eq:split_gain}
\end{align}
because the child supports partition $\Omega_q$ and $\rho(s)-\rho(s/2)=1$. Therefore, if $\mathcal S$ denotes the set of internal nodes that are split,
\begin{equation}
J(\mathcal Q)
=
J_{\mathrm{root}}
-
\sum_{q\in\mathcal S}M(q),
\label{eq:objective_split_form}
\end{equation}
where $J_{\mathrm{root}}$ is the cost of the initial 64-root forest. Since every split replaces one leaf by four, it increases the leaf count by exactly three. Hence any $N$-leaf solution contains exactly
\begin{equation}
K=\frac{N-R}{3}
\label{eq:split_budget}
\end{equation}
internal splits; with $R=64$ and $N=256$, $K=64$. Minimizing \cref{eq:objective_split_form} is therefore equivalent to selecting exactly $K$ feasible splits with maximum total importance mass.

Crucially, $w(\mathbf{x})\geq0$ implies
\begin{equation}
M(q')\leq M(q),
\qquad
\forall q'\in\operatorname{ch}(q),
\label{eq:mass_monotonicity}
\end{equation}
so split gains are non-increasing along every root-to-leaf path.
Any feasible split set is ancestor-closed because a node can be split only after its ancestors have been split.
Consequently, in the unconstrained case, repeatedly selecting the currently available node with the largest $M(q)$ produces a globally non-increasing feasible ordering of split gains and maximizes the total gain for any fixed number of splits.

The same argument applies after imposing hard structural constraints.
Let $\mathcal S_{\mathrm{hard}}$ denote the ancestor-closed set of mandatory splits required to make all current leaves admissible.
Conditioned on including $\mathcal S_{\mathrm{hard}}$, its induced leaf frontier forms a new forest whose remaining split gains retain the monotonicity in \cref{eq:mass_monotonicity}.
Therefore, selecting the next $K-K_{\mathrm{hard}}$ splits by the same max-heap rule is optimal among all exact-budget partitions satisfying the current hard constraints.
Equal-mass ties may yield different leaf layouts but identical objective values.

\textbf{Hard constraints and feasibility.}
We use the following ordered hard-support schedule:
\begin{equation}
\begin{aligned}
\mathcal H_1
&=
\{S_{\mathrm{iris}},S_{\mathrm{lid}},
S_{\mathrm{lens}},S_{\mathrm{anc}}\},\\
\mathcal H_2
&=
\{S_{\mathrm{iris}},S_{\mathrm{lid}},
S_{\mathrm{anc}}\},\\
\mathcal H_3
&=
\{S_{\mathrm{anc}}\},\\
\mathcal H_4
&=
\varnothing .
\end{aligned}
\label{eq:constraint_schedule}
\end{equation}
For each relaxation level $\mathcal H_j$, the solver starts from a fresh copy of the original $R=64$ root forest.
It recursively splits every inadmissible leaf until all frontier leaves satisfy
\cref{eq:leaf_admissibility}.
Let $\mathcal S_{\mathrm{hard}}(\mathcal H_j)$ denote the resulting set of mandatory internal splits and
\begin{equation}
K_{\mathrm{hard}}
=
\left|
\mathcal S_{\mathrm{hard}}(\mathcal H_j)
\right|.
\end{equation}
If $K_{\mathrm{hard}}>K$, that relaxation level is infeasible and is discarded entirely.
Otherwise, the remaining $K-K_{\mathrm{hard}}$ splits are selected from the induced leaf frontier using a max-heap keyed by $M(q)$.
The first feasible relaxation level is used.

Importantly, mandatory splits from an infeasible stricter level are not carried into the next level; the forest and split count are reinitialized before each solve.
Relaxing a hard support does not remove its contribution from $w$.
For example, the lens region retains its soft refinement weight even after its hard $8\times8$ maximum-leaf-size constraint is relaxed.

\textbf{Geometry fallback.}
We use all available ocular cues without requiring complete iris detection. For each detected face, missing iris landmarks remove only the corresponding iris prior, while available eye and periocular supports are retained. For multi-face inputs, the eye, iris, periocular, and lens supports from the detected faces are jointly incorporated into the spatial prior and renderer support. Adaptive allocation falls back to the uniform $16\times16$ grid only when no detected face provides reliable bilateral periocular geometry and eye centers; available partial region priors are still retained for conditioning and renderer support. If face geometry is unavailable altogether, the region prior becomes background-only and the lens support is set to zero, so the final renderer preserves the input.
\begin{algorithm}[t!]
	\caption{Exact-budget quadtree allocation}
	\label{alg:exact_heap}
	\begin{algorithmic}[1]
		\Require Importance field $w$; supports $S_{\mathrm{iris}},S_{\mathrm{lid}},S_{\mathrm{lens}},S_{\mathrm{anc}}$; target budget $N$
		\Ensure Partition $\mathcal Q^*$ with $|\mathcal Q^*|=N$
		\State Initialize the $R=64$ root forest $\mathcal Q_0$
		\State Build integral images for $w$ and all structural supports
		\If{$N<R$ or $(N-R)\bmod 3\neq0$}
		\State \Return \textbf{infeasible}
		\EndIf
		\State $K\gets(N-R)/3$
		\State Define $\mathcal H_1,\ldots,\mathcal H_4$ as in \cref{eq:constraint_schedule}
		\For{$\mathcal H\in[\mathcal H_1,\mathcal H_2,\mathcal H_3,\mathcal H_4]$}
		\State $\mathcal Q\gets$ fresh copy of $\mathcal Q_0$
		\State $\mathcal S_{\mathrm{hard}}\gets\varnothing$
		\While{there exists $q\in\mathcal Q$ with $\neg\operatorname{adm}(q;\mathcal H)$}
		\State Select an inadmissible frontier leaf $q$
		\State Replace $q$ in $\mathcal Q$ by $\operatorname{ch}(q)$
		\State $\mathcal S_{\mathrm{hard}}\gets\mathcal S_{\mathrm{hard}}\cup\{q\}$
		\If{$|\mathcal S_{\mathrm{hard}}|>K$}
		\State \textbf{break}
		\EndIf
		\EndWhile
		\If{$|\mathcal S_{\mathrm{hard}}|>K$}
		\State \textbf{continue}
		\EndIf
		\State $\mathcal S\gets\mathcal S_{\mathrm{hard}}$
		\State Initialize an empty max-heap $\mathsf{PQ}$
		\ForAll{$q\in\mathcal Q$ with $s_q>4$}
		\State Insert $q$ into $\mathsf{PQ}$ with key $M(q)$
		\EndFor
		\While{$|\mathcal S|<K$}
		\If{$\mathsf{PQ}$ is empty}
		\State \textbf{break}
		\EndIf
		\State $q\gets\operatorname{ExtractMax}(\mathsf{PQ})$
		\State Replace $q$ in $\mathcal Q$ by $\operatorname{ch}(q)$
		\State $\mathcal S\gets\mathcal S\cup\{q\}$
		\ForAll{$q'\in\operatorname{ch}(q)$ with $s_{q'}>4$}
		\State Insert $q'$ into $\mathsf{PQ}$ with key $M(q')$
		\EndFor
		\EndWhile
		\If{$|\mathcal S|=K$}
		\State \Return $\mathcal Q$ as $\mathcal Q^*$
		\EndIf
		\EndFor
		\State \Return \textbf{infeasible}
	\end{algorithmic}
\end{algorithm}
\textbf{Complexity.}
Let $A=256^2$ be the logical-canvas area.
Integral images for the importance field and structural supports require $O(A)$ time and memory.
At each of the four fixed relaxation levels, mandatory refinement is terminated once more than $K$ splits would be required, so only $O(N)$ quadtree nodes need to be materialized.
The active frontier contains at most $R+3K=N$ leaves, and each optional split performs one heap extraction and at most four insertions, each in $O(\log N)$ time.
Since the number of relaxation levels is constant, the specialized solver runs in
$O(A+N\log N)$ time and $O(A+N)$ memory.
In contrast, our direct budget-table DP implementation maintains up to $N$ states per quadtree node and combines child tables by min-plus convolution, giving a worst-case cost of $O(|V|N^2)$ and $O(|V|N)$ storage for a forest with $|V|$ possible nodes. 
The heap solver therefore exploits the particular scale-linear objective to remove the budget dimension while preserving exact optimality.
\cref{alg:exact_heap} summarizes the solver used in practice.
We retain the dynamic-programming implementation as an exact regression verifier: at each tested hard-support level, both solvers are required to agree on feasibility and objective value, although tied constant-importance regions may lead to different but equally optimal leaf layouts.
\textbf{Native-coordinate patch sampling.}
After obtaining $\mathcal Q^*$, each logical leaf support is mapped through the same square-canvas geometry back to the native image coordinate system. RGB content is sampled directly from the native-resolution image and resampled to a canonical $24\times24$ patch, yielding one patch per adaptive leaf. Thus, the $256\times256$ domain determines only the spatial partition, rather than imposing a $256\times256$ image-resolution bottleneck.
\textbf{Leaf-wise semantic priors.}
Given the adaptive partition $\mathcal Q^*$, we construct four semantic maps corresponding to background, periocular, eye, and iris regions. For each leaf $q_i$ with support $\Omega_i$, average pooling over these maps yields the leaf-wise semantic prior
\begin{equation}
\mathbf p_i
=
\begin{bmatrix}
p_{i,\mathrm{bg}}\\
p_{i,\mathrm{peri}}\\
p_{i,\mathrm{eye}}\\
p_{i,\mathrm{iris}}
\end{bmatrix}
\in[0,1]^4 .
\label{eq:leaf_semantic_prior}
\end{equation}
These pooled occupancies provide ocular-region conditioning to the restoration backbone.

\subsection{Geometry-Guided One-Step pMF}
\label{app:pmf}

\subsubsection{Geometry and Cross-Scale Conditioning}

We further specify the conditioning representation $\mathbf h$ used in \cref{sec:pmf}. For each adaptive leaf $q_i=(\mathbf c_i,s_i)$, the current path state $\mathbf z_{t,i}$ defined by \cref{eq:geometry_aware_source} and the reflected observation patch $\mathbf g_i$ are encoded separately into a state token $\mathbf e_i$ and an observation token $\mathbf o_i$, respectively. We additionally embed the continuous leaf geometry as 
$\mathbf e_i^{\mathrm{geo}} = \operatorname{PE}\!\left( \mathbf c_i/256,\, \log s_i \right)$, 
where $\operatorname{PE}$ encodes the continuous position and scale of each adaptive leaf. The geometry embedding is added to the query and key features of the conditioning attention, while values contain appearance features only. Thus, geometry controls spatial correspondence without being treated as image content.

We first inject reflected-observation features into the restoration state. Define the active token set as 
$\mathcal A = \left\{ i: 1-p_{\mathrm{bg},i}>10^{-4} \ \lor\ s_i\leq 8 \right\}$.
The first condition selects leaves with non-background ocular support, whereas the second retains all fine leaves with $s_i\in\{4,8\}$, including those allocated around lens boundaries. Using geometry-aware cross-attention, the intermediate state tokens are
\begin{equation}
\mathbf e_i'
=
\mathbf e_i
+
\mathbf 1_{\{i\in\mathcal A\}}\,
\boldsymbol{\gamma}_{\mathrm{obs}}
\odot
\operatorname{Attn}_{\mathrm{geo}}
\!\left(
\mathbf e_i,\,
\{\mathbf o_j\}_{j\in\mathcal A}
\right),
\label{eq:supp_observation_conditioning}
\end{equation}
where $\boldsymbol{\gamma}_{\mathrm{obs}}$ is a learned channel-wise residual gate. Consequently, only geometry-relevant restoration tokens are updated by this branch, while their observation features are associated through continuous position and scale.
We then supply larger-context information specifically to high-resolution ocular tokens. 
Let $\mathcal F=\{i:s_i\leq8\}$ and $\ \mathcal C=\{i:s_i\geq16\}$ denote the fine and coarse token sets. 
A second geometry-aware attention uses coarse state tokens as context and updates only fine tokens:
\begin{equation}
\mathbf h_i
=
\mathbf e_i'
+
\mathbf 1_{\{i\in\mathcal F\}}\,
\boldsymbol{\gamma}_{\mathrm{ctx}}
\odot
\operatorname{Attn}_{\mathrm{geo}}
\!\left(
\mathbf e_i',\,
\{\mathbf e_j'\}_{j\in\mathcal C}
\right).
\label{eq:supp_cross_scale_conditioning}
\end{equation}
Hence, fine tokens with $s_i\in\{4,8\}$ retain access to the broader facial appearance and illumination represented by coarse supports with $s_i\in\{16,32\}$, while coarse tokens are left unchanged by the cross-scale branch. The collection $\mathbf h=\{\mathbf h_i\}_{i=1}^{N}$ constitutes the conditioned representation used by the restoration backbone. The leaf-wise semantic prior is additionally supplied to the backbone as ocular-region conditioning. Both attention branches use learned residual gates, and their output projections are initialized to zero so that the added conditioning paths begin as identity-preserving residual branches.

\subsubsection{Observation-Anchored Source}

Let $\mathbf g=\{\mathbf g_i\}_{i=1}^{N}$ denote the adaptive patches sampled from the reflected observation. For each leaf, we define the modulation weight
\begin{equation}
m_i
=
\max\!\left(
1-p_{\mathrm{bg},i},
\tau\,\mathbf 1_{\{s_i\leq8\}}
\right),
\label{eq:supp_source_modulation}
\end{equation}
where $p_{\mathrm{bg},i}$ is the leaf-wise background occupancy and $\tau=0.75$ provides a minimum modulation level for fine leaves. The observation-anchored source is then
\begin{equation}
\widetilde{\mathbf g}
=
\mathbf g
+
\sigma\mathbf m\odot\boldsymbol\epsilon,
\qquad
\boldsymbol\epsilon\sim\mathcal N(0,\mathbf I),
\label{eq:supp_observation_source}
\end{equation}
where $\mathbf m$ stacks the token-wise modulation weights and is broadcast over the corresponding RGB patches. We use $\sigma=0.05$.

The term $1-p_{\mathrm{bg},i}$ increases perturbation according to non-background ocular occupancy, while the floor $\tau$ prevents fine leaves, including those near lens boundaries, from receiving negligible perturbation when their pooled occupancy is small. For fine leaves, $\tau=0.75$ guarantees a modulation weight of at least $0.75$, while $\sigma=0.05$ keeps the perturbed source close to the reflected observation. Both values are fixed for all experiments.

\subsubsection{pMF Parameterization}

Along the linear path defined in \cref{eq:geometry_aware_source}, the clean adaptive patches $\mathbf x_0$ correspond to $t=0$, while the observation-anchored source $\widetilde{\mathbf g}$ corresponds to $t=1$.

For an interval $0\leq r\leq t\leq1$, let $\Delta=t-r$. Given the current path state $\mathbf z_t$, the image predictor
$\widehat{\mathbf x}_{\theta}(\mathbf z_t,\Delta;\mathbf h)$
induces the average velocity
\begin{equation}
\mathbf u_{\theta}
(\mathbf z_t,r,t;\mathbf h)
=
\frac{
	\mathbf z_t-
	\widehat{\mathbf x}_{\theta}
	(\mathbf z_t,\Delta;\mathbf h)
}{t},
\qquad t>0.
\label{eq:supp_pmf_velocity}
\end{equation}
The corresponding update from $t$ to the earlier endpoint $r$ is
\begin{equation}
\widehat{\mathbf z}_r
=
\mathbf z_t
-
(t-r)\mathbf u_{\theta}
(\mathbf z_t,r,t;\mathbf h)
=
\frac{r}{t}\mathbf z_t
+
\frac{t-r}{t}
\widehat{\mathbf x}_{\theta}
(\mathbf z_t,\Delta;\mathbf h).
\label{eq:supp_pmf_update}
\end{equation}

For one-step inference, setting $(r,t)=(0,1)$ gives $\Delta=1$ and $\mathbf z_1=\widetilde{\mathbf g}$. The update therefore reduces to
\begin{equation}
\widehat{\mathbf x}
\equiv
\widehat{\mathbf z}_0
=
\widehat{\mathbf x}_{\bar\theta}
(\widetilde{\mathbf g},1;\mathbf h),
\end{equation}
which is the one-evaluation restoration rule used in \cref{sec:pmf}.

\subsubsection{MeanFlow Time Sampling}

During training, we sample the MeanFlow interval endpoints through two independent logit-normal draws. Specifically,
\begin{equation}
\xi_t,\xi_r
\overset{\mathrm{i.i.d.}}{\sim}
\mathcal N(0,1),
\quad
\widetilde t
=
\operatorname{sigmoid}(\mu+\sigma_t\xi_t),
\quad
\widetilde r
=
\operatorname{sigmoid}(\mu+\sigma_t\xi_r).
\label{eq:supp_time_candidates}
\end{equation}
To jointly expose the model to finite MeanFlow intervals and their instantaneous limit, we draw $b\sim\operatorname{Bernoulli}(p_{\mathrm{FM}})$ and collapse the pair when $b=1$:
\begin{equation}
\widetilde r
\leftarrow
\begin{cases}
\widetilde t, & b=1,\\
\widetilde r, & b=0.
\end{cases}
\quad
t=\max(\widetilde t,\widetilde r),
\quad
r=\min(\widetilde t,\widetilde r).
\label{eq:supp_time_sampling}
\end{equation}
The sorting guarantees $0\leq r\leq t\leq1$. Collapsed samples satisfy $r=t$ and therefore have zero interval length, whereas the remaining samples provide finite-interval MeanFlow conditioning. We use $\mu=-0.4$, $\sigma_t=1.0$, and $p_{\mathrm{FM}}=0.5$ in our default restoration setting.

The restoration backbone is conditioned by the current path time $t$ and the interval length
$\Delta=t-r$.
Thus, the endpoint $r$ is represented implicitly by $(t,\Delta)$. 
At inference, the one-step setting used in \cref{sec:pmf} corresponds to $(r,t)=(0,1)$ and hence $\Delta=1$.

\subsection{Native-Resolution Frequency-Preserving Synthesis}
\label{app:renderer}

\begin{figure}[t!]
	\centering
	\includegraphics[width=\textwidth]{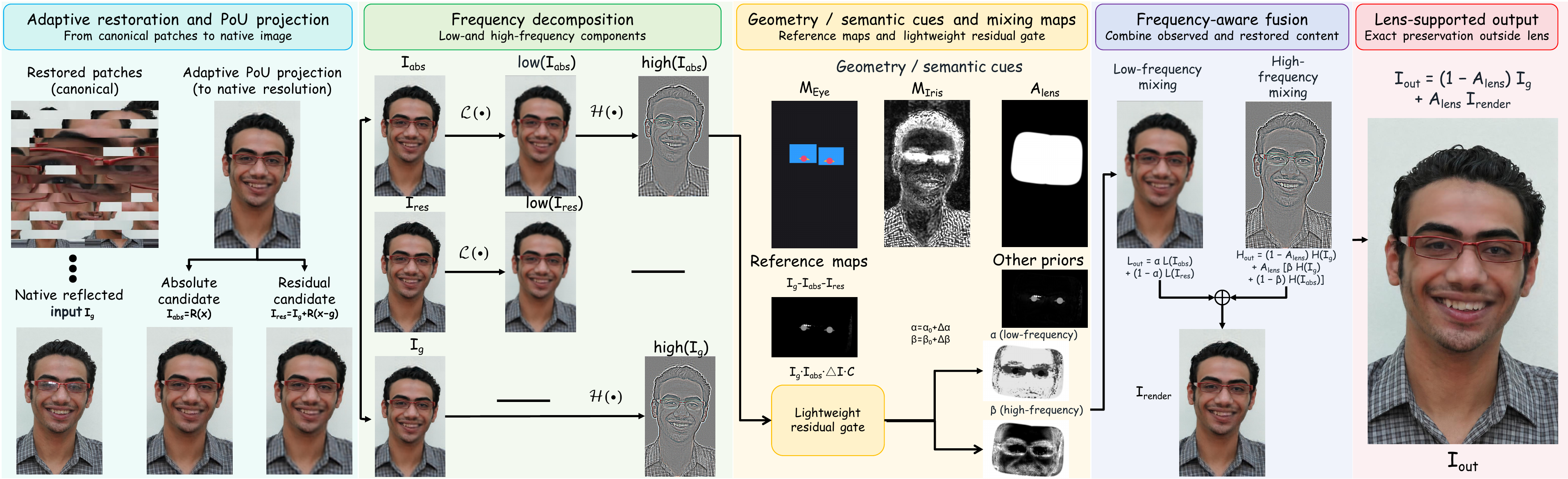}
	\caption{Overview of native-resolution frequency-preserving synthesis. Projected patch predictions form absolute and residual reconstructions. Spatial maps $\alpha$ and $\beta$ mix $\mathcal{L}(I_{\mathrm{abs}})$ with $\mathcal{L}(I_{\mathrm{res}})$ and $\mathcal{H}(I_g)$ with $\mathcal{H}(I_{\mathrm{abs}})$, respectively. Final lens-supported compositing preserves the observed input outside the spectacle region.}
	\label{fig:frequency_renderer}
\end{figure}

We provide additional details of the native-resolution synthesis module introduced in \cref{sec:renderer}. As illustrated in \cref{fig:frequency_renderer}, the restored canonical patches are first projected back to their native spatial supports. We then construct complementary absolute and residual reconstruction candidates and combine them in the frequency domain. A lightweight residual gate locally adjusts the frequency mixing, while the final lens-supported composition preserves the observed image exactly outside the editable region.

\subsubsection{Native Projection and Lens Support}

We first detail the projection operator $\mathcal R$ used in \cref{sec:renderer}. For each adaptive leaf $q_i$, let $\mathcal W_i(\mathbf y_i)$ denote bicubic resampling of its canonical patch $\mathbf y_i$ onto the corresponding native-resolution support. Each support is extended by a small proportional halo and weighted by a separable raised-cosine window $w_i$ that equals one over the interior and smoothly decays to zero across the halo. Overlapping predictions are then combined by normalized partition-of-unity blending:
\begin{equation}
\mathcal R(\mathbf y)(\mathbf p)
=
\frac{
	\sum_i w_i(\mathbf p)\,\mathcal W_i(\mathbf y_i)(\mathbf p)
}{
	\sum_i w_i(\mathbf p)+\varepsilon
}.
\label{eq:supp_native_projection}
\end{equation}
This normalized overlap suppresses seams between neighboring adaptive supports while preserving the linearity of $\mathcal R$, which is used by both absolute and residual reconstruction in \cref{eq:render_candidates}.

The soft lens-support map $A_{\mathrm{lens}}$ is constructed independently of the adaptive partition. Spectacle geometry derived from the ocular landmarks is rasterized on the logical canvas, expanded to cover the lens region, and feathered before being mapped to the native resolution. As a result, the final compositing support is spatially smooth and does not inherit quadtree boundaries from $\mathcal Q^{*}$.

\subsubsection{Spatial Gate Construction}

The frequency gates distinguish regions that require reconstruction from those where native photographic information can be safely retained. We implement the native-resolution low-pass operator $\mathcal L$ as a $15\times15$ box filter, and define the corresponding high-frequency component as in \cref{sec:renderer}.

To estimate where the restored patches consistently request a change to the observation, we first form the projected correction
\begin{equation}
\Delta I
=
\mathcal R(\widehat{\mathbf x}-\mathbf g),
\quad
C
=
\operatorname{mean}_{\mathrm{RGB}}
\left|
\mathcal L_c(\Delta I)
\right|,
\label{eq:supp_correction_evidence}
\end{equation}
where $\mathcal L_c$ is an $11\times11$ box filter. We convert this magnitude into a soft reconstruction-need map
\begin{equation}
P_{\mathrm{corr}}
=
s\!\left(
\frac{C-0.03}{0.10-0.03}
\right),
\quad
s(u)
=
\bar u^{\,2}(3-2\bar u),
\quad
\bar u=\operatorname{clip}(u,0,1).
\label{eq:supp_reconstruction_need}
\end{equation}
Thus, correction magnitudes below $0.03$ receive little reconstruction evidence, while values above $0.10$ saturate the gate.

We additionally obtain eye and iris supports from the ocular prior. Their binary maps are smoothed independently with a $7\times7$ box filter and passed through the same smoothstep function, yielding soft maps $\widetilde M_{\mathrm{eye}}$ and $\widetilde M_{\mathrm{iris}}$. To identify strong near-white reflections, we convert the native observation from $[-1,1]$ to $[0,1]$ as
$I_g^{01}=(I_g+1)/2$ and define
\begin{equation}
B
=
\operatorname{mean}_{\mathrm{RGB}} I_g^{01},
\quad
\chi
=
\max_{\mathrm{RGB}} I_g^{01}
-
\min_{\mathrm{RGB}} I_g^{01}.
\end{equation}
Here $B$ is a brightness cue and $\chi$ measures local chroma spread. The soft highlight response is
\begin{equation}
S
=
A_{\mathrm{lens}}\,
s\!\left(
\frac{B-0.82}{0.96-0.82}
\right)
s\!\left(
\frac{0.16-\chi}{0.16}
\right).
\label{eq:supp_highlight_gate}
\end{equation}
A response therefore requires both high brightness and low chroma within the lens support.

These signals define analytical initializations for the two frequency gates. For the low-frequency branch,
\begin{equation}
\alpha_0
=
A_{\mathrm{lens}}\,
\max\!\left(
0,\,
0.35\widetilde M_{\mathrm{eye}},\,
0.55\widetilde M_{\mathrm{iris}},\,
0.90S
\right).
\label{eq:supp_alpha_init}
\end{equation}
Thus, ordinary lens regions initially favor residual reconstruction, whereas eye, iris, and strong-highlight regions receive progressively larger contributions from the absolute clean candidate.

For the high-frequency branch, let
$\operatorname{lerp}(a,b;m)
=
(1-m)a+mb$.
We first define the within-lens native-detail weight
\begin{align}
\widetilde\beta_0
={}
\operatorname{lerp}\!\left(
\operatorname{lerp}
\left(
0.90,\,
0.30;\,
\widetilde M_{\mathrm{eye}}
\right),
0.15;\,
\widetilde M_{\mathrm{iris}}
\right)
\times
\left(
1-0.75P_{\mathrm{corr}}
\right)
(1-S),
\end{align}
and then extend it to the full image as
\begin{equation}
\beta_0
=
(1-A_{\mathrm{lens}})
+
A_{\mathrm{lens}}\widetilde\beta_0.
\label{eq:supp_beta_init}
\end{equation}
Accordingly, native high-frequency information receives an initial weight of $0.90$ over ordinary lens regions, $0.30$ over the eye, and $0.15$ over the iris, before attenuation by correction evidence and the highlight cue. Outside the soft lens support, $\alpha_0=0$ and $\beta_0=1$.

A lightweight learned gate refines these analytical maps rather than predicting the rendered image itself. Its input comprises
$\left[
I_g,\,
I_{\mathrm{abs}},\,
\Delta I,\,
C,\,
\widetilde M_{\mathrm{eye}},\,
\widetilde M_{\mathrm{iris}},\,
A_{\mathrm{lens}},\,
\alpha_0,\,
\beta_0
\right]$,
forming $15$ channels. Before entering the gate, $\Delta I$ is scaled by $0.5$ and clipped to $[-1,1]$, while $C$ is normalized by $0.10$ and clipped to $[0,2]$. All inputs are bilinearly resized to a fixed $256\times256$ gate canvas.

The gate uses a $3\times3$ convolutional stem with width $32$, followed by three depthwise-separable residual blocks with dilation rates $1$, $2$, and $4$. Each block consists of a depthwise $3\times3$ convolution, a pointwise $1\times1$ convolution, GroupNorm, and SiLU activation. A final $1\times1$ convolution predicts two residual maps. The entire gate contains only $8{,}578$ trainable parameters. Its output head is initialized to zero, so optimization starts exactly from the analytical gates $\alpha_0$ and $\beta_0$.

Let
$(\delta_\alpha,\delta_\beta)
=
\tanh
G_\phi
\left(
I_g,
I_{\mathrm{abs}},
\Delta I,
C,
\widetilde M_{\mathrm{eye}},
\widetilde M_{\mathrm{iris}},
A_{\mathrm{lens}},
\alpha_0,
\beta_0
\right)$
denote the predicted residuals after bilinear upsampling to the native image resolution. The final gates are
\begin{equation}
\alpha
=
\operatorname{clip}
\left(
\alpha_0
+
A_{\mathrm{lens}}\delta_\alpha,\,
0,1
\right),
\quad
\beta
=
\operatorname{clip}
\left(
\beta_0
+
A_{\mathrm{lens}}\delta_\beta,\,
0,1
\right).
\label{eq:supp_frequency_gate}
\end{equation}
They are used directly in \cref{eq:frequency_render}. The learned module therefore adjusts only the two spatial mixing maps, while the frequency decomposition and final lens-supported composition remain unchanged.

\section{Experimental Details}
\label{app:Experiment_Details}

\subsection{Implementation and Evaluation}
\label{app:implementation}


\subsubsection{Training Setup and Objectives}
\label{app:objectives}
\label{app:resolution}

\textbf{Training data handling.}
Ocu-S pairs are stored at $1008\times1296$. For Ocu-P, each real reflected input retains its original resolution, and the corresponding pseudo-clean target is generated at the same resolution. During training, adaptive patches are extracted directly from the stored paired images at their respective resolutions.

OcuFlow is trained with a velocity-space restoration objective together with image-space constraints for reconstruction, ocular fidelity, and preservation. The native-resolution renderer is optimized with a separate frequency-aware objective.

\textbf{Velocity-space restoration.}
For the conditional path in \cref{sec:pmf}, let
$\mathbf v^{\star}=\widetilde{\mathbf g}-\mathbf x_0$
denote the instantaneous path velocity, and let $(r,t)$ be sampled as described in \cref{app:pmf}. The image-valued prediction induces the average velocity $\mathbf u_\theta$. We evaluate its total derivative along the ground-truth path using a Jacobian--vector product,
\begin{equation}
D_t\mathbf u_\theta
=
\operatorname{JVP}_{(\mathbf z,t,r)}
\!\left[
\mathbf u_\theta;
(\mathbf v^{\star},1,0)
\right],
\qquad
\mathbf U_\theta
=
\mathbf u_\theta
+
(t-r)\operatorname{sg}\!\left[D_t\mathbf u_\theta\right],
\label{eq:supp_meanflow_field}
\end{equation}
where $\operatorname{sg}[\cdot]$ denotes stop-gradient. We supervise $\mathbf U_\theta$ toward $\mathbf v^{\star}$ and retain a training-only instantaneous-velocity head $\mathbf v_\theta^{\mathrm{aux}}$. We apply detached per-image adaptive normalization
\begin{equation}
\operatorname{AN}(E)
=
\frac{E}
{\operatorname{sg}\!\left[(E+\epsilon_{\mathrm n})^p\right]},
\qquad
p=1,
\qquad
\epsilon_{\mathrm n}=0.01.
\label{eq:supp_adaptive_norm}
\end{equation}

For image $b$, we define
\begin{align}
E_U^{(b)}
&=
\kappa
\sum_{i=1}^{N}
\left\|
\mathbf U_{\theta,i}^{(b)}
-
\mathbf v_i^{\star,(b)}
\right\|_2^2,
\\
E_V^{(b)}
&=
\kappa
\sum_{i=1}^{N}
\left\|
\mathbf v_{\theta,i}^{\mathrm{aux},(b)}
-
\mathbf v_i^{\star,(b)}
\right\|_2^2,
\qquad
\kappa=\left(\frac{16}{24}\right)^2.
\end{align}
The factor $\kappa$ compensates for the increased canonical pixel count relative to the $16\times16$ reference scale. The velocity-space objective is
\begin{equation}
\mathcal L_{\mathrm{vel}}
=
\frac{1}{B}
\sum_{b=1}^{B}
\left[
\operatorname{AN}(E_U^{(b)})
+
\operatorname{AN}(E_V^{(b)})
\right].
\label{eq:supp_velocity_loss}
\end{equation}
The auxiliary velocity branch is used only during training.

\textbf{Restoration regularization.}
We impose reconstruction fidelity at medium and coarse canonical scales. Let $\mathcal D_k$ denote area resampling to $k\times k$. We define
\begin{equation}
\mathcal E_k
=
\frac{1}{BN}
\sum_{b=1}^{B}
\sum_{i=1}^{N}
\frac{16^2}{k^2}
\left\|
\mathcal D_k(\widehat{\mathbf x}_{0,i}^{(b)})
-
\mathcal D_k(\mathbf x_{0,i}^{(b)})
\right\|_2^2,
\qquad
\mathcal L_x
=
\mathcal E_{12}
+
0.5\,\mathcal E_6.
\label{eq:supp_reconstruction_loss}
\end{equation}

Semantic priors weight first-order gradient supervision over ocular structures:
\begin{equation}
w_i^{\mathrm{eye}}
=
0.05\,p_{i,\mathrm{peri}}
+
p_{i,\mathrm{eye}}
+
2\,p_{i,\mathrm{iris}},
\end{equation}
\begin{equation}
e_i^\nabla
=
\frac{1}{2}
\left(
\left\|
\nabla_x\widehat{\mathbf x}_{0,i}
-
\nabla_x\mathbf x_{0,i}
\right\|_1^{\mathrm{mean}}
+
\left\|
\nabla_y\widehat{\mathbf x}_{0,i}
-
\nabla_y\mathbf x_{0,i}
\right\|_1^{\mathrm{mean}}
\right),
\end{equation}
and
\begin{equation}
\mathcal L_{\mathrm{eye}}
=
\frac{1}{B}
\sum_{b=1}^{B}
\frac{
	\sum_i
	w_i^{\mathrm{eye},(b)}
	e_i^{\nabla,(b)}
}{
	\sum_i
	w_i^{\mathrm{eye},(b)}
	+\varepsilon
}.
\label{eq:supp_eye_loss}
\end{equation}

Using the geometry-guided restoration support $m_i$ defined in \cref{sec:pmf}, let
\begin{equation}
d_i
=
\left\|
\widehat{\mathbf x}_{0,i}
-
\mathbf g_i
\right\|_1^{\mathrm{mean}}.
\end{equation}
Leaves are weighted by their corresponding spatial area:
\begin{equation}
\mathcal L_{\mathrm{pres}}
=
\frac{1}{B}
\sum_{b=1}^{B}
\frac{
	\sum_i
	s_i^2(1-m_i)d_i
}{
	\sum_i
	s_i^2(1-m_i)
	+\varepsilon
}.
\label{eq:supp_preservation_loss}
\end{equation}

The complete restoration objective is
\begin{equation}
\mathcal L_{\mathrm{rest}}
=
\mathcal L_{\mathrm{vel}}
+
0.05\,\mathcal L_x
+
0.005\,\mathcal L_{\mathrm{eye}}
+
0.01\,\mathcal L_{\mathrm{pres}}.
\label{eq:supp_full_restoration_loss}
\end{equation}

\textbf{Renderer objective.}
The frequency-preserving renderer is supervised over the full-resolution lens support. Let $A=A_{\mathrm{lens}}$ and define the lens-normalized average of a $C$-channel field $F$ as
\begin{equation}
\langle F\rangle_A
=
\frac{
	\sum_{\mathbf x,c}
	A(\mathbf x)F_c(\mathbf x)
}{
	C\sum_{\mathbf x}A(\mathbf x)+\varepsilon
}.
\end{equation}

We use reconstruction, low-frequency, and first-order gradient terms:
\begin{align}
\mathcal L_{\mathrm{rec}}
&=
\left\langle
\sqrt{
	(I_{\mathrm{out}}-I_0)^2+\epsilon_c^2
}
\right\rangle_A,
\qquad
\epsilon_c=10^{-3},
\\
\mathcal L_{\mathrm{low}}
&=
\left\langle
\left|
\mathcal L(I_{\mathrm{out}})
-
\mathcal L(I_0)
\right|
\right\rangle_A,
\\
\mathcal L_{\mathrm{grad}}
&=
\frac{1}{2}
\sum_{d\in\{x,y\}}
\left\langle
\left|
\nabla_d I_{\mathrm{out}}
-
\nabla_d I_0
\right|
\right\rangle_{A_d},
\end{align}
where $\mathcal L$ is the same $15\times15$ low-pass operator used by the renderer and $A_d$ denotes the valid lens support for the corresponding finite-difference direction.

Let
\begin{equation}
\mathbf r_\phi
=
\left[
A_{\mathrm{lens}}\Delta\alpha_\phi,
A_{\mathrm{lens}}\Delta\beta_\phi
\right]
\end{equation}
denote the lens-masked residual gate fields. Their magnitude and spatial variation are regularized by
\begin{equation}
\mathcal L_{\mathrm{gate}}
=
\|\mathbf r_\phi\|_1^{\mathrm{mean}},
\qquad
\mathcal L_{\mathrm{TV}}
=
\frac{1}{2}
\left(
\|\nabla_x\mathbf r_\phi\|_1^{\mathrm{mean}}
+
\|\nabla_y\mathbf r_\phi\|_1^{\mathrm{mean}}
\right).
\end{equation}

The complete renderer objective is
\begin{equation}
\mathcal L_{\mathrm{rend}}
=
\mathcal L_{\mathrm{rec}}
+
0.5\,\mathcal L_{\mathrm{low}}
+
0.1\,\mathcal L_{\mathrm{grad}}
+
0.01\,\mathcal L_{\mathrm{gate}}
+
0.01\,\mathcal L_{\mathrm{TV}}.
\label{eq:supp_renderer_loss}
\end{equation}

\textbf{Optimization.}
We initialize the restoration backbone from pretrained \textbf{pMF-B-16} weights~\citep{lu2026onestep} and train OcuFlow for $60$k steps with batch size $6$. Fixed-size adaptive patch representations are processed jointly across the batch, while native-resolution renderer supervision is evaluated sample by sample and accumulated before each optimizer update.

The first $45$k steps train the restoration backbone and adaptive conditioner with the renderer frozen. We then freeze the restoration network and optimize the renderer gate for $5$k steps, followed by $10$k steps of joint fine-tuning. The renderer objective is linearly ramped to its full weight during the first $2$k joint steps.

We use AdamW with $(\beta_1,\beta_2)=(0.9,0.95)$ and weight decay $10^{-4}$. The backbone and conditioner use learning rate $2\times10^{-5}$ during the first stage and $2\times10^{-6}$ and $10^{-5}$ during joint fine-tuning, respectively. The renderer uses learning rate $10^{-4}$ whenever it is optimized. Training uses BF16 precision, gradient clipping at global norm $5$, and EMA decay $0.999$.


\subsubsection{Inference Details}
\label{app:inference_details}

For all reported results, we use the final EMA checkpoint. Given an input image of size $H\times W$, OcuFlow edge-pads the shorter spatial dimension to a square canvas of side $S=\max(H,W)$. The estimated ocular geometry is mapped to the $256\times256$ logical domain, where the adaptive quadtree produces $N=256$ leaves.

Each leaf is mapped to its corresponding support on the $S\times S$ image canvas, and its RGB content is bicubically resampled into a $24\times24$ canonical patch. The observation-anchored source perturbation described in \cref{sec:pmf} is applied using a fixed evaluation seed. Restoration uses the endpoint setting $(r,t)=(0,1)$ and requires one restoration-network evaluation (1-NFE).

The restored canonical patches are projected onto their corresponding full-resolution supports and combined with the padded observation by the frequency-preserving renderer. The padded region is then removed to recover the supplied $H\times W$ image extent. Inference uses BF16 precision.


\subsubsection{Evaluation Protocols}
\label{app:evaluation}

\textbf{Common paired-evaluation protocol.}
For the paired benchmark, all methods receive inputs standardized to $1008\times1296$ using the same aspect-ratio-preserving resize-and-pad transform. Each method then applies its own internal preprocessing as required. Predictions are mapped back through the inverse external transform to the original paired-image resolution, where all full-reference metrics are computed.

\textbf{Evaluation rationale.}
Eyeglass reflections typically affect a relatively small portion of the image while directly degrading the appearance and local structure of the ocular region~\citep{yi2025dl2g}. Consequently, full-image PSNR can be dominated by the much larger unaffected facial and background regions: because PSNR is derived from spatially averaged mean-squared error, errors within a small but task-critical eye region may contribute only weakly to the final score. This limitation is consistent with the well-established shortcomings of MSE-based fidelity measures for perceptually important image distortions~\citep{wang2009mean}. We therefore report Mask-PSNR and Mask-SSIM within the eyeglass region, together with Mixed-PSNR and Mixed-SSIM to measure restoration inside the affected region while accounting for preservation elsewhere.

Regional pixelwise agreement alone does not fully characterize the quality of ocular restoration. PSNR and SSIM can remain insensitive to perceptually meaningful changes in local appearance and structure~\citep{zhang2018unreasonable}. We therefore additionally report Eye-ROI LPIPS for perceptual similarity and landmark-based errors for ocular geometry. This task-oriented evaluation is also motivated by prior eyeglass-reflection-removal studies, which have shown that reflections affect iris localization and facial landmark detection and have evaluated restoration through downstream iris or landmark accuracy~\citep{sandhan2017anti,zou2024eyeglass}. Eye-NME and Iris-NME directly quantify displacement of eye-contour and iris landmarks relative to the clean reference, complementing the regional reconstruction and perceptual metrics. \cref{tab:evaluation_metrics} summarizes the paired and real-world evaluation metrics.

\begin{table}[t]
	\centering
	\caption{Summary of evaluation metrics.}
	\label{tab:evaluation_metrics}
	\resizebox{\linewidth}{!}{%
		\begin{tabular}{@{}llc@{}}
			\toprule
			Metric & Region / Purpose & Better \\
			\midrule
			Mask-PSNR / SSIM
			& Eyeglass region; local restoration fidelity
			& $\uparrow$ \\
			Mixed-PSNR / SSIM
			& Eyeglass and non-eyeglass regions; restoration--preservation balance
			& $\uparrow$ \\
			Out-PSNR
			& Outside eyeglass region; RGB preservation
			& $\uparrow$ \\
			Out-HF-PSNR
			& Outside eyeglass region; high-frequency detail preservation
			& $\uparrow$ \\
			Eye-ROI LPIPS
			& Binocular ROI; perceptual ocular fidelity
			& $\downarrow$ \\
			Eye-NME
			& Eye contours; ocular geometric fidelity
			& $\downarrow$ \\
			Iris-NME
			& Iris landmarks; iris geometric fidelity
			& $\downarrow$ \\
			\midrule
			NIQE~\citep{mittal2012making}
			& Full real image; no-reference naturalness
			& $\downarrow$ \\
			CLIP-IQA+~\citep{wang2023exploring}
			& Full real image; perceptual quality
			& $\uparrow$ \\
			User preference
			& Real-world study set; visual preference
			& $\uparrow$ \\
			\bottomrule
		\end{tabular}
	}
\end{table}

\textbf{Masks and regional reconstruction metrics.}
RGB values are normalized to $[0,1]$. Binary eyeglass masks are initialized using Grounded-SAM-2~\citep{ren2024grounded}: Grounding DINO uses the text prompt ``glasses frames'' with box and text thresholds of $0.25$, and the detected components are segmented using the SAM~2.1 Hiera-Large checkpoint~\citep{ravi2025sam}. All detected eyeglass components are merged and manually corrected before evaluation. The resulting evaluation mask $\mathcal M$ defines the local region used for regional reconstruction metrics and is distinct from the soft renderer support $A_{\mathrm{lens}}$ in \cref{sec:renderer}. Let $\overline{\mathcal M}$ denote its complement within the valid image support.

Mask-PSNR is computed from the RGB mean-squared error within $\mathcal M$. To additionally account for preservation outside the eyeglass region, Mixed-PSNR combines the regional errors in the MSE domain:
\begin{equation}
E_{\mathrm{mix}}
=
\gamma E_{\mathrm{mask}}
+
(1-\gamma)E_{\mathrm{out}},
\qquad
\operatorname{Mixed\text{-}PSNR}
=
10\log_{10}\frac{1}{E_{\mathrm{mix}}},
\qquad
\gamma=0.5,
\label{eq:mixed_psnr}
\end{equation}
where $E_{\mathrm{mask}}$ and $E_{\mathrm{out}}$ are the RGB mean-squared errors over $\mathcal M$ and $\overline{\mathcal M}$, respectively.

For the preservation-oriented ablations, we additionally report Out-PSNR and Out-HF-PSNR over the valid image region outside the eyeglass mask. Out-PSNR is obtained directly from the outside-region RGB error:
\begin{equation}
\operatorname{Out\text{-}PSNR}
=
10\log_{10}\frac{1}{E_{\mathrm{out}}},
\label{eq:out_psnr}
\end{equation}
where $E_{\mathrm{out}}$ is defined above.

To specifically evaluate preservation of fine image detail, Out-HF-PSNR compares high-frequency residuals outside the eyeglass mask.
For an image $I$, we define
\begin{equation}
H_{\mathrm{eval}}(I)
=
I-G_{\sigma}(I),
\end{equation}
where $G_{\sigma}$ denotes Gaussian filtering with a $7\times7$ kernel and $\sigma=1.5$. Let $\Omega_{\mathrm{out}}$ denote the valid image support outside $\mathcal M$. The high-frequency error is
\begin{equation}
E_{\mathrm{out}}^{\mathrm{HF}}
=
\frac{1}{3|\Omega_{\mathrm{out}}|}
\sum_{\mathbf{x}\in\Omega_{\mathrm{out}}}
\left\|
H_{\mathrm{eval}}(I_{\mathrm{out}})(\mathbf{x})
-
H_{\mathrm{eval}}(I_0)(\mathbf{x})
\right\|_2^2,
\end{equation}
and we report
\begin{equation}
\operatorname{Out\text{-}HF\text{-}PSNR}
=
10\log_{10}\frac{1}{E_{\mathrm{out}}^{\mathrm{HF}}}.
\label{eq:out_hf_psnr}
\end{equation}
Higher values indicate closer preservation of RGB appearance and high-frequency content, respectively, outside the eyeglass region.

For SSIM, we compute a full grayscale SSIM map using a $7\times7$ window. The mask and complementary support are each eroded by three pixels before regional averaging so that the local SSIM window remains within the corresponding region. Mask-SSIM is the mean SSIM over the eroded eyeglass support. Let $S_{\mathrm{mask}}$ and $S_{\mathrm{out}}$ denote the regional SSIM means; we report
\begin{equation}
\operatorname{Mixed\text{-}SSIM}
=
\gamma S_{\mathrm{mask}}
+
(1-\gamma)S_{\mathrm{out}},
\qquad
\gamma=0.5.
\label{eq:mixed_ssim}
\end{equation}

\textbf{Ocular perceptual and geometric fidelity.}
Regional PSNR and SSIM quantify reconstruction agreement within the task-relevant area, while Eye-ROI LPIPS and landmark-based errors characterize perceptual and geometric changes to the ocular structures themselves. Eye-ROI LPIPS is evaluated over a binocular region defined from clean-reference MediaPipe landmarks~\citep{lugaresi2019mediapipe}. The joint eye bounding box is enlarged by $25\%$ horizontally and $35\%$ vertically, and the same crop is applied to the prediction and reference. Both crops are aspect-preservingly resized and reflect-padded to $128\times128$, after which LPIPS is computed using the AlexNet backbone.

For geometric fidelity, MediaPipe landmarks with iris refinement are independently detected from the clean reference and restored image. Eye-NME measures displacement of the left- and right-eye contour landmarks, while Iris-NME measures displacement of the refined iris landmarks. Following standard face-alignment evaluation practice~\citep{kumar2020luvli}, we report normalized mean error (NME) using an inter-ocular normalization distance. For landmark set $\mathcal K$,
\begin{equation}
\operatorname{NME}_{\mathcal K}
=
\frac{100}{|\mathcal K|}
\sum_{j\in\mathcal K}
\frac{
	\left\|
	\widehat{\mathbf p}_j-\mathbf p_j
	\right\|_2
}{
	\left\|
	\mathbf p_{33}-\mathbf p_{263}
	\right\|_2
},
\label{eq:ocular_nme}
\end{equation}
where $\mathbf p_j$ and $\widehat{\mathbf p}_j$ are landmarks detected from the clean reference and restored image, respectively. The normalization distance is measured between reference landmarks $33$ and $263$, corresponding to the two outer eye corners. Eye-NME and Iris-NME are reported as percentages.

\textbf{Downstream iris localization.}
Building on the iris-localization evaluation used in Anti-Glare~\citep{sandhan2017anti}, we evaluate iris localization on the reflected inputs and restoration outputs. For each paired sample, iris landmarks detected from the clean image are converted into left- and right-iris circles and used as reference targets. The same iris-localization pipeline is independently applied to the reflected input and to the output of each restoration method.

For each eye, localization accuracy is measured by the intersection-over-union (IoU) between the predicted and reference iris circles. An eye is considered correctly localized when its circle IoU is at least $\tau$. At the image level, a sample is counted as successful when both eyes satisfy the threshold. We report the bilateral localization success rate across IoU thresholds and the minimum IoU of the two eyes. All methods are evaluated using the same downstream detector without downstream fine-tuning or method-specific adaptation.

\textbf{Real-world metrics and blind user study.}
We evaluate all $458$ Ocu-R images at their native resolutions using NIQE~\citep{mittal2012making} and CLIP-IQA+~\citep{wang2023exploring}. Both metrics are computed using the default PyIQA configurations from the final restored images, without additional manual resizing or normalization.

We additionally conduct a blind multi-way preference study on $30$ images randomly sampled from Ocu-R using a fixed random seed of $2026$. All $133$ participants evaluate the same $30$ images, yielding $133\times30=3{,}990$ selections. For each trial, the original reflected input is displayed above the outputs of all nine compared methods. Method identities are hidden, and the presentation order of the nine outputs is independently randomized for each participant and image. Participants select exactly one result, with no tie option, based jointly on reflection-removal effectiveness and preservation of the original image content. All $30$ trials are required for submission, and all $133$ submitted questionnaires are complete. Informed consent is obtained before participation.

We report each method's overall preference share over all $3{,}990$ selections and its image wins, defined as the number of test images on which it receives the largest number of participant selections. No image-level ties occur. Uncertainty in the overall preference share is estimated using a two-way bootstrap over participants and images. For each of $5{,}000$ bootstrap replicates, $133$ participants and $30$ images are independently resampled with replacement, and the preference share is recomputed over the resulting crossed responses. The $2.5$th and $97.5$th percentiles define the reported $95\%$ confidence interval.


\subsubsection{Baseline Configurations}
\label{app:baselines}

\textbf{Training and checkpoint selection.}
All learned restoration baselines are retrained or fine-tuned on the OcuBench training data using their respective training recipes and initialization strategies. When a method defines a validation protocol, its checkpoint is selected according to that protocol using a validation subset drawn from the OcuBench training data. Otherwise, the final available checkpoint is used.

We compare against Stable Diffusion~1.5~\citep{rombach2022high}, ER$^2$Net~\citep{zou2024eyeglass}, DSRNet~\citep{hu2023single}, Stable Diffusion~3.5~\citep{esser2024scaling}, DSIT~\citep{hu2024single}, RDNet~\citep{zhao2025reversible}, ReflexSplit~\citep{Lee_2026_CVPR}, and DL2G~\citep{yi2025dl2g}. Their method-specific inference configurations are summarized in \cref{tab:baseline_configs}.

\begin{table}[t]
	\centering
	\caption{Method-specific inference configurations. Spatial dimensions follow the W$\times$H convention.}
	\label{tab:baseline_configs}
	\begin{adjustbox}{max width=\linewidth}
		\begin{tabular}{@{}llll@{}}
			\toprule
			Method & Internal input handling & Key inference settings & Precision \\
			\midrule
			Stable Diffusion 1.5
			& Full-image resize to $992\times768$; masked inpainting
			& DDIM; 50 configured / 35 executed; strength $0.7$; CFG $6.0$
			& FP16 \\
			ER$^2$Net
			& Short-side resize to 256; dimensions aligned to multiples of 16
			& Single-pass restoration
			& FP32 \\
			DSRNet
			& Dimensions aligned to multiples of 32; $768\times768$ tiles
			& 64-px overlap; overlapping-tile reconstruction
			& FP32 \\
			Stable Diffusion 3.5
			& Full-image resize to $512\times512$; masked inpainting
			& FlowMatchEulerDiscreteScheduler; 28 configured / 15 executed; strength $0.55$; CFG $4.5$
			& BF16 \\
			DSIT
			& Reflect-padding to multiples of 32; $384\times384$ tiles
			& 64-px overlap; overlapping-tile reconstruction
			& FP32 \\
			RDNet
			& $320\times320$ tiles; boundary padding when required
			& 64-px overlap; Hann-weighted fusion
			& FP32 \\
			ReflexSplit
			& Reflect-padding to multiples of 32; $768\times768$ tiles
			& 64-px overlap; overlapping-tile reconstruction
			& FP32 \\
			DL2G
			& Full-image resize to $256\times256$
			& DEM--LSDM--GCR; $64\times64$ LSDM patches, stride 16; 100 DDIM steps
			& FP16 autocast \\
			\textbf{OcuFlow}
			& Edge padding; $256\times256$ logical domain; $24\times24$ canonical patches
			& $N=256$ adaptive leaves; 1-NFE; native-resolution synthesis
			& BF16 autocast \\
			\bottomrule
		\end{tabular}
	\end{adjustbox}
\end{table}

\textbf{Diffusion baseline details.}
For Stable Diffusion~1.5, we use the pretrained
\href{https://huggingface.co/stable-diffusion-v1-5/stable-diffusion-v1-5}{Stable Diffusion~1.5 weights}
together with our ReGlass-IP Lite adapter. The positive prompt is
\emph{``Remove the reflection on the eyeglasses.''}, and the negative prompt is
\emph{``low quality, worst quality, blurry, distorted face, deformed eyes, extra eyes, bad anatomy, artifacts, watermark, text''}.

For Stable Diffusion~3.5, we use the pretrained
\href{https://huggingface.co/stabilityai/stable-diffusion-3.5-medium}{Stable Diffusion~3.5 Medium weights}
with our task-adapted ControlNet. The positive prompt is
\emph{``Remove the reflection on the eyeglasses.''}, and the negative prompt is
\emph{``reflection, glare, specular highlight, white streak, bright spot, lens reflection''}.
The task-adapted ControlNet is evaluated at its $512\times512$ training resolution. Both diffusion baselines use a fixed base seed of $42$ with deterministic per-image seed offsets.


\subsubsection{Efficiency Protocol}
\label{app:efficiency}

All methods are benchmarked on the same NVIDIA RTX~4090 using the standardized paired-benchmark inputs defined in \cref{app:evaluation}. The common benchmark preprocessing is completed before timing. Timing covers the complete method-specific inference pipeline, including internal resizing, padding, tiling, sampling, preprocessing, and postprocessing. Three warm-up runs are excluded.

Runtime is reported as the average wall-clock time per image under sustained inference. Each method uses its normal CPU--GPU parallelism and asynchronous execution during benchmarking. File encoding and disk I/O are excluded.

OcuFlow is evaluated with batch size $1$, BF16 precision, and 1-NFE. Its timed pipeline includes ocular-geometry extraction, adaptive-grid construction, native-resolution patch sampling, canonical resampling, one-step pMF restoration, partition-of-unity projection, frequency-preserving synthesis, and final cropping. Ocular-geometry extraction uses four asynchronous CPU workers and runs concurrently with GPU inference.

\subsection{Additional Quantitative Analysis}
\label{app:additional_qutitity_results}

\textbf{Complementary restoration and ocular gains.} 
The results in Tab.~\ref{tab:comparison_all} show that OcuFlow improves restoration quality, ocular perceptual fidelity, and geometric consistency simultaneously rather than optimizing a single reconstruction criterion. Among the seven paired-reference metrics, OcuFlow ranks first on six and second only on Mixed-SSIM, where it differs from the best result by $0.0004$ ($0.7074$ versus $0.7078$). Compared with DSIT, the strongest competing baseline across the four reconstruction metrics, OcuFlow improves Mask-PSNR by $0.5654$ dB, Mixed-PSNR by $0.1889$ dB, and Mask-SSIM by $0.0155$, while reducing Eye-ROI LPIPS by $29.2\%$, Eye-NME by $8.7\%$, and Iris-NME by $6.7\%$. Relative to RDNet, the closest competitor in Eye-ROI LPIPS, OcuFlow further reduces Eye-ROI LPIPS by $10.9\%$, Eye-NME by $10.2\%$, and Iris-NME by $9.5\%$, while improving Mask-PSNR and Mixed-PSNR by $1.0366$ dB and $0.6277$ dB, respectively. These results show that the gains in regional reconstruction are accompanied by improved ocular appearance and geometry rather than obtained at their expense.

\textbf{Quality--efficiency trade-off.}
OcuFlow requires $56.4551$ ms/image of steady-state wall-clock time and $1.04$ GB of GPU memory. Relative to ER$^2$Net, the second-fastest baseline, OcuFlow is approximately $4.3\times$ faster. Compared with DSIT, OcuFlow is approximately $54.8\times$ faster and uses $64.9\%$ less GPU memory while improving six of the seven paired-reference metrics. Relative to RDNet, it is approximately $47.9\times$ faster and uses $30.2\%$ less GPU memory while outperforming RDNet on all seven paired restoration and ocular metrics. These results reflect the fixed adaptive-token budget, one-step pMF restoration, and lightweight native-resolution synthesis used by OcuFlow.

\textbf{Real-world evaluation.} The main paper summarizes no-reference image quality and human preference on the independent real-world set in \cref{fig:realworld_evaluation}. For completeness, Tabs.~\ref{tab:realworld_nr} and~\ref{tab:user_study} provide the corresponding numerical results. The evaluation protocols, including the sampling and statistical analysis of the user study, are detailed in \cref{app:evaluation}.

\begin{table}[t]
	\centering
	\begin{minipage}[t]{0.44\linewidth}
		\caption{No-reference evaluation on $458$ independent real-world images. The reflected input is included for reference but excluded from restoration-method ranking. Best and second-best restoration results are shown in \textbf{bold} and \underline{underlined}, respectively.}
		\label{tab:realworld_nr}
		\centering
		\resizebox{!}{5em}{%
			\begin{tabular}{lrr}
				\toprule
				Method & NIQE $\downarrow$ & CLIP-IQA+ $\uparrow$ \\
				\midrule
				\rowcolor{blockgray}
				Input & 5.2583 & 0.6556 \\
				\midrule
				Stable Diffusion 1.5 & 9.7805 & 0.6127 \\
				ER$^2$Net & 12.8999 & 0.5568 \\
				DSRNet & \underline{5.5638} & 0.6312 \\
				Stable Diffusion 3.5 & 11.4307 & 0.5737 \\
				DSIT & 5.7683 & 0.6265 \\
				RDNet & 9.3812 & \textbf{0.6442} \\
				ReflexSplit & 5.6417 & 0.6128 \\
				DL2G & 12.7582 & 0.5473 \\
				\midrule
				\rowcolor{oursgreen}
				\textbf{Ours} & \textbf{5.2369} & \underline{0.6419} \\
				\bottomrule
			\end{tabular}
		}
	\end{minipage}
	\hfill
	\begin{minipage}[t]{0.53\linewidth}
		\caption{Blind user study on $30$ images randomly sampled from the $458$-image real-world set with seed $2026$. All $133$ participants completed all $30$ trials, yielding $3{,}990$ selections. Image Wins counts images on which a method receives the largest number of selections.}
		\label{tab:user_study}
		\centering
		\resizebox{!}{5em}{%
			\begin{tabular}{lrrr}
				\toprule
				Method & Votes & Preference (\%) $\uparrow$ & Image Wins $\uparrow$ \\
				\midrule
				DSIT & 433 & 10.85 & 3/30 \\
				ReflexSplit & 276 & 6.92 & 1/30 \\
				ER$^2$Net & 220 & 5.51 & 0/30 \\
				Stable Diffusion 1.5 & 89 & 2.23 & 0/30 \\
				DSRNet & 84 & 2.11 & 0/30 \\
				RDNet & 78 & 1.95 & 0/30 \\
				DL2G & 75 & 1.88 & 0/30 \\
				Stable Diffusion 3.5 & 49 & 1.23 & 0/30 \\
				\midrule
				\rowcolor{oursgreen}
				\textbf{Ours} & \textbf{2686} & \textbf{67.32} & \textbf{26/30} \\
				\bottomrule
			\end{tabular}
		}
	\end{minipage}
\end{table}

\textbf{No-reference image quality.} As shown in Tab.~\ref{tab:realworld_nr}, OcuFlow achieves the lowest NIQE among all restoration methods, improving from $5.5638$ for the second-best DSRNet to $5.2369$. For CLIP-IQA+, RDNet obtains the highest restoration score of $0.6442$, while OcuFlow ranks second at $0.6419$, a difference of only $0.0023$. Notably, the reflected input itself receives a CLIP-IQA+ score of $0.6556$, higher than any restored output despite retaining the target reflection artifacts. This observation illustrates that generic full-image no-reference metrics do not directly measure reflection-removal correctness. We therefore treat NIQE and CLIP-IQA+ as complementary indicators of global perceptual quality rather than task-specific measures of successful ocular restoration.

\textbf{Blind user preference.} The human study in Tab.~\ref{tab:user_study} provides substantially stronger separation between methods. OcuFlow receives $2{,}686$ of the $3{,}990$ selections, corresponding to an overall preference share of $67.32\%$. The next most selected method, DSIT, receives $10.85\%$, giving OcuFlow a $56.47$ percentage-point margin and approximately $6.2\times$ its preference share. OcuFlow receives the largest number of selections on $26$ of the $30$ images ($86.7\%$), compared with three images for DSIT and one for ReflexSplit. A two-way participant-by-image bootstrap with $5{,}000$ replicates gives a $95\%$ confidence interval of $[59.32\%,74.71\%]$ for OcuFlow's overall preference share. Its lower bound remains $9.32$ percentage points above $50\%$, while the $26/30$ image-win rate shows that the aggregate preference is distributed broadly across the evaluated examples rather than concentrated on only a small subset.

\textbf{Cross-protocol consistency.} The paired-reference benchmark, generic no-reference metrics, and blind user study capture complementary aspects of restoration. The paired benchmark measures regional reconstruction fidelity, ocular perceptual similarity, and landmark consistency against paired references; NIQE and CLIP-IQA+ characterize global perceptual quality on the independent real-world set; and the blind study measures task-level human preference under simultaneous consideration of reflection removal and content preservation. OcuFlow performs strongly across these protocols, while the relatively modest separation under generic NR-IQA and the substantially larger separation in human preference demonstrate that no single metric family fully characterizes eyeglass-reflection removal. Together with the Eye-ROI LPIPS and ocular NME results, these findings motivate reporting complementary regional, geometric, no-reference, and human-perceptual criteria.

\subsection{Downstream Iris Localization}
\label{app:downstream_iris}

\begin{figure}[t!]
	\centering
	\includegraphics[width=\linewidth]
	{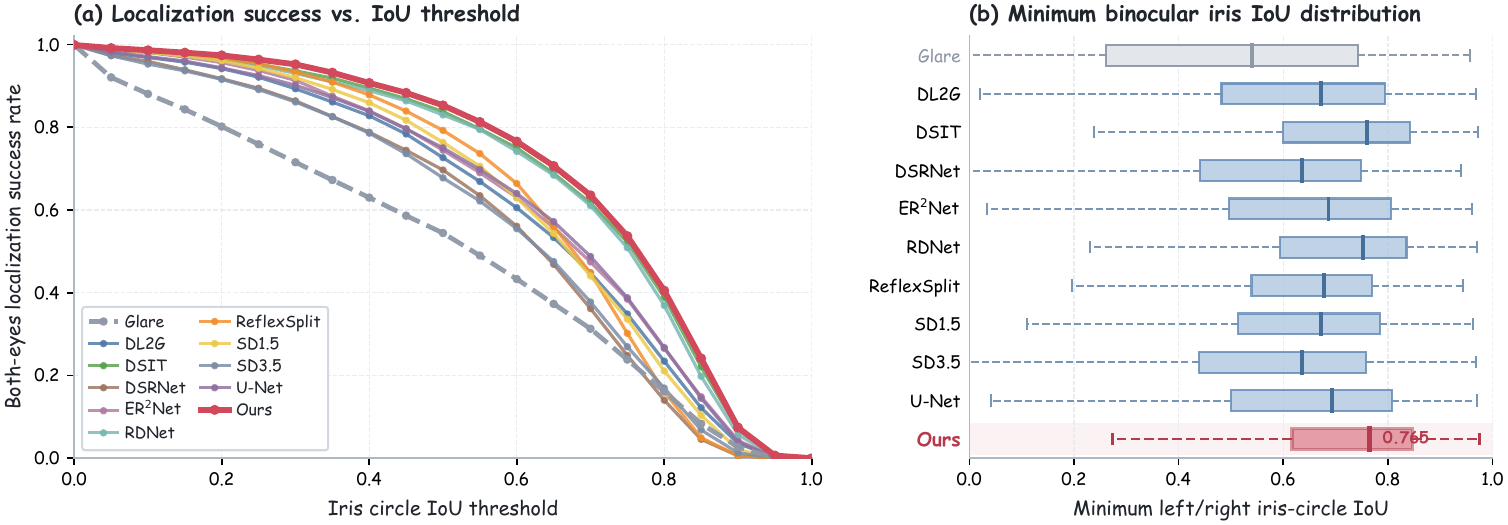}
	\caption{
		Downstream iris-localization evaluation.
		\textbf{(a)} Both-eyes localization success rate under different iris-circle
		IoU thresholds.
		\textbf{(b)} Distribution of the minimum left/right iris-circle IoU for each
		method. Glare denotes the original reflected input without reflection removal,
		while OcuFlow is highlighted in red.
	}
	\label{fig:downstream_iris}
\end{figure}

Beyond restoration fidelity, we further examine whether eyeglass-reflection removal improves the reliability of a downstream ocular-analysis task. Specifically, we evaluate iris localization using a fixed iris-landmark detector, without method-specific adaptation or downstream fine-tuning. The goal is to assess whether the restored ocular structures are not only visually plausible, but also sufficiently consistent for subsequent vision processing.

For each paired sample, iris circles estimated from the clean image are used as pseudo-ground truth. The same detector is independently applied to the reflected input and to the output of each restoration method. Let $\operatorname{IoU}_{\mathrm{left}}$ and
$\operatorname{IoU}_{\mathrm{right}}$ denote the circle IoUs of the two eyes.
We define the image-level localization score as
$s_{\mathrm{iris}}
=
\min\left(
\operatorname{IoU}_{\mathrm{left}},
\operatorname{IoU}_{\mathrm{right}}
\right)$,
such that an image is considered successfully localized at threshold $\tau$
only when
$s_{\mathrm{iris}} \geq \tau$.
This conservative criterion requires both eyes to be accurately localized.

As shown in \cref{fig:downstream_iris}(a), eyeglass reflections substantially degrade downstream iris localization, with the original reflected input exhibiting a markedly lower success rate over a broad range of IoU thresholds. Reflection removal recovers a large portion of this lost localization performance, while OcuFlow remains among the strongest methods throughout the moderate-to-high threshold range. At the default threshold $\tau=0.5$, OcuFlow achieves a both-eyes localization success rate of $85.30\%$, compared with $54.43\%$ for the reflected input and $83.78\%$ for the strongest competing method, DSIT. This corresponds to a $30.87$ percentage-point improvement over the unprocessed reflected input.
\cref{fig:downstream_iris}(b) further examines the distribution of the minimum left/right iris IoU, which reflects the localization quality of the more difficult eye in each image. OcuFlow exhibits a clear rightward shift relative to the reflected input and achieves a median minimum IoU of $0.765$. This distributional improvement indicates that the downstream gain is not restricted to a single operating threshold, but instead reflects more consistently recovered iris geometry across the evaluation set.

These results complement Eye-NME and Iris-NME by measuring functional downstream utility rather than geometric error alone. While the landmark-based metrics quantify agreement with the clean reference, the downstream localization experiment directly evaluates whether a fixed ocular-analysis pipeline can reliably operate on the restored images. The results therefore show that reflection removal restores task-relevant iris structure, with OcuFlow providing strong overall downstream localization performance.

\subsection{Additional Qualitative Results}
\label{app:additional_qualitative_results}

We provide additional paired qualitative comparisons to complement \cref{fig:qualitative_comparison}. The examples cover diverse subjects, eyeglass styles, reflection intensities, and ocular appearances. Each comparison includes the reflected input, clean reference, OcuFlow, and the competing methods evaluated in the main paper. Enlarged ocular crops are shown to facilitate inspection of residual reflections, ocular structure, and local appearance preservation.

\begin{figure}[t!]
	\centering
	\includegraphics[width=\linewidth]{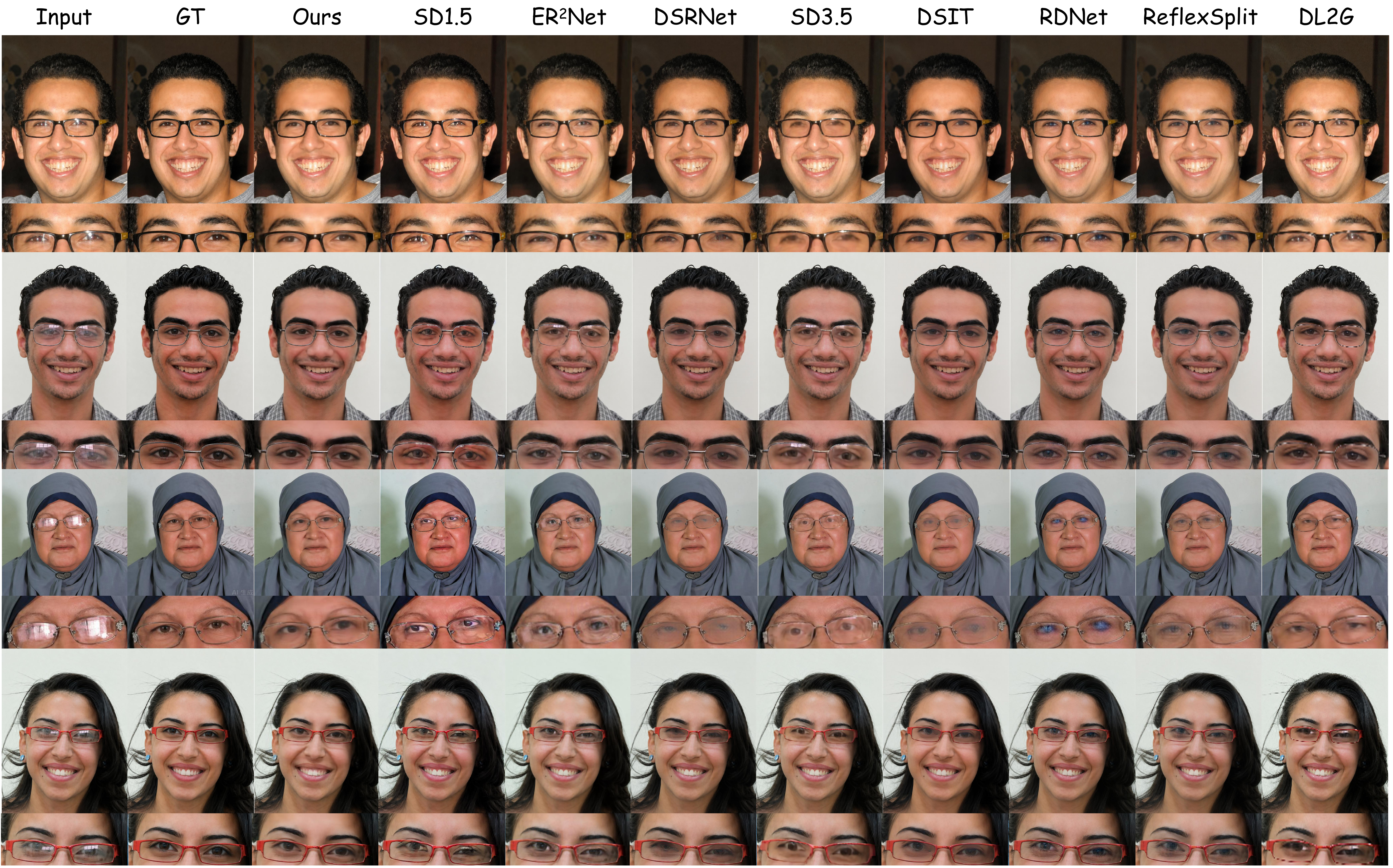}
	\caption{
		Additional qualitative comparisons on held-out paired OcuBench examples.
		Enlarged ocular regions are shown below the corresponding full-image comparisons to highlight residual reflections, iris and eyelid structure, and local detail preservation.
	}
	\label{fig:additional_quality_pub}
\end{figure}

\begin{figure}[t!]
	\centering
	\includegraphics[width=\linewidth]{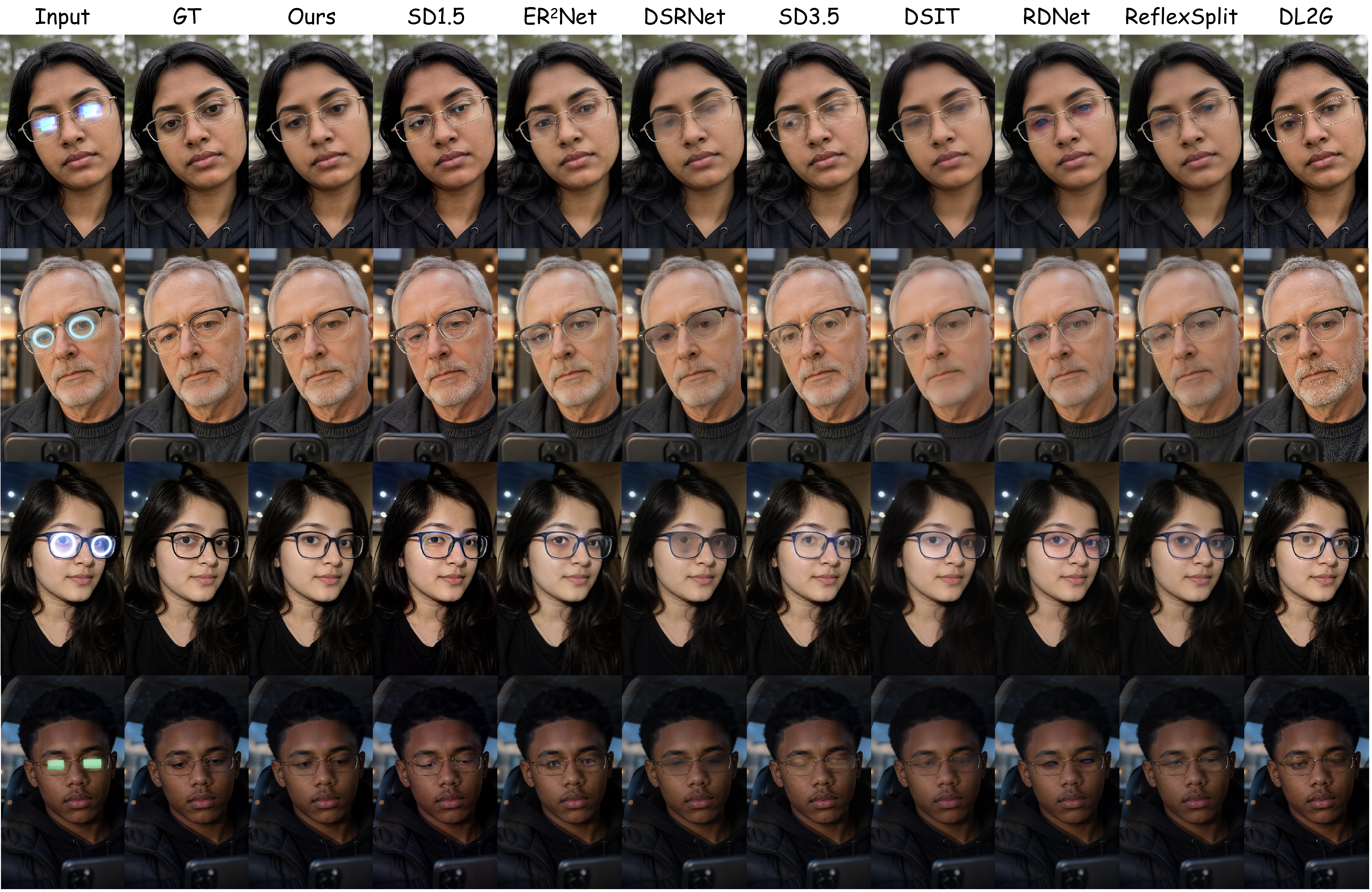}
	\caption{
		Additional qualitative comparisons on held-out paired OcuBench examples.
		Methods are shown in the same order as in \cref{fig:additional_quality_pub}.
	}
	\label{fig:additional_quality_syn}
\end{figure}

\begin{figure}[t!]
	\centering
	\includegraphics[width=\linewidth]{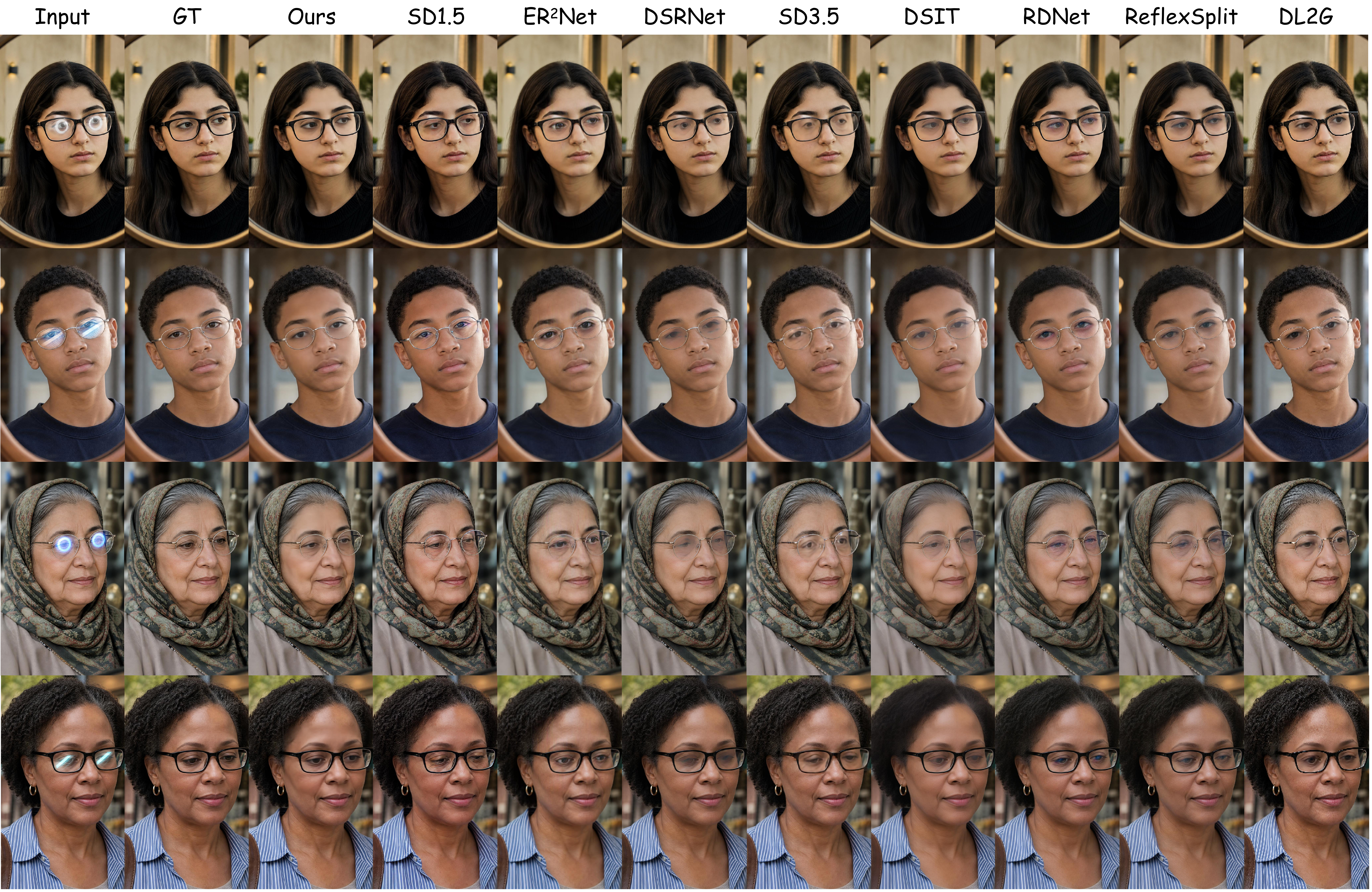}
	\caption{
		Further qualitative comparisons on held-out paired OcuBench examples, covering additional subjects, reflection patterns, and ocular appearances.
	}
	\label{fig:additional_quality_syn2}
\end{figure}

\subsubsection{Qualitative comparison.}
Across Figs.~\ref{fig:additional_quality_pub}--\ref{fig:additional_quality_syn2}, the competing methods exhibit several recurring types of restoration error. ER$^2$Net~\citep{zou2024eyeglass} and DSRNet~\citep{hu2023single} often suppress the strongest reflection response while leaving residual glare or reduced contrast within the lens region. Under stronger occlusion, the recovered iris and eyelid boundaries can remain less distinct than in the clean reference. DSIT~\citep{hu2024single} and ReflexSplit~\citep{Lee_2026_CVPR} generally remove a larger fraction of the visible reflection, but several examples exhibit smoother ocular appearance and attenuated local contrast. RDNet~\citep{zhao2025reversible} frequently retains sharper local structure, although blue or purple chromatic residuals remain around the iris or lens region in multiple cases. Stable Diffusion~1.5~\citep{rombach2022high} and Stable Diffusion~3.5~\citep{esser2024scaling} can substantially alter local color, contrast, or facial appearance beyond the original reflection pattern. DL2G~\citep{yi2025dl2g} preserves global facial appearance well in several examples but can introduce locally inconsistent structures around the eyeglass rims and eyelid boundaries.

These differences become particularly visible for saturated highlights, broad colored reflections, and glare overlapping the pupil or iris. In such cases, incomplete suppression leaves visible contamination, while stronger correction can introduce smoothing, chromatic shifts, or local structural changes. Across the shown examples, OcuFlow more closely follows the clean references in both the spatial extent of the correction and the recovered ocular appearance. Localized glare is removed while eyeglass boundaries and surrounding facial texture remain stable, and strongly colored reflections are restored with fewer persistent chromatic residuals. Under severe ocular occlusion, the recovered iris position, eyelid contour, and local eye contrast remain closer to the clean reference than the smoother or more strongly altered outputs observed in several competing methods. These comparisons illustrate the restoration--preservation trade-off in eyeglass reflection removal: effective suppression of corrupted content should be accompanied by preservation of ocular geometry and reliable local detail.

\begin{figure}[t!]
	\centering
	\includegraphics[width=\linewidth]{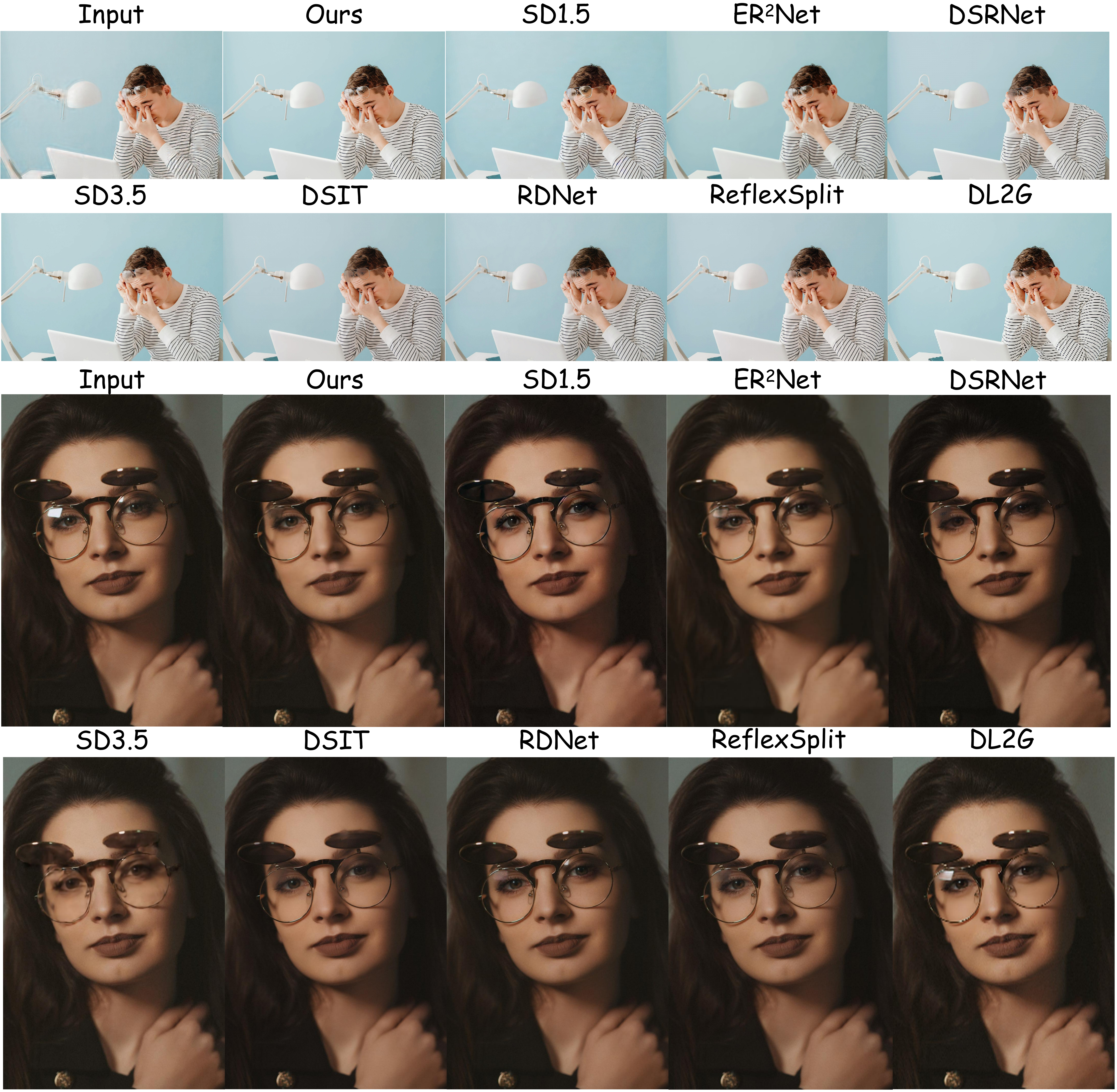}
	\caption{
		Ocular-prior-guided selective restoration.
		In the upper example, the eyeglasses are displaced away from the visible eyes, and OcuFlow leaves the lens appearance largely unchanged.
		In the lower example, a strong reflection overlaps the eye; OcuFlow applies stronger correction to the ocularly relevant region while making smaller changes to surrounding lens reflections.
	}
	\label{fig:selective_preservation}
\end{figure}

\subsubsection{Ocular-prior-guided selective restoration.}
\cref{fig:selective_preservation} illustrates the spatial selectivity induced by the ocular prior. In the upper example, the eyeglasses are displaced above the eyes, so the visible lens reflections do not overlap the ocular structures. OcuFlow consequently leaves the input largely unchanged. Several competing methods modify the lens regions more extensively, producing partial reflection suppression, local contrast changes, or irregular residual structures. In this example, DSIT and ReflexSplit introduce eye-like ghosting patterns within the lenses, while DL2G produces noticeable local texture distortion.

The lower example presents the complementary case. A strong bright reflection overlaps the left eye and obscures useful ocular detail, whereas the larger reflections near the upper parts of the lenses lie mostly outside the visible eye regions. OcuFlow applies stronger restoration around the affected eye while making smaller changes to the surrounding lens appearance. Several competing methods show broader or less localized changes: ER$^2$Net and DSRNet retain visible degradation around the ocular region, SD3.5 substantially alters the overall lens appearance, and DL2G leaves a strong bright residual near the eye together with local structural artifacts.

These examples illustrate how the ocular prior modulates the spatial extent of restoration according to the relationship between lens reflections and visible ocular structures. The resulting behavior reduces unnecessary changes to lens regions that do not interfere with eye visibility while concentrating stronger correction on reflections that overlap the eyes.

\subsubsection{Failure cases.}

\begin{figure}[t!]
	\centering
	\includegraphics[width=0.55\linewidth]
	{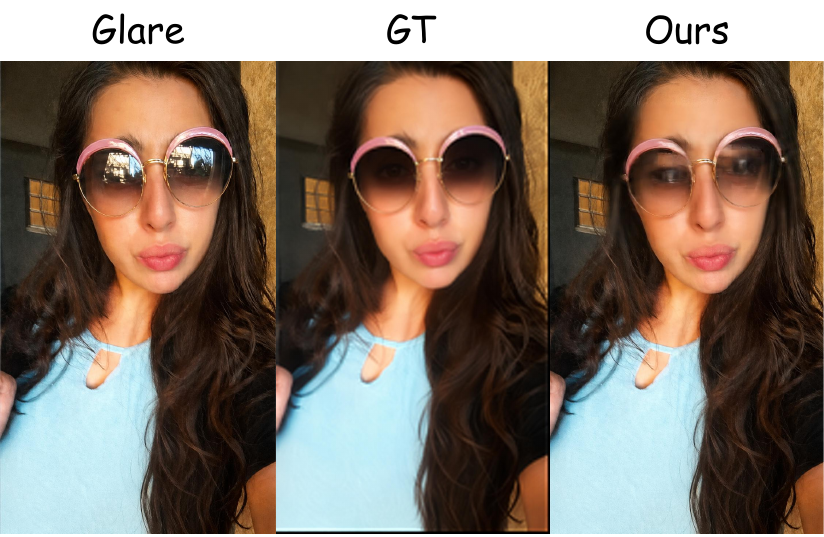}
	\caption{
		Failure case on sunglasses.
		OcuFlow reconstructs visible eye content behind the intentionally dark lenses, deviating from the clean reference.
	}
	\label{fig:failure_sunglasses}
\end{figure}

OcuFlow can fail on sunglasses, where reduced ocular visibility is caused by intrinsic lens attenuation rather than reflection alone. As shown in \cref{fig:failure_sunglasses}, the ocular-restoration prior encourages reconstruction of visible eye content behind the dark lenses, leading to a result that deviates from the clean reference. This failure mode indicates that future models should better distinguish reflection-induced occlusion from intrinsic lens properties such as dark tint or strong attenuation.

\subsection{Additional Ablation and Inference Analysis}
\label{app:additional_ablation}

The main paper studies the three high-level components of OcuFlow and the effect of training-data composition. We further examine training-time sampling and auxiliary objectives, semantic conditioning, the learned residual gates of the frequency-preserving renderer, and the behavior of multi-step inference. Unless otherwise specified, all variants are evaluated on the complete held-out paired OcuBench test set using the same metrics as in the main paper.


\subsubsection{Training and Conditioning Ablations}

\cref{tab:additional_training_ablation} complements the component study in \cref{tab:ablation_core} by isolating several training-time choices not included in the main-paper ablation.

\begin{table}[t]
	\centering
	\caption{Additional ablations of data sampling, auxiliary supervision, and semantic conditioning. All variants are evaluated on the complete paired OcuBench test set.}
	\label{tab:additional_training_ablation}
	\resizebox{\linewidth}{!}{%
		\begin{tabular}{l*{7}{c}}
			\toprule
			Setting
			& Mask-PSNR $\uparrow$
			& Mixed-PSNR $\uparrow$
			& Mask-SSIM $\uparrow$
			& Mixed-SSIM $\uparrow$
			& Eye-ROI LPIPS $\downarrow$
			& Eye-NME (\%) $\downarrow$
			& Iris-NME (\%) $\downarrow$ \\
			\midrule
			Balanced Source Sampling
			& 21.9821 & 23.3859 & 0.6588 & 0.7050
			& 0.1125 & 1.4409 & 1.6286 \\
			w/o Eye Gradient
			& 21.9915 & 23.3927 & 0.6592 & 0.7052
			& 0.1125 & 1.4374 & 1.6241 \\
			w/o Semantic Prior
			& 21.9883 & 23.3920 & 0.6593 & 0.7053
			& 0.1122 & 1.4305 & 1.6171 \\
			w/o Multi-scale Reconstruction ($\mathcal L_x$)
			& 21.3276 & 22.9074 & 0.6367 & 0.6939
			& \textbf{0.1015} & 1.6258 & 1.9470 \\
			w/o Preserve Loss
			& 21.9931 & 23.3945 & 0.6591 & 0.7052
			& 0.1128 & 1.4360 & 1.6255 \\
			\midrule
			\rowcolor{oursgreen}
			\textbf{Ours (Natural Sampling)}
			& \textbf{22.1054} & \textbf{23.4710}
			& \textbf{0.6633} & \textbf{0.7074}
			& 0.1035 & \textbf{1.4030} & \textbf{1.5863} \\
			\bottomrule
	\end{tabular}}
\end{table}

Natural sampling outperforms balanced source sampling across all seven reported metrics, indicating that retaining the empirical mixture of the training data is beneficial for the final model. Removing the eye-gradient objective, semantic prior, or preservation objective results in smaller but consistent degradation across reconstruction, perceptual, and ocular-geometry metrics. These components therefore provide complementary structural and preservation supervision in addition to the primary restoration objective.

The multi-scale reconstruction term $\mathcal L_x$ has a larger effect. Removing it reduces Mask-PSNR from $22.1054$ to $21.3276$~dB and Mixed-PSNR from $23.4710$ to $22.9074$~dB, while Mask-SSIM decreases from $0.6633$ to $0.6367$. Eye-NME and Iris-NME increase from $1.4030\%$ and $1.5863\%$ to $1.6258\%$ and $1.9470\%$, respectively. Eye-ROI LPIPS, in contrast, decreases from $0.1035$ to $0.1015$. This divergence shows that improvement in a single perceptual metric does not necessarily correspond to better reference-aligned ocular reconstruction. The multi-scale reconstruction term contributes substantially to pixel-level fidelity and ocular geometric consistency.


\subsubsection{Frequency-Renderer Gate Ablation}

The main-paper ablation progresses from direct pMF prediction to residual rendering and frequency-preserving synthesis. Here, we isolate the content-adaptive residual gates within the frequency renderer. Both variants use the same frequency decomposition, lens support, and native-resolution synthesis. The fixed variant uses the handcrafted mixing maps directly, while the learned variant predicts residual corrections to these maps.

For this ablation, we additionally report Out-PSNR, defined as PSNR computed over the complement of the eyeglass evaluation mask, $\overline{\mathcal M}$.

\begin{table}[t]
	\centering
	\caption{Ablation of the frequency-renderer gates. Both variants use the same native-resolution frequency decomposition and synthesis; the learned variant predicts content-adaptive residual corrections to the fixed mixing maps.}
	\label{tab:renderer_gate_ablation}
	\resizebox{\linewidth}{!}{%
		\begin{tabular}{l*{8}{c}}
			\toprule
			Setting
			& Mask-PSNR $\uparrow$
			& Mixed-PSNR $\uparrow$
			& Mask-SSIM $\uparrow$
			& Mixed-SSIM $\uparrow$
			& Eye-ROI LPIPS $\downarrow$
			& Eye-NME (\%) $\downarrow$
			& Iris-NME (\%) $\downarrow$
			& Out-PSNR $\uparrow$ \\
			\midrule
			Fixed Mixing Maps
			& 21.8975 & 23.3077 & 0.6502 & 0.6967
			& \textbf{0.0999} & 1.4075 & \textbf{1.5821} & 26.2977 \\
			\rowcolor{oursgreen}
			\textbf{Learned Residual Gates (Ours)}
			& \textbf{22.1054} & \textbf{23.4710}
			& \textbf{0.6633} & \textbf{0.7074}
			& 0.1035 & \textbf{1.4030} & 1.5863
			& \textbf{26.4147} \\
			\bottomrule
	\end{tabular}}
\end{table}

Learning the residual gates improves Mask-PSNR by $0.2079$~dB and Mixed-PSNR by $0.1633$~dB, while increasing Mask-SSIM and Mixed-SSIM by $0.0131$ and $0.0107$, respectively. Eye-NME decreases from $1.4075\%$ to $1.4030\%$, and Out-PSNR increases from $26.2977$ to $26.4147$~dB. Overall, the learned residual gates improve six of the eight reported metrics, indicating that content-adaptive refinement of the frequency mixing maps benefits both restoration within the eyeglass region and preservation outside it.

The fixed variant obtains slightly lower Eye-ROI LPIPS ($0.0999$ versus $0.1035$) and Iris-NME ($1.5821\%$ versus $1.5863\%$). Together with the training ablations above, these results show that individual perceptual and landmark metrics can favor slightly different reconstruction trade-offs. The learned gates provide stronger masked and mixed reconstruction fidelity while retaining comparable ocular perceptual and geometric quality.


\subsubsection{Multi-step Inference}

OcuFlow uses a single pMF network function evaluation by default. We additionally evaluate repeated MeanFlow updates with
$K\in\{1,2,4,8\}$ evaluations. For $K$ steps, the interval $[0,1]$ is divided uniformly as
\begin{equation}
t_k
=
1-\frac{k}{K},
\qquad
r_k
=
1-\frac{k+1}{K},
\qquad
k=0,\ldots,K-1,
\end{equation}
and the state is updated according to
\begin{equation}
\mathbf z_{r_k}
=
\mathbf z_{t_k}
-
(t_k-r_k)
\mathbf u_\theta
(\mathbf z_{t_k},r_k,t_k).
\end{equation}

\begin{table}[t]
	\centering
	\caption{Effect of the number of pMF network function evaluations (NFE).}
	\label{tab:multistep_ablation}
	\begin{adjustbox}{max width=\linewidth}
		\begin{tabular}{c*{8}{c}}
			\toprule
			NFE
			& Mask-PSNR $\uparrow$
			& Mixed-PSNR $\uparrow$
			& Mask-SSIM $\uparrow$
			& Mixed-SSIM $\uparrow$
			& Eye-ROI LPIPS $\downarrow$
			& Eye-NME (\%) $\downarrow$
			& Iris-NME (\%) $\downarrow$
			& Time (ms) $\downarrow$ \\
			\midrule
			\rowcolor{oursgreen}
			\textbf{1}
			& \textbf{22.1054}
			& \textbf{23.4710}
			& \textbf{0.6633}
			& \textbf{0.7074}
			& 0.1035
			& \textbf{1.4030}
			& \textbf{1.5863}
			& \textbf{56.4551} \\
			2
			& 21.4412 & 22.9939 & 0.6329 & 0.6917
			& 0.0826 & 1.4695 & 1.7152 & 60.3832 \\
			4
			& 21.3870 & 22.9516 & 0.6311 & 0.6907
			& \textbf{0.0822} & 1.4638 & 1.7025 & 62.0009 \\
			8
			& 21.3066 & 22.8885 & 0.6279 & 0.6888
			& 0.0826 & 1.4614 & 1.7057 & 72.6149 \\
			\bottomrule
		\end{tabular}
	\end{adjustbox}
\end{table}

\cref{tab:multistep_ablation} shows that additional MeanFlow evaluations do not progressively improve the reference-aligned restoration metrics. The 1-NFE setting achieves the highest Mask/Mixed PSNR and SSIM and the lowest Eye-NME and Iris-NME. Increasing to 2-NFE lowers Eye-ROI LPIPS from $0.1035$ to $0.0826$, but Mask-PSNR decreases by $0.6642$~dB and both ocular NME measures increase. The 4-NFE setting yields the lowest Eye-ROI LPIPS at $0.0822$, while PSNR and SSIM continue to decrease as the number of evaluations increases.

These results suggest that repeated MeanFlow updates change the reconstruction trade-off rather than progressively refining the 1-NFE solution. Multi-step inference produces lower LPIPS values but weaker pixel-level agreement and ocular geometric consistency with the clean reference. Runtime also increases with the number of evaluations, from $56.4551$~ms/image at 1-NFE to $60.3832$, $62.0009$, and $72.6149$~ms/image at 2-, 4-, and 8-NFE, respectively. We therefore use 1-NFE for the reported OcuFlow results.

\subsubsection{Inference-time Breakdown and Pipeline Parallelism}

\begin{wraptable}{r}{0.42\linewidth}
	\vspace{-\intextsep}
	\centering
	\caption{
		Steady-state runtime breakdown of OcuFlow under sustained pipelined execution ($56.4551$~ms/image).
	}
	\label{tab:runtime_breakdown}
	\resizebox{\linewidth}{!}{%
		\begin{tabular}{lrr}
			\toprule
			Component & Time (ms) & Share (\%) \\
			\midrule
			Exposed CPU-preparation wait & 17.6631 & 31.2869 \\
			Native-resolution renderer & 14.1714 & 25.1021 \\
			pMF & 12.5873 & 22.2962 \\
			Patch preparation & 5.3085 & 9.4030 \\
			PIL materialization & 2.8555 & 5.0580 \\
			Batch packing & 2.4048 & 4.2597 \\
			H2D transfer & 0.5426 & 0.9612 \\
			Quantization + D2H & 0.4070 & 0.7208 \\
			Other pipeline overhead & 0.5150 & 0.9122 \\
			\midrule
			\textbf{Steady-state total} & \textbf{56.4551} & \textbf{100.0000} \\
			\bottomrule
		\end{tabular}
	}
\end{wraptable}

The steady-state profile shows that the one-step pMF predictor accounts for $12.5873$~ms ($22.3\%$) of the measured $56.4551$~ms/image processing interval, while the native-resolution renderer accounts for $14.1714$~ms ($25.1\%$). The largest individual contribution is the $17.6631$~ms ($31.3\%$) of CPU preparation that remains exposed after asynchronous overlap. Within the renderer, partition-of-unity projection accounts for $66.1\%$ of the rendering cost, compared with $33.6\%$ for frequency/decode operations and $0.3\%$ for lens resizing, making projection the dominant rendering operation.

Pipeline parallelism substantially reduces the exposed cost of ocular-geometry and preparation work. The aggregate CPU preparation work corresponds to $220.3822$~ms/image. With four asynchronous preprocessing workers, the corresponding idealized preparation interval is approximately $55.0956$~ms/image. The remaining inference-stage work accounts for approximately $38.7920$~ms/image. Without CPU--GPU overlap, these two stages would yield an estimated processing interval of $93.8876$~ms/image.

Under sustained pipelined execution, CPU preparation overlaps with GPU inference, reducing the measured steady-state processing interval to $56.4551$~ms/image. Under ideal full overlap, throughput is bounded by the slower of the two pipeline stages, giving an idealized interval of $55.0956$~ms/image. The measured interval is therefore within approximately $2.5\%$ of this idealized bound.

\end{document}